%% file: main.tex
\documentclass[twoside,11pt]{article}

\usepackage{amsmath}
\usepackage{amssymb}
\usepackage{amsthm}

\usepackage{jmlr2e}
\usepackage{lmodern}

\usepackage[utf8]{inputenc} %
\usepackage[T1]{fontenc}    %
\usepackage{hyperref}       %
\usepackage{url}            %
\usepackage{booktabs}       %
\usepackage{amsfonts}       %
\usepackage{nicefrac}       %
\usepackage{microtype}      %
\usepackage{amsmath}
\usepackage{graphicx}
\usepackage{wrapfig}
\usepackage{caption}
\usepackage{booktabs}
\usepackage{multirow}
\usepackage{tikz}
\usepackage{enumitem}
\usepackage{natbib}

\usepackage{dsfont}
\usepackage{graphicx}
\usepackage{float}
\usepackage{wrapfig}
\usepackage{caption}
\usepackage{lipsum}
\usepackage{pdflscape}
\usepackage{xcolor}
\usepackage{footnote}
\usepackage{microtype}
\usepackage{graphicx}
\usepackage{subfig}
\usepackage{booktabs} %
\usepackage{wrapfig}
\usepackage{setspace}
\usepackage{hyperref}

\usepackage{makecell}
\usepackage{tabularx}
\usepackage[export]{adjustbox}
\usepackage{setspace}

\usepackage{algorithmic}
\usepackage{algorithm}
\usepackage[figuresright]{rotating}
\usepackage{bbm}

\let\proof\relax

\usepackage{zzhang}
\usepackage{lastpage}
\jmlrheading{24}{2023}{1-\pageref{LastPage}}{4/21; Revised
3/23}{6/23}{21-0434}{Zhen Zhang, Mohammed Haroon Dupty, Fan Wu, Javen Shi and Wee Sun Lee}
\ShortHeadings{Factor Graph Neural Networks}{Zhang, Dupty, Wu, Shi and Lee}

\graphicspath{{./figure/},{./}}

\newcommand*\samethanks[1][\value{footnote}]{\footnotemark[#1]}

\firstpageno{1}

\begin{document}

\title{Factor Graph Neural Networks}

\author{%
  \name Zhen Zhang\thanks{Equal contribution.} \email{zhen.zhang02@adelaide.edu.au} \\
  \addr Australian Institute for Machine Learning \&\\
  \addr University of Adelaide, Australia          
  \AND
  \name Mohammed Haroon Dupty\samethanks  \email dmharoon@comp.nus.edu.sg \\
  \addr
  National University of Singapore, Singapore
  \AND 
  \name Fan Wu \email fanw6@illinois.edu\\ 
  \addr University of Illinois at Urbana-Champaign, USA
  \AND 
  \name Javen Qinfeng Shi \email javen.shi@adelaide.edu.au \\
  \addr Australian Institute for Machine Learning \&\\
  University of Adelaide, Australia
  \AND 
  \name Wee Sun Lee \email leews@comp.nus.edu.sg \\
  \addr
  National University of Singapore, Singapore
}
\editor{Tommi Jaakkola}

\maketitle

\begin{abstract}%
  In recent years, we have witnessed a surge of Graph Neural Networks (GNNs), most of which can learn powerful representations in an end-to-end fashion with great success in many real-world applications. They have resemblance to Probabilistic Graphical Models (PGMs), but break free from some limitations of PGMs. By aiming to provide expressive methods for representation learning instead of computing marginals or most likely configurations, GNNs provide flexibility in the choice of information flowing rules while maintaining good performance. Despite their success and inspirations, they lack efficient ways to represent and learn higher-order relations among variables/nodes. More expressive higher-order GNNs which operate on k-tuples of nodes need increased computational resources in order to process higher-order tensors. We propose Factor Graph Neural Networks (FGNNs) to effectively capture higher-order relations for inference and learning. 
  To do so, we first derive an efficient approximate Sum-Product loopy belief propagation inference algorithm for discrete higher-order PGMs. We then neuralize the novel message passing scheme into a Factor Graph Neural Network (FGNN) module by allowing richer representations of the message update rules; this facilitates both efficient inference and powerful end-to-end learning. %
  We further show that with a suitable choice of message aggregation operators, our FGNN is also able to represent
  Max-Product belief propagation, providing a single family of architecture that can represent both Max and Sum-Product loopy belief propagation. Our extensive experimental evaluation on synthetic as well as real datasets demonstrates the potential of the proposed model. 
\end{abstract}


\section{Introduction}
\label{sec:introduction}
Deep neural networks are powerful function approximators that have been
extremely successful in practice. While fully connected networks are
universal approximators, successful networks in practice tend to be
structured, \eg, grid-structured convolutional neural networks 
and chain-structured gated recurrent
neural networks (\eg, LSTM, GRU). Graph neural networks~\citep{gilmer2017neural,xu2018powerful,yoon2019inference} have recently been successfully used with graph-structured data to capture pairwise dependencies between variables and to propagate the information to the entire graph. 

GNNs learn node representations by iteratively passing messages between nodes within their neighbourhood and updating the node embeddings based on the messages received. Though successful, these models are limited by the first order approximations they make in aggregating information from the neighbouring nodes. The dependencies in the real-world data are often of higher-order which cannot be captured by only pairwise modeling.
For example,
in LDPC encoding, the bits of a signal are grouped into several clusters and in each cluster, the parity of all bits is constrained to be equal to
zero \citep{zarkeshvari2002implementation}. For good performance, these higher-order constraints should be exploited in the decoding procedure. Furthermore, many naturally occurring graphs like molecules exhibit repeating substructures like motifs; atoms in a molecule additionally satisfy higher-order  valence constraints~\citep{wu2018moleculenet,agarwal2006higher}.
We can learn better node representations if we can design better message passing schemes that can directly utilize the higher-order dependencies that are not captured using pair-wise dependencies. 

Node interactions have also been modeled by Probabilistic Graphical Models (PGMs) \citep{wainwright2008graphical,koller2009probabilistic} wherein nodes are viewed as random variables with a factorized joint distribution defined over them. Research that focuses on such models has been directed towards finding approximate inference algorithms to compute node marginals or the most likely configuration of the variables. There is rich literature of theoretically grounded PGM inference algorithms to find the state of a node given its statistical relations with other variable nodes in the graph. Knowledge of such inference algorithms can provide good inductive bias if it can be encoded in the neural network. The inductive bias lets the network favour certain solutions over others, and if the bias is indeed consistent with the target task, it helps in better generalization~\citep{battaglia2018relational}. So, while we use deep learning methods to learn representations in an end-to-end setting, better generalization may be achievable for some tasks with the inductive bias of classical inference algorithms. Unfortunately, these approximate inference algorithms become inefficient at higher-order.

One such well-known algorithm to find approximate node marginals is Loopy Belief Propagation (LBP)~\citep{murphy2013loopy,pearl2014probabilistic} which operates on the factor graph data structure. A factor graph is a bipartite graph with a set
of variable nodes connected to a set of factor nodes; each factor node indicates the presence of dependencies among its connected variables. In this paper, we propose to leverage Sum-Product LBP to formulate the message passing updates and build a better graph representation learning model. To this end, we derive an efficient message passing algorithm based on LBP with arbitrary higher-order factors on discrete graphical models, with the assumption that higher-order factors are of low rank, parameterized in the form of a mixture of rank-1 tensors. The derived message passing updates only need two operations, matrix multiplication and Hadamard product, and their complexity grows linearly with the number of variables in the factor.
Furthermore, this parameterization can represent any factor exactly with a large enough set of rank-1 tensors; the number of rank-1 tensors required can grow exponentially for some problems but often in practice, a small number is sufficient for a good approximation. 

Further, we represent the message passing updates in a neural network module and unroll the inference over discrete graphical models as a computational graph. We allow the messages to be arbitrary real valued vectors (instead of being constrained to be positive as in LBP) and treat the messages as latent vectors in a network; the latent vectors produced by the network can then be used for learning the target task through end-to-end training. Instead of using just the product operations to aggregrate the set of latent vectors, we allow the use of other aggregrators, potentially even universal set functions, to provide more flexibility in representation learning. We refer to the process of unrolling the algorithm, relaxing some of the constraints, modifying some of the components to potentially make the network more powerful, and using the resulting network as a component for end-to-end learning as \emph{neuralizing} the algorithm. %
We call the neural module as a Factor Graph Neural Network (FGNN).

\begin{figure*}[t]
  \centering
  \includegraphics[width=0.9\textwidth]{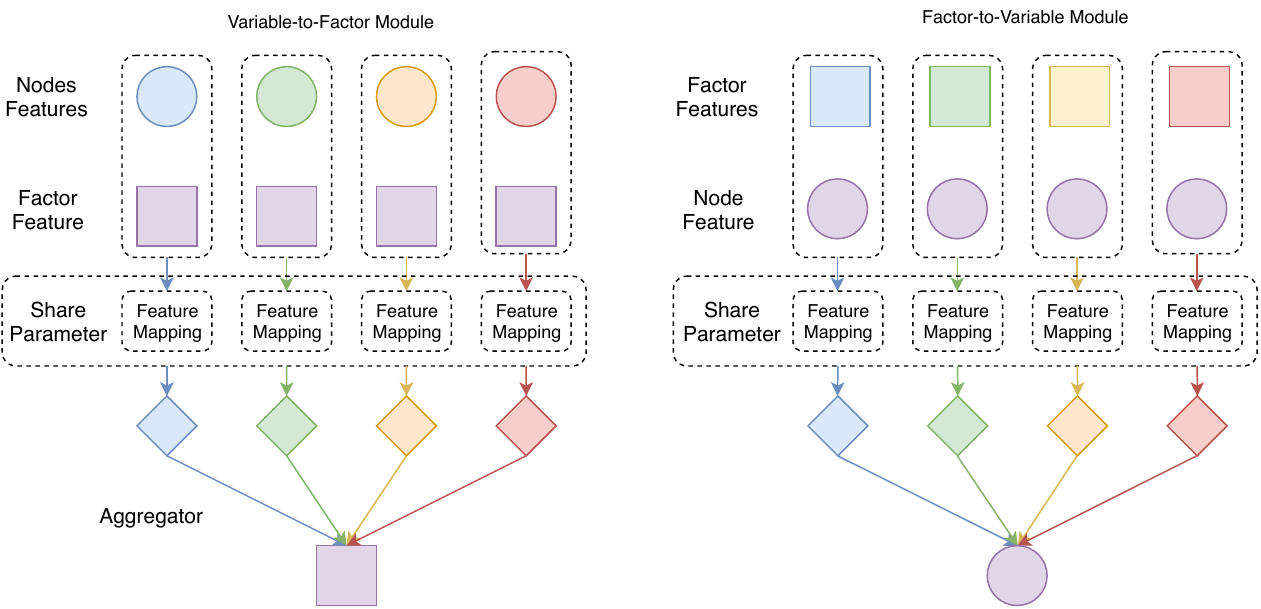}
  
  \caption{\small The structure of the Factor Graph Neural
    Network (FGNN): the Variable-to-Factor (VF) module is shown on the
    left and the Factor-to-Variable (FV) module is shown on the
    right. 
  }  \label{fig:FGNN}
\end{figure*}

The FGNN is
defined using two types of modules, the Variable-to-Factor (VF) module
and the Factor-to-Variable (FV) module (see
Figure~\ref{fig:FGNN}). These modules are combined into a layer, and
the layers are stacked together into an algorithm. Though the FGNN is motivated with Sum-Product LBP, we show that by using a different form of low-rank tensor representation and aggregator function, it is able to exactly parameterize the Max-Product Belief
Propagation, which is a widely used approximate
\map (MAP) inference algorithm for PGMs, as well. 
Theoretically, this shows that FGNN can represent both Max-Product and Sum-Product within a single architecture simply by changing the aggregator function; furthermore if a universal approximator is used as the aggregator, it would be able to learn to approximate the better of the two message passing algorithms on the problem.

The theoretical relationship with Sum-Product and Max-Product provides understanding on the representational capabilities of GNNs in general, and of FGNN in particular, \eg it can solve problems solvable by graphical model message passing algorithms \eg, \citep{bayati2008max,kim1983computational}.  From the practical perspective, the factor graph provides a flexible way for specifying dependencies among the variables, including higher-order dependencies. Furthermore, inference algorithms for many types of graphs, \eg, graphs with typed edges or nodes, are easily developed using the factor graph representation. Edges, or more generally factors, can be typed by tying together parameters of factors of the same type, or can also be conditioned from input features by making the edge or factor parameters a function of the features; nodes can similarly have types or features with the use of factors that depend on a node variable. With typed or conditioned factors, the factor graph can also be assembled dynamically for each graph instance. FGNN provides a flexible learnable architecture for exploiting these graphical structures---just as factor graph allows easy specification of different types of PGMs, FGNN allows easy specification of both typed and conditioned variables and dependencies as well as a corresponding data-dependent approximate inference algorithm. 

To be practically useful, the FGNN architecture needs to be
practically \emph{learnable} without being trapped in poor local minima.
We performed experiments to explore the
practical potential of FGNN on both problems where the graphical model inference aspects are clear, as well as problems where the FGNN is used mostly as a graph neural network that allows the structure of the higher-order interactions to be specified and exploited.
On problems closely related to PGM inference, we experimented with a synthetic higher-order PGM inference problem, LDPC decoding, as well as a graph matching problem. FGNN performed well on the synthetic PGM inference %
and outperforms the standard LDPC decoding method under some noise conditions. On the graph matching problem, FGNN outperforms both a graphical model approximate inference algorithm as well as a graph neural network. 
To show that FGNN can be used purely as a graph neural network to exploit higher-order dependencies, we conducted experiments on handwritten character recognition within words where there is strong correlation in the character sequences, human motion prediction where different joint positions are constrained by the human body structure and molecular property prediction where higher-order correlations in the molecular graphs are likely to be present. We demonstrate that FGNN is able to exploit the higher-order information with state-of-the-art results on human motion prediction. Furthermore, FGNN is able to outperform other recent $k$-order GNNs~\citep{morris2019weisfeiler,maron2019provably} substantially on two challenging large molecular datasets (QM9 and Alchemy).

\section{Related work} 
Belief Propagation (BP) inference algorithms have been used in variety of applications spanning computer vision, natural language processing and other machine learning domains. Due to the intractability of the algorithm in its higher-order form, first-order pairwise BP has been predominantly used in most problems. However, multiple works have proposed approaches to efficiently run higher-order BP in various settings~\citep{lan2006efficient,potetz2008efficient,kohli2010energy}. Most of the approaches are designed either with assumptions on input graph structures or with restrictions on the kind of higher-order functions being modelled. 
~\citet{lan2006efficient} used an adaptive state space to handle the increased complexity for higher-order 2x2 MRF clique structures. \citet{potetz2008efficient} used specific assumption of linear constraints on higher-order functions to efficiently run the BP equations. ~\citet{kohli2010energy} used linear envelope approximations to model the higher-order functions and showed its usefulness for semantic segmentation.%

The inspiration for our low-rank formulation of higher-order functions comes from~\citet{wrigley2017tensor} where higher-order potentials were decomposed with CP decompositions to efficiently run the sampling-based junction-tree algorithm. They used CP tensor decomposition to efficiently sample from higher-order functions in the junction-tree updates, whereas in our work, we assume tensor decomposition form of higher-order function and derive efficient message update equations for the LBP algorithm. Furthermore, our method is a deterministic approximation of LBP and we neuralize it to learn the parameters in an end-to-end training.

There have been a significant number of works which have tried to incorporate inference algorithms in deep networks~\citep{zheng2015conditional,chen2015learning, lin2016efficient, lin2015deeply,tamar2016value,karkus2017qmdp,xu2017scene}. A large number of these works focus on learning pairwise potentials of graphical models on top of CNN to capture relational structures among their outputs for structured prediction. These methods are largely focused on modeling for a specific task like semantic segmentation~\citep{zheng2015conditional,lin2016efficient}, scene graph prediction~\citep{xu2017scene}, and image denoising~\citep{wu2016deep}. On the other hand, ~\citep{tamar2016value,karkus2017qmdp} represent planning algorithms as neural network layers in sequential decision making tasks.

Various graph neural network models have been proposed for graph
structured data, including methods based  on the graph Laplacian
\citep{bruna2013spectral, defferrard2016convolutional,kipf2016semi},
gated networks \citep{li2015gated}, and  various other
neural networks structures for updating the information
\citep{duvenaud2015convolutional,battaglia2016interaction,kearnes2016molecular,schutt2017quantum}.
\citet{gilmer2017neural} show that these methods can be viewed as
applying message passing on pairwise graphs and are special cases of Message Passing Neural
Networks (MPNNs). 

There has been some recent work on extending the graph convolutional neural networks to hyper-graphs in order to capture higher-order information~\citep{feng2019hypergraph,yadati2019hypergcn, jiang2019dynamic,zhang2019hyper}. ~\citet{feng2019hypergraph,jiang2019dynamic} used clique-expansion of hyper-edges to extend convolutional operation to hyper-graphs. Such modeling is equivalent to decomposing an hyper-edge to a set of pairwise edges. Similar approximation is applied in~\citet{yadati2019hypergcn} where the number of pairwise edges added are reduced and are linearly dependent on the size of the hyperedge. Although, these methods operate on hyper-graphs, effectively the hyper-graphs are reduced to graphs with pairwise edges. 

Recently,~\citet{morris2019weisfeiler} and~\citet{maron2019provably} used Weisfeiler Lehman (WL) graph isomorphism tests to construct increasingly powerful GNNs. They proposed models of message passing called $k$-order GNNs to capture higher-order structures and compared their expressiveness with higher-order WL tests. In contrast to $k$-order GNNs which build on graph isomorphism testing, FGNN builds on probabilistic graphical models, which provide a rich modeling language allowing the designer to specify prior knowledge in the form of pairwise as well as higher-order dependencies in a factor graph. 
As $k$-order GNNs are theoretically shown to be more expressive in capturing higher-order information, we compare our model with the $k$-order GNNs on the molecular datasets. 

Running inference algorithms on graph neural networks has been previously explored in~\citet{yoon2019inference,dai2016discriminative} and \citet{chen2018supervised}. These works showed methods of running graphical model inference with GNN message passing updates. However, they operate with the assumption of pairwise graphical models and not more general higher-order models. In contrast, our work deals mainly with running the inference algorithms on the higher-order graphical model as a graph neural network. Following our initial work, \citet{satorras2020neural} proposed a graph neural network based on the factor graph for LDPC code decoding that also exploits the loopy belief propagation messages. However they did not connect message passing on factor graphs with higher-order factors represented using tensor decompositions as we have done which further gives a theoretical understanding of the relation between inference algorithms and GNNs. %

\section{Preliminaries}
In this section, we briefly review two concepts central to our approach, low rank tensor decomposition and the loopy belief propagation inference algorithm.

\subsection{Tensor decompositions}\label{sec:tensor_decomp}
Tensors are generalizations of matrices to higher dimensions. An order-$m$ tensor $\mathbf{T}$ is an element in $\mathbb{R}^{N_1\times N_2\cdots\times N_m}$ with $N_k$ possible values in $k^{th}$ dimension for $k \in \{1,2,\dots ,m\}$. Tensor rank decompositions provide succinct representation of tensors. 
In CANDECOMP / PARAFAC (CP) decomposition, a tensor $\mathbf{T}$ can be represented as a linear combination of outer products of vectors as
\begin{equation}
  \mathbf{T} = \sum_{r=1}^R \lambda^r w_{1}^r \otimes w_{2}^r \otimes \cdots \otimes w_{m}^r
  \label{eq:tensor}
\end{equation}
where $\lambda^r\in\mathbb{R}$, $w_{k}^r \in
\mathbb{R}^{N_k}$, $\otimes$ is the outer product operator,
\ie, $\mathbf{T}(i_1,i_2\dots,i_m)=\sum_{r=1}^R  \lambda^r
w_{1_{i_1}}^{r} w_{2_{i_2}}^{r} \cdots w_{m_{i_m}}^{r}$,
and the term $w_{1}^r \otimes w_{2}^r \otimes \cdots \otimes w_{m}^r$
is a rank-1 tensor. The scalar coefficients $\lambda^r$ can
optionally be absorbed into  $\{w_{k}^r\}$. The smallest $R$
for which an exact $R$-term decomposition exists is the rank
of tensor $\mathbf{T}$ and the decomposition (\ref{eq:tensor})
is its $R$-rank approximation. With this compact
representation, an exponentially large tensor $\mathbf{T}$
with ${N_1\times N_2\cdots\times N_m}$ entries can be
represented with $R$ vectors for each variable in
$\mathbf{T}$, \ie, with a total of $R(N_1+N_2+\dots+N_m)$
parameters. More information about tensor decompositions can
be found in~\citet{kolda2009tensor}, and~\citet{rabanser2017introduction}. %

\subsection{Graphical models and Loopy belief propagation}
\begin{wrapfigure}{r}{.5\textwidth}
  \centering
  \includegraphics[scale=0.175]{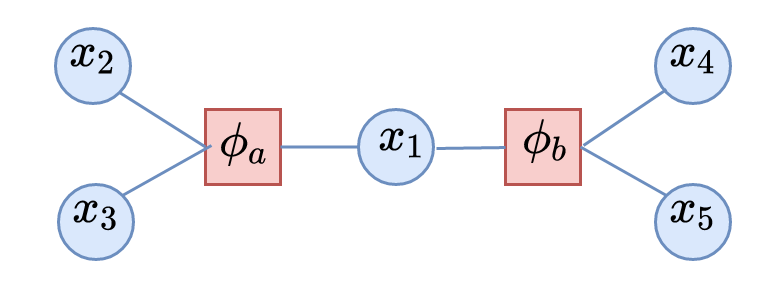}
  \par\vfill
  \caption{A factor graph where $\phi_a$ depends
    on $\{x_1, x_2,x_3\}$ while $\phi_b$ depends on $\{x_1, x_4, x_5\}$.}\label{fig:5}	
\end{wrapfigure}

Probabilistic Graphical Models (PGMs) use graphs to model dependencies among random variables. These dependencies are conveniently represented using a factor graph, which is a bipartite graph $\Gcal = (\Vcal,\Ccal,\Ecal)$
where each vertex $i\in\Vcal$ in the graph is associated with a random
variable $x_i \in \xb$, each vertex $c\in \Ccal$ is associated with a non-negative function $\phi_c$, and an edge connects a variable vertex $i$ to a factor vertex $c$ if $\phi_c$ depends on $x_i$. An example factor graph is shown in Figure~\ref{fig:5}.

We consider PGMs which model dependencies between discrete random variables. Let $\xb$ be the set of all variables and let $\xb_c$ be the subset of
variables that $\phi_c$ depends on.  The joint distribution of variables factorizes over $\mathcal{C}$ as %
\begin{align}
  P(\xb) = \frac{1}{Z} \prod_{c \in \mathcal{C}} \phi_c(\xb_c)
  && Z=\sum_{\xb}\prod_{c \in \mathcal{C}} \phi_c(\xb_c)
     \label{eq:pgm}
\end{align}
$Z$ is the normalizing constant. Without loss of generality, we assume all variables can take $d$ values and consequently $\phi_c(\xb_c) \in \mathbb{R}^{d^{|\xb_c|}}$. %

\paragraph{Marginal and MAP Inference} In this paper we
consider two main inference tasks --- the \emph{marginal} inference and the
\emph{maximum a posteriori} (MAP) inference. The aim of marginal inference is to
compute the marginals $p_i(x_i)$
\begin{align}
  \label{eq:marginal_inference}
  p_i(x_i) = \sum_{\xb_{\Vcal \setminus \{i\}}} P(\xb) = \sum_{\xb_{\Vcal \setminus \{i\}}}\frac{1}{Z} \prod_{c \in \mathcal{C}} \phi_c(\xb_c),
\end{align}
and the aim of MAP inference is to find the assignment which maximizes
$P(\xb)$, that is
\begin{align}
  \label{eq:map_inference}\nonumber
  \xb^{\ast} &= \argmax_{\xb}\prod_{c\in\mathcal{C}}\phi_c(\xb_c)\\
  &= \argmax_{\xb}\sum_{c\in\mathcal{C}}\log\phi_c(\xb_c)
\end{align}

\paragraph{Loopy Belief Propagation}
The marginal inference and MAP inference problem are  NP-hard
in general, and thus approximation algorithms are usually required. 
Different versions of the loopy belief propagation
(LBP) algorithms ~\citep{pearl2014probabilistic,murphy2013loopy} compute approximate marginals $p(x_i)$ at each node $x_i$, or the
approximate MAP assignment, by sending messages
between factor and variables nodes  on a factor graph. First we introduce the Sum-Product
loopy belief propagation. 
Essentially, the Sum-Product LBP starts by initializing two kinds of messages, factor-to-variable $m_{c\rightarrow i}(x_i)$ and variable-to-factor $m_{i\rightarrow c}(x_i)$. Messages are a function of the variable in the variable node, updated with the following recursive equations,
\begin{equation}
  m_{i\rightarrow c }(x_i) = \prod_{d \in  N(i) \setminus \{c\}} m_{d\rightarrow i} (x_i)
  \label{eq:msg_i_a} 
\end{equation}

\begin{figure}[t]
    \centering
    \subfloat[Variable to Factor message]{\includegraphics[scale=0.135]{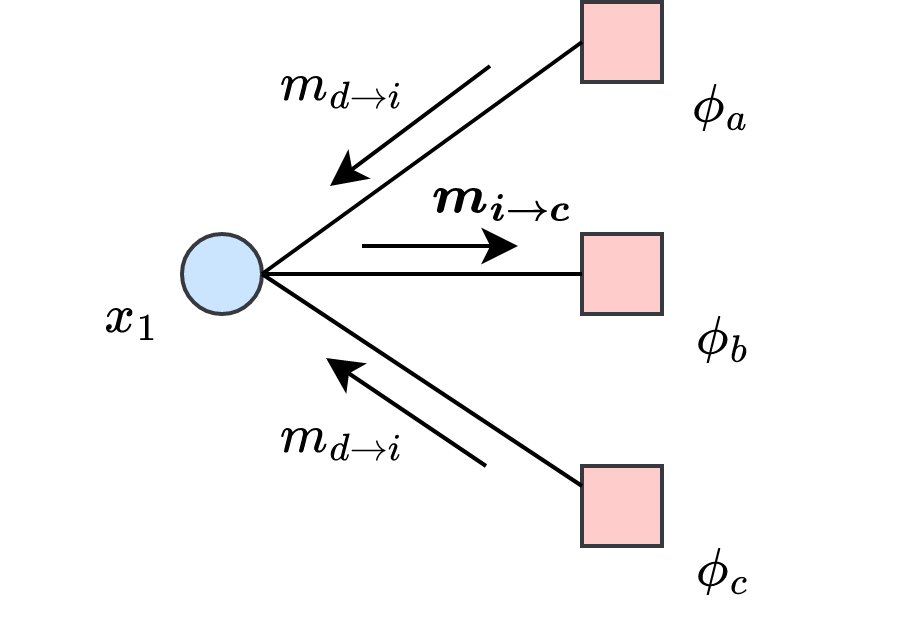}}
    \hfil
    \subfloat[Factor to Variable message]{\includegraphics[scale=0.135]{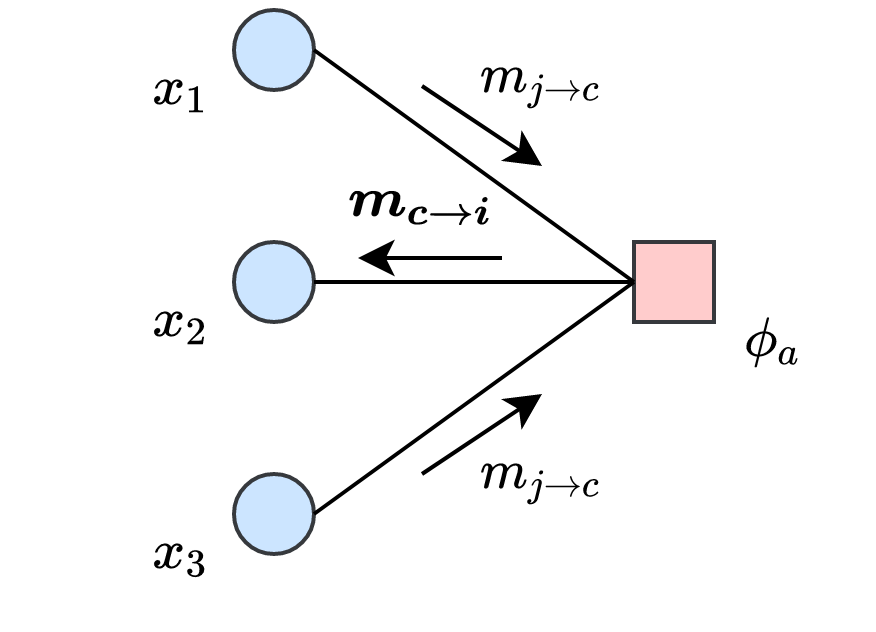}}
    \caption{Loopy Belief Propagation messages}\label{fig:factor_graph_messages}
\end{figure}
        
\begin{equation}
  m_{c\rightarrow i}(x_i) = \sum_{\xb_c \setminus \{x_i\}} \phi_c(\xb_c) \prod_{j \in  N(c) \setminus \{i\}} m_{j\rightarrow c} (x_j)
  \label{eq:msg_a_i}
\end{equation}
where $N(i)$ is the set of neighbours of $i$ and messages $m_{i\rightarrow c}, m_{c\rightarrow i}\in \mathbb{R}^d$. As illustrated in Figure~\ref{fig:factor_graph_messages}, $m_{i\rightarrow c}$ is the message from variable $i$ to factor $c$ and $ m_{c\rightarrow i}$ is the message from factor $c$ to variable $i$. After sufficient number of iterations, the belief of variables is computed by
\begin{equation}
  b_i(x_i) = \prod_{c \in N(i)} m_{c\rightarrow i} (x_i).
  \label{eq:bel_i}
\end{equation}

\noindent Similarly, the Max-Product belief propagation algorithm can be
formulated as 
      
\begin{align}
\label{eq:mp_lbp1}
    m_{i\rightarrow c }(x_i)& = \prod_{d \in  N(i) \setminus \{c\}}
    m_{d\rightarrow i} (x_i)\\     \label{eq:mp_lbp2}
  m_{c\rightarrow i}(x_i) &= \max_{\xb_c \setminus \{x_i\}}
                            \phi_c(\xb_c) \prod_{j \in  N(c) \setminus
                            \{i\}} m_{j\rightarrow c} (x_j)\\     \label{eq:mp_lbp3}
    b_i(x_i) &= \prod_{c \in N(i)} m_{c\rightarrow i} (x_i).
\end{align}

\noindent Performed in log space, the product operator in Eqns~\ref{eq:mp_lbp1}, \ref{eq:mp_lbp2}, \ref{eq:mp_lbp3} becomes sum and we have the Max-Sum algorithm which may be better behaved numerically.

\section{Proposed Method}
In this section, we derive the FGNN model through neuralizing the
Sum-Product loopy belief propagation that utilizes the low rank decomposition of higher-order potentials. Then we show that with a slight
modification the model can also mimic the Max-Sum or equivalently the Max-Product belief propagation. 
\subsection{Low Rank Sum-Product Loopy Belief Propagation}
We start the derivation by writing the LBP equations in vectorized form. Consider a factor $\phi_c(\xb_c)$ over $n_c$ number of variables \ie $\xb_c = [x_1, x_2 \dots, x_{n_c}]$. Then the message update equations are,
\begin{equation}
m_{i\rightarrow c } = \prod_{d \in  N(i) \setminus \{c\}} m_{d\rightarrow i}
\label{eq:msg_i_a_vec} 
\end{equation}
\begin{equation}
m_{c\rightarrow i} = \sum_{\xb_c \setminus \{x_i\}} \phi_c(\xb_c) \prod_{j \in  N(c) \setminus \{i\}} m_{j\rightarrow c}
\label{eq:msg_a_i_vec}
\end{equation}
where $\phi_c(\xb_c) \in \mathbb{R}^{d^{n_c}}$ and the messages $m_{j\rightarrow c}, m_{c\rightarrow i}$ are in $\mathbb{R}^d$. A tensorized way to implement Equation~(\ref{eq:msg_a_i_vec}) would be to take the outer product of all incoming messages $m_{j\rightarrow c}$, expand a dimension corresponding to the dimension of the $i^{th}$ variable and elementwise multiply with the $\phi_c(\xb_c)$ tensor. Then, we can marginalize all variables except $x_i$ to get the message $m_{c\rightarrow i}$ \ie, 
\begin{equation}
m_{c\rightarrow i} = \sum_{\xb_c \setminus \{x_i\}} \phi_c(\xb_c) \odot \big(  m_{1\rightarrow c}  \otimes \dots \otimes  \mathbf{1}_{i\rightarrow c} \otimes \dots \otimes m_{n_c\rightarrow c} \big)
\label{eq:msg_a_i_later}
\end{equation}
where $\mathbf{1}_{i\rightarrow c}$ is a vector of ones in $\mathbb{R}^d$, $\otimes$ is the outer product and $\odot$ is elementwise multiplication or Hadamard product.
Note that the summation operator $\sum_{\xb_c \setminus \{x_i\}}$ performs ``marginalization'' operation or summation across all dimensions except the $i^{th}$ dimension corresponding to variable $x_i$ which reduces the order-${n_c}$ tensor to an $R^d$ vector.

Since $\phi_c(\xb_c)$ is a tensor in $\mathbb{R}^{d^{n_c}}$, it can be represented as a sum of $R$ fully factored terms in CP decomposition form (\ref{eq:tensor}). 
\begin{equation}
\phi_c(\xb_c) = \sum_{r=1}^R \lambda_r w_{c,1}^r \otimes w_{c,2}^r \otimes \cdots \otimes w_{c,{n_c}}^r
\label{eq:fa_tensor}
\end{equation}
where $w_{c,j}^r \in \mathbb{R}^d$ and $\lambda_r$ are real-valued scalars. This representation is efficient if $\phi_c$ is of low-rank i.e., $R \ll d^{n_c}$, and
in that case, a $d^{n_c}$ dimensional tensor is compressed into a set of $n_c\cdot R$ vectors of $d$ dimensons each. Please refer Section~\ref{sec:tensor_decomp} on Preliminaries for details on CP Tensor Decomposition.

\begin{figure}[t]
  \centering
  \includegraphics[width=0.9\linewidth]{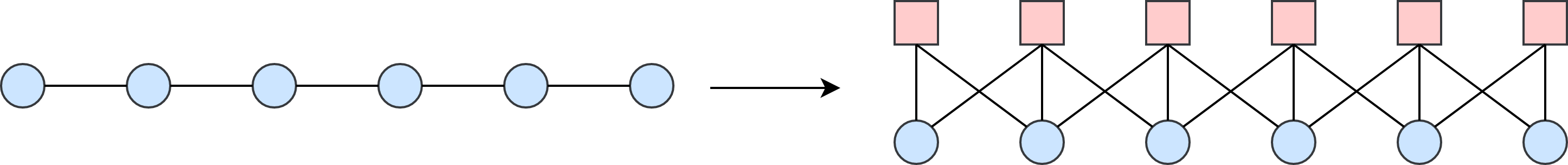}
  \caption{Factor Graph model for the Handwriting character recognition.}
    \label{fig:ocr_graph_b}
\end{figure}

We posit that such a low-rank representation is often a good approximation of $\phi_c$ in practice; the method is likely to be useful when this assumption holds. 
Previously, such low rank approximations have been shown to be useful
in variety of real world tasks including semantic segmentation and
knowledge graph
embedding~\citep{wrigley2017tensor,kohli2010energy,trouillon2017knowledge}. Moreover,
we provide supporting evidence on the low-rank assumption in our
ablation experiments in Section~\ref{sec:handwriting-char-recog}
and~\ref{sec:molecular_data} (See Figure \ref{fig:ocr_graph_b} and \ref{fig:qm9_mae-rank}).

Absorbing $\lambda_r$ in~(\ref{eq:fa_tensor}) into weights $\{
w_{c,j}^r \}$ and substituting in (\ref{eq:msg_a_i_later}), we have
\begin{subequations}
	\begin{align}
	m_{c\rightarrow i} = &\sum_{\xb_c \setminus \{x_i\}} \Big(\sum_{r=1}^R w_{c,1}^r \otimes w_{c,2}^r \otimes \cdots \otimes w_{c,{n_c}}^r \Big) %
	\odot \Big( m_{1\rightarrow c} \otimes \dots  \otimes \mathbf{1}_{i\rightarrow c} \otimes \dots \otimes m_{n_c\rightarrow c} \Big)
	\label{eq:lbp_1}\\
	=& \sum_{\xb_c \setminus \{x_i\}} \sum_{r=1}^R \big( w_{c,1}^r \odot m_{1\rightarrow c} \big) \otimes \cdots w_{c,i}^r \cdots \otimes  \big(w_{c,{n_c}}^r \odot m_{{n_c}\rightarrow c}\big) 
	\label{eq:lbp_2}
	\end{align}
\end{subequations}
In Equation~(\ref{eq:lbp_2}), we have used the distribution rule \ie $\forall \;u,v\in\mathbb{R}^d$ and $u',v'\in\mathbb{R}^{d'}$, then $(u\otimes u')\odot(v\otimes v') = (u\odot v) \otimes (u'\odot v')$.

The variables are grouped together with the factor parameters corresponding to them. %
Now we can marginalize out a variable easily as we have a sum of fully
factorized functions. We simply push the outer summation inside,
distribute and separately evaluate it over each of the univariate
products $\big(w_{c,j}^r \odot m_{j\rightarrow c} \big)$. This
gives us,
\begin{subequations}
	\begin{align}
	m_{c\rightarrow i} =&  \sum_{r=1}^R \sum_{\xb_c \setminus \{x_i\}} \big( w_{c,1}^r \odot m_{1\rightarrow c} \big) \otimes \cdots w_{c,i}^r \cdots \otimes  \big(w_{c,{n_c}}^r \odot m_{{n_c}\rightarrow c}\big) 
	\label{eq:lbp_2_1}\\
	=& \sum_{r=1}^R w_{c,i}^r   \Big(\sum_{x_1} w_{c,1}^r \odot m_{1\rightarrow c} \Big) \cdots \Big( \sum_{x_{n_c}} w_{c,{n_c}}^r\odot m_{{n_c}\rightarrow c} \Big)  \label{eq:lbp_3} \\
	= &\sum_{r=1}^R  w_{c,i}^r \gamma_{c,1}^r \cdots \gamma_{c,{n_c}}^r
	\qquad\qquad\qquad\qquad\quad\quad ;\text{with } \gamma_{c,i}^r=1\label{eq:lbp_4}\\
	= &\sum_{r=1}^R  w_{c,i}^r  \boldsymbol{\gamma}_c^r \label{eq:lbp_4_1}
	\end{align}
\end{subequations}
In Equation~(\ref{eq:lbp_2_1}), we have swapped the summations and in Equation~(\ref{eq:lbp_3}), have distributed the summation $\sum_{\xb_c \setminus \{x_i\}}$ over each variable. In Equation~(\ref{eq:lbp_4}), we evaluate the summation over each variable to get $\gamma_{c,j}^r$ \ie $(\sum_{x_j} w_{c,j}^r \odot m_{j\rightarrow c}) = w_{c,j}^{r^T} m_{j\rightarrow c} = \gamma_{c,j}^r \in \mathbb{R}$, which is a scalar. To rule out the message from variable $x_i$, we set $\gamma_{c,i}^r=1$ and therefore, the product
$\boldsymbol{\gamma}_c^r = \gamma_{c,1}^r \cdot \gamma_{c,2}^r \dots \gamma_{c,{n_c}}^r  \in \mathbb{R}$ is also a scalar in Equation~(\ref{eq:lbp_4_1}).
 
Since $m_{c\rightarrow i}$ is a linear combination of $R$ number of $w_{c,i}^r$ vectors in Equation~(\ref{eq:lbp_4_1}), we can rewrite it in matrix form. For this, we stack the $R$ component weight vectors for each variable as matrix $\mathbf{W}_{c,i} = [ w_{c,i}^1,w_{c,i}^2,\dots, w_{c,i}^R] \in \mathbb{R}^{d\times R}$.
Similarly,  we stack $R$ number of $\boldsymbol{\gamma}_c^r$'s together as vector  
$\boldsymbol{\Gamma}_{c} = [\boldsymbol{\gamma}_c^1, \boldsymbol{\gamma}_c^2, \dots, \boldsymbol{\gamma}_c^R]^T\in \mathbb{R}^{R\times 1}$. Then, we can rewrite the Equation (\ref{eq:lbp_4_1}) in matrix form as 
\begin{equation}
m_{c\rightarrow i} = \mathbf{W}_{c,i}\boldsymbol{\Gamma}_c 
\label{eq:lbp_5}
\end{equation}

Note that since $\boldsymbol{\Gamma}_{c} = [\boldsymbol{\gamma}_c^1, \boldsymbol{\gamma}_c^2, \dots, \boldsymbol{\gamma}_c^R]^T \in \mathbb{R}^{R\times 1}$ where each $\boldsymbol{\gamma}_c^r$ is a product of $\gamma_{c,j}^r$'s \ie $\boldsymbol{\gamma}_c^r = \gamma_{c,1}^r \cdot \gamma_{c,2}^r \dots \gamma_{c,{n_c}}^r$ (from Equations~(\ref{eq:lbp_4}),~(\ref{eq:lbp_4_1}), we can rewrite $\boldsymbol{\Gamma}_{c}$ as elementwise product of vectors in each variable as:

\begin{gather}
\boldsymbol{\Gamma}_{c} 
 =
\begin{bmatrix}
\boldsymbol{\gamma}_c^1 \\
\boldsymbol{\gamma}_c^2 \\
\vdots\\
\boldsymbol{\gamma}_c^R \\
\end{bmatrix}
=
\begin{bmatrix}
\gamma_{c,1}^1 \cdot \gamma_{c,2}^1 \dots \gamma_{c,{n_c}}^1\\
\gamma_{c,1}^2 \cdot \gamma_{c,2}^2 \dots \gamma_{c,{n_c}}^2\\
\vdots\\
\gamma_{c,1}^R \cdot \gamma_{c,2}^R \dots \gamma_{c,{n_c}}^R\\
\end{bmatrix}
=
\begin{bmatrix}
\gamma_{c,1}^1 \\
\gamma_{c,1}^2 \\
\vdots\\
\gamma_{c,1}^R \\
\end{bmatrix}
\odot
\begin{bmatrix}
\gamma_{c,2}^1 \\
\gamma_{c,2}^2 \\
\vdots\\
\gamma_{c,2}^R \\
\end{bmatrix}
\dots 
\odot
\begin{bmatrix}
\gamma_{c,{n_c}}^1 \\
\gamma_{c,{n_c}}^2 \\
\vdots\\
\gamma_{c,{n_c}}^R \\
\end{bmatrix}
\end{gather}
This gives us, 
\begin{equation}
    \boldsymbol{\Gamma}_c = [\Gamma_{c,1} \odot \Gamma_{c,2} \dots \odot \Gamma_{c,{n_c}}]
    \label{eq:lbp_gamma_mat}
\end{equation}
where each $\Gamma_{c,j} = [\gamma_{c,j}^1, \gamma_{c,j}^2, \dots, \gamma_{c,j}^R]^T \in \mathbb{R}^{R\times 1}$. Note that from Equation~ (\ref{eq:lbp_4}), $\boldsymbol{\Gamma_{c}}$ contains $\Gamma_{c,i}$ as a vector of all ones \ie $\Gamma_{c,i} = [\gamma_{c,i}^1, \gamma_{c,i}^2, \dots, \gamma_{c,i}^R]^T=[1, 1, \dots, 1]^T$. Therefore, Equation~(\ref{eq:lbp_5}) can now be written as

\begin{equation}
m_{c\rightarrow i} = \mathbf{W}_{c,i} \big[\Gamma_{c,1} \odot \Gamma_{c,2} \dots \odot \Gamma_{c,{n_c}} \Big]
\label{eq:lbp_5_1}
\end{equation}
\noindent Furthermore, from Equation~(\ref{eq:lbp_4}) we have $\gamma_{c,j} =\sum_{x_j} w_{c,j}^r  m_{j\rightarrow c} =  w_{c,j}^{r^T} m_{j\rightarrow c}$. Therefore, we can write $\Gamma_{c,j}$ as
\begin{subequations}
  \begin{align}
   \Gamma_{c,j} =&\; [\gamma_{c,j}^1, \gamma_{c,j}^2, \dots, \gamma_{c,j}^R]^T\\
                =&\; [w_{c,j}^{1},w_{c,j}^{2},\dots,w_{c,j}^R]^T m_{j\rightarrow c}\\
    =&\; \mathbf{W}_{c,j}^{T} m_{j\rightarrow c}
  \end{align}
\end{subequations}

Now, combining with Equation~(\ref{eq:lbp_5_1}) we have new message passing updates for the low-rank loopy belief propagation algorithm,
\begin{subequations}
  \begin{equation}    \label{eq:lbp_msg_a}
    m_{c\rightarrow i} = \mathbf{W}_{c,i} \Big[\big(\mathbf{W}_{c,1}^{T} m_{1\rightarrow c} \big) \odot \big(\mathbf{W}_{c,2}^{T} m_{2\rightarrow c} \big)  \dots \odot \big(\mathbf{W}_{c,{n_c}}^{T} m_{{n_c}\rightarrow c} \big) \Big]_{\setminus i}
  \end{equation}
  \begin{equation}
    m_{i\rightarrow c } = \bigodot_{d \in  N(i) \setminus \{c\}} m_{d\rightarrow i} 
    \label{eq:lbp_msg_b} 
  \end{equation}
\end{subequations}
Belief update is simple. For the variable $x_i$, we project the messages from the other variables sharing a factor with $x_i$ to $\mathbb{R}^R$, perform elementwise multiplication and then project the product back to $\mathbb{R}^d$. Thereafter, multiply such messages from all the factors $x_i$ is connected to get the updated belief of $x_i$. 

Clearly, the computational complexity of the message updates grows linearly with the addition of variables to factors and thereby the algorithm is efficent enough to run with higher-order factors.

\subsection{Neuralizing Low Rank Sum-Product LBP}
To learn better node and consequently, graph representations in an end-to-end setting, we seek to neuralize the Low Rank LBP algorithm by writing the message passing updates as a functionally equivalent  neural  network module and \textit{unroll} the inference algorithm as a computational graph. We further replace the positive message vectors in LBP with unconstrained real valued hidden latent vector representations initialized from a feature extractor network.

In the Low-rank LBP algorithm, a factor is parameterized by the set of matrices $\{\mathbf{W}_{c,j}\}$ i.e. it has as many $\mathbf{W}_{c,j}$'s as the number of variables adjacent to it in the factor graph. We can relax this constraint and maintain $2|\{\mathbf{W}_{c,j}\}|$ matrices at each factor, one set is used for transformation of messages before Hadamard product and the other set after the product in equation (\ref{eq:lbp_msg_a}). The additional parameters are helpful as the two message updates are not tied with shared parameters and can be run parallelly. Also, with more parameters, this may increase the representative power of the neural-network while still being able to represent equation (\ref{eq:lbp_msg_a}) as the extra set of matrices can be learnt to be same as the first set.%
With this relaxation, we can push the outer $\mathbf{W}_{c,i}$ to
equation (\ref{eq:lbp_msg_b}) and rewrite the LBP updates of
(\ref{eq:lbp_msg_a}) and (\ref{eq:lbp_msg_b}) as,
\begin{subequations}
\begin{equation}
  m_{c\rightarrow i} =  
  \bigodot_{j \in  N(c) \setminus \{i\}} 
  \mathbf{W}_{c,j}^{T} m_{j\rightarrow c}  
  \label{eq:lbpnet_msg_a}
\end{equation}

\begin{equation}
  m_{i\rightarrow c} = \bigodot_{d \in  N(i) \setminus \{c\}} \mathbf{W}_{d,i} m_{d\rightarrow i} 
  \label{eq:lbpnet_msg_b} 
\end{equation}
\end{subequations}
The message updates of (\ref{eq:lbpnet_msg_a}) and  (\ref{eq:lbpnet_msg_b}) only involve operations like matrix-vector product and elementwise multiplication. These operations can very well be represented in a neural network for end-to-end learning. But the multiplication of several terms can lead to numerical instability in learning due to overflow and underflow errors. Therefore, we replace the Hadamard product with other generalized set-function aggregators. It has been shown that given a set of input vectors, an MLP followed by \emph{sum}~\citep{zaheer2017deep} or \emph{max}~\citep{qi2017pointnet} aggregator is a universal set approximator; it can approximate any non-linear set function and hence can be used in place of Hadamard product, approximating the Hadamard product if necessary, and providing the capability to possibly learn a better aggregator than the Hadamard product.  %
To provide even more approximation capabilities, we allow a multi-layer perceptron (MLP) to transform the message after aggregation just like typical graph neural networks. 

\begin{figure}
\begin{center}
\includegraphics[scale=0.5]{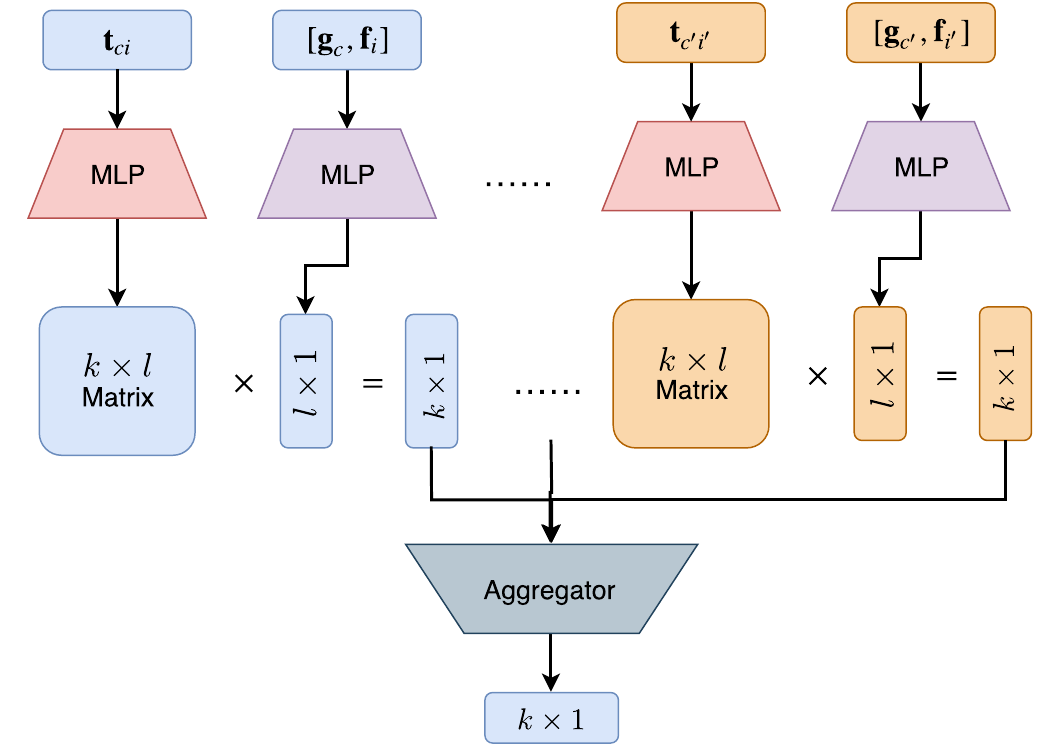}
\caption{A single aggregator of the FGNN architecture.}\label{fig:FGNN_agg}
\end{center}
\end{figure}

A single aggregator of FGNN is shown in Figure~\ref{fig:FGNN_agg}. For a message between node $c$ and node $i$, we optionally allow both the the latent vectors from $c$ and $i$ to be used as input to the MLP at node $c$. %
The matrix $\mathbf{W}$ that is used to multiply the message can be learned directly, if necessary. Alternatively, if some feature $t_{ci}$ is available, the $\mathbf{W}_{c,i}$ can be conditioned on $t_{ci}$ by using a MLP to output the matrix conditioned on the feature.

For implementing Equations~(\ref{eq:lbpnet_msg_a}) and (\ref{eq:lbpnet_msg_b}), the nodes in Figure~\ref{fig:FGNN_agg} correspond to messages $m_{c\rightarrow i}$ and $m_{i\rightarrow c}$. If we want to use the architecture as a graph neural network instead, it is convenient to define the latent vectors to correspond to factor and variable nodes as it would result in a smaller network. This can be done if we do not exclude the vector from the target node in the aggregration operation, i.e. use $m_{c} =  \bigodot_{j \in  N(c)} \mathbf{W}_{c,j}^{T} m_{j}$ and $m_{i} = \bigodot_{d \in  N(i)} \mathbf{W}_{d,i} m_{d}$ as the starting points for neuralization instead. The original belief propagation equations are exact inference algorithms when there are no loops. For correctness, the information in each message is only used once for computing a node marginal. As a graph neural network, the network is trained end-to-end and can be trained to account for the repeated information. We find experimentally that the results are similar when we use networks where nodes correspond to the belief propagation messages and networks where the nodes correspond to variables and factors. Hence, we use the networks where nodes correspond to variables and factors except for theoretical results where we are simulating the Sum-Product and Max-Product algorithms. 

\subsection{Factor Graph Neural Network}
\label{sec:fact-mess-pass}
We now describe the Factor Graph Neural Network (FGNN), a graph neural network model based on the derived low-rank LBP equations. It consists of two modules, Variable-to-Factor module and Factor-to-Variable module.
Given a factor graph $\Gcal = (\Vcal,\Ccal,\Ecal)$, unary features $[\fb_i]_{i \in \Vcal}$ and factor features $[\gb_c]_{c\in\Ccal}$, assume that for each edge $(c, i)\in \Ecal$, with $c\in\Ccal, i \in \Vcal$, there is an associated edge feature vector $[t_{ci}]$. Let $k$ and $l$ be hidden dimensions of variable and factor node embeddings, respectively. Then, the pseudo code for a FGNN layer on $\Gcal$ is shown in Algorithm \ref{algo:FGNN}, where $[\Phi_{\text{VF}}, \Theta_{\text{VF}}]$ are parameters for the
Variable-to-Factor module, and $[\Phi_{\text{FV}},
\Theta_{\text{FV}}]$ are parameters for the Factor-to-Variable
module. 
The factor to variable messages are computed as $\tilde \fb_i = \AGG\limits_{c:(c,i) \in \Ecal}\Qcal(\tb_{ci}|\Phi_{\text{FV}})\Mcal([\ \gb_c, \fb_i]|\Theta_{\text{FV}})$
where $\Mcal$ is a neural network mapping feature vectors to a length-$k$ feature vector,
and $\Qcal(\eb_{ij})$ is a neural network mapping $\eb_{ij}$ to a $k\times l$ matrix. Then by matrix multiplication and aggregation operator $\AGG$, a new length-$l$ variable feature $\tilde \fb_i$ is generated. Similarly, the variable to factor messages are computed as $\tilde \gb_c \hspace{-1mm}= \AGG\limits_{i:(c,i) \in \Ecal}\Qcal(\tb_{ci}|\Phi_{\text{VF}})\Mcal([\gb_c, \fb_i ]|\Theta_{\text{VF}})$ to generate the updated factor feature $\tilde \gb_c$.

      \begin{algorithm}[h]
        \small
        \setstretch{1.75}
        \renewcommand{\algorithmicrequire}{\textbf{Input:}}
        \renewcommand{\algorithmicensure}{\textbf{Output:}}
        \caption{\small The FGNN layer}\label{algo:FGNN}
        \begin{algorithmic}[1]
          \REQUIRE $\Gcal=(\Vcal,\Ccal,\Ecal),[\fb_i]_{i \in\Vcal},[\gb_c]_{c\in\Ccal},[t_{ci}]_{(c,i) \in\Ecal}$
          \ENSURE $[\tilde \fb_i]_{i \in\Vcal}$, $[\tilde
          \gb_c]_{c\in\Ccal}$
          \STATE \textbf{Variable-to-Factor:}
          \STATE \quad$\tilde \gb_c \hspace{-1mm}= \AGG\limits_{i:(c,i) \in \Ecal}\Qcal(\tb_{ci}|\Phi_{\text{VF}})\Mcal([\gb_c, \fb_i ]|\Theta_{\text{VF}})$
          
          \STATE \textbf{Factor-to-Variable:}
          \STATE \quad $\tilde \fb_i = \AGG\limits_{c:(c,i) \in \Ecal}\Qcal(\tb_{ci}|\Phi_{\text{FV}})\Mcal([\ \gb_c,
          \fb_i]|\Theta_{\text{FV}})$
        \end{algorithmic}
      \end{algorithm}
      
By stacking multiple FGNN layers together, we obtain a FGNN that transforms the initial factor graph features to variable and factor embeddings. To assist learning, we can add other architecture details such as residual connections. As graph neural networks, unlike belief propagation algorithms, the parameters do not need to be tied across different layers, potentially giving better representational power. Furthermore, the latent variables can have different dimensions in different layers. Different types of layers such as fully connected layers can also be interleaved with the FGNN layers.

\subsection{FGNN for Max-Product Belief Propagation}
\label{sec:belief-propagation}
We motivated the construction of FGNN based on Low-rank Sum-Product LBP message updates. In this section, we prove that another widely used approximate inference
algorithm, Max-Product Belief Propagation can
be exactly parameterized by the FGNN 
using \emph{maximization} as $\AGG$ operator in Algorithm (\ref{algo:FGNN}). For convenience of description, we use a log-linear formulation of PGMs. 

Let $\xb$ be the set of all variables and let $\xb_c$ be the subset of
variables that factor $\phi_c$ depends on. Denote the set of indices of
variables in $\xb_c$ by $s(c)$. Then, an equivalent formulation of PGMs in equation (\ref{eq:pgm}) is as follows%
\begin{align}
  \label{eq:3}
  p(\xb) = \frac{1}{Z}\exp\bigg[\sum_{c\in\Ccal}\theta_c(\xb_c
  ) +\sum_{i\in \Vcal}\theta_i(x_i) \bigg],
\end{align}
where $\exp(\theta_c(\xb_c))= \phi_c(\xb_c)$ and $\exp(\theta_i(x_i))=\phi_i(x_i)$ are non-negative factor potential
functions (with $\theta_c(\cdot)$,
$\theta_i(\cdot)$ as the corresponding log-potentials) and
$Z$ is a normalizing constant. Using the log-linear formulation, the
MAP inference problem in \eqref{eq:map_inference} can be reformulated
as 
\begin{align}
  \label{eq:map_def}
  \xb^{\ast} = \argmax_{\xb}\sum_{c\in\Ccal}\theta_c(\xb_c
  ) +\sum_{i\in \Vcal}\theta_i(x_i),
\end{align}
and the  Max-Product loopy belief propagation in Eqns~\ref{eq:mp_lbp1}, \ref{eq:mp_lbp2}, \ref{eq:mp_lbp3} can
be reformulated as %
\begin{subequations}\label{max_product_factor}
  \begin{align}
    n_{i\rightarrow c}(x_i) =& \theta_i(x_i) +  \sum_{d: d\neq c, i \in
                               s(d)}m_{d\rightarrow i}(x_i),\\
    m_{c\rightarrow i}(x_i) =& \max_{\xb_c \setminus
                                \{x_i\}}\left[\theta_c(\xb_c) + \sum_{{j}\in
                                s(c), {j}\neq i}n_{{j}\rightarrow
                                c} (x_{{j}}) \right],\\
    b_i(x_i) = &  \theta_i(x_i) + \sum_{c: i \in s(c)}m_{c\rightarrow
  i}(x_i)
  \end{align}    \label{eq:max_product}
\end{subequations}%
We prove that Max-Product Belief Propagation can
be exactly parameterized by the FGNN. The proof of the propositions and lemmas are provided in the appendix. The sketch of the proof is as
follows. First, instead of parameterizing higher-order potentials as sum of rank-1 tensors, we show that the higher-order potentials can be also decomposed as maximization over a set of rank-1 tensors, and that the decomposition can be represented by a FGNN layer. After the decomposition, a single Max-Product iteration only requires two operations: (1) maximization over rows or columns of a matrix, and (2) summation over a group of features. We show that the two operations can be exactly parameterized by the FGNN and that $k$ Max-Product iterations can be simulated using $\Ocal(k)$ FGNN layers. %

Initially, we represented the higher-order potential functions as sum of rank-1 tensors which is suitable for Sum-Product type inference algorithms. %
However for Max-Product type
algorithms, a decomposition as a maximum of a finite number of rank-1
tensors is more appropriate. It has been shown that there is always a
finite decomposition of this type  \citep{kohli2010energy}. 

\begin{lemma}[\citep{kohli2010energy}]\label{lemma:decompose}
  Given an arbitrary potential function $\phi_c(\mathbf{x}_c)$, there
  exists a variable $z_c\in\Zcal_c=\{1, 2, \ldots, Z_c\}$ with $Z_c < \infty$  and a set of univariate potentials $\{\phi_{ic}(x_i,
  z_c)| i \in c\}$, s.t. %
  \begin{align}
    \log \phi_c(\mathbf{x}_c) = \log \max_{z_c\in \Zcal_c}\prod_{i\in s(c)}\phi_{ic}(x_i,
    z_c) = \max_{z_c \in \Zcal_c}\sum_{i\in s(c)}\varphi_{ic}(x_i,
    z_c),    \label{eq:lemma1}
  \end{align}
  where $\varphi_{ic}(x_i, z_c) = \log \phi_{ic}(x_i, z_c)$.
\end{lemma}
In Lemma \ref{lemma:decompose}, a higher-order potential are formulated as maximization over a set of rank-one tensors. It is notable that in worst cases, the size of set $\Zcal_c$, denoted by $Z_c$, is exponential against the order of the factor, but in practice higher-order potentials with low-rank properties can often be decomposed as \eqref{eq:lemma1} with relatively small $Z_c$.
Using ideas from \citep{kohli2010energy}, we show that a PGM %
can be converted into single layer FGNN with the non-unary potentials represented as a finite number of rank-1 tensors.

\begin{proposition} \label{propos:decomposition}
  A factor graph $\Gcal = (\Vcal,\Ccal,\Ecal)$ with variable log potentials $\theta_i(x_i)$ and 
  factor log potentials $\varphi_c(\xb_c)$ can be converted to a factor graph $\Gcal'$ with the same variable potentials and the decomposed  log-potentials $\varphi_{ic}(x_{i}, z_c)$ using a one-layer FGNN. 
\end{proposition}

With the
decomposed higher-order potential, one iteration of  the Max-Product \eqref{eq:max_product} can be rewritten using the following three steps \footnote{Detailed derivation are in appendix}:
\begin{subequations}\label{eq:decomposed}
  \begin{align}
      b_{c\rightarrow i}(z_c) \leftarrow & \sum_{{j}\in s(c),{j}\neq i}\max_{x_{{j}}}\left[\varphi_{{j}c}(x_{{j}}, z_c) - m_{c\rightarrow {j}}(x_{{j}}) + b_{{j}}(x_{{j}})\right],\label{eq:to_z} \\
      m_{c\rightarrow i }(x_i) \leftarrow & \max_{z_c}\left[b_{c\rightarrow i}(z_c) + \varphi_{ic}(x_i, z_c) \right], \label{eq:reparam_max_z}\\
      b_{i}(x_i) \leftarrow& \theta_i(x_i) +  \sum_{c: i \in s(c)} m_{c\rightarrow i} (x_i) %
    \label{eq:to_x}
  \end{align}
\end{subequations}
Given the log potentials represented as a set of rank-1 tensors at
each factor node, we show that each iteration of the Max-Product
message passing update can be represented by a Variable-to-Factor (VF)
layer and a Factor-to-Variable (FV) layer, forming a FGNN layer,
followed by a linear layer (that can be absorbed into the VF layer for the next iteration).

With decomposed log-potentials, Max-Product belief propagation mainly requires two operations: (1) maximization over rows or columns of a matrix; (2) summation over a group of features. We first show that the maximization operation in \eqref{eq:to_z} and \eqref{eq:reparam_max_z} (producing max-marginals) can be done using neural networks that can be implemented by the $\Mcal$ units in the VF layer.

\begin{proposition}\label{propos:matrix_max}
  For arbitrary feature
  matrix $\mathbf{X} \in \mathbb{R}^{k \times l}$ with $x_{ij}$ as its
  entry in the $i^{\text{th}}$ row and $j^{\text{th}}$ column, the feature mapping operation
  $
  \hat \xb = [\max_{j}x_{ij}]_{i=1}^l
  $
  can be exactly parameterized with a 2$\log_2 l$-layer neural network
  with at most $\Ocal(l^2 \log_2 l)$ parameters. 
\end{proposition}

Following the maximization operations, Eq. (\ref{eq:to_z}) requires
summation of a group of features. However, the VF layer uses max
instead of sum operators to aggregate features.
Assuming that the $\Mcal$ operator has performed
the maximization component of equation (\ref{eq:to_z}) producing
max-marginals, Proposition \ref{propos:feature_sum} shows how the
$\Qcal$ layer can be used to produce a matrix $\mathbf{W}$ that
converts the max-marginals into an intermediate form to be used with
the max aggregators. The output of the max aggregators can then be
transformed with a linear layer ($\mathbf{Q}$ in Proposition
\ref{propos:feature_sum}) to complete the computation of the summation
operation required in equation (\ref{eq:to_z}).
Hence, equation
(\ref{eq:to_z}) can be implemented using the VF layer together with a
linear layer that can be absorbed in the $\Mcal$ operator of the
following FV layer.

\begin{proposition}\label{propos:feature_sum}
  For arbitrary non-negative valued feature
  matrix $\mathbf{X} \in \mathbb{R}_{\geqslant 0}^{k \times l}$ with $x_{ij}$ as its
  entry in the $i^{\text{th}}$ row and $j^{\text{th}}$ column, there
  exists a constant tensor $\mathbf{W} \in
  \mathbb{R}^{k \times l \times kl}$ that can be used to transform $\mathbf{X}$ into an intermediate representation $y_{ir} = \sum_{ij}x_{ij}w_{ijr}$, such that after maximization operations are done to obtain $\hat y_r = \max_{i}y_{ir}$, we can use another constant matrix $\mathbf{Q}\in \mathbb{R}^{l
    \times kl }$ to obtain 
  $
  [\sum_ix_{ij}]_{j=1}^{l} = \Qb[\hat y_r]_{r=1}^{kl}.
  $
\end{proposition}
Eq. \eqref{eq:reparam_max_z} and (\ref{eq:to_x}) can be implemented in the same way as
(\ref{eq:to_z}) by the FV layer. First the max operations are
done by the $\Mcal$ units to obtain max-marginals. The max-marginals
are then transformed into an intermediate form using the $\Qcal$ units
which are further transformed by the max aggregators. An additional
linear layer is then sufficient to complete the summation operation
required in \eqref{eq:to_x}. The final linear layer can be
absorbed into the next FGNN layer, or as an additional linear layer in
the network in the case of the final Max-Product iteration.

Using the above two proposition, we can implement all important
operations \eqref{eq:decomposed}. 
Firstly, by Proposition \ref{propos:matrix_max}, we can construct the Variable-to-Factor module
using the following proposition. 
\begin{proposition}\label{propos:first_layer}
  The operation in \eqref{eq:to_z} can be parameterized by one MPNN layer
  with $\Ocal(|X| \max_{c\in\Ccal}|\Zcal_c|)$ parameters followed by a
  $\Ocal(\log_2|X|)$-layer neural network with at most $\Ocal(|X|^2\log_2|X|)$ hidden
  units. 
\end{proposition}
Meanwhile, by Proposition \ref{propos:matrix_max} and Proposition \ref{propos:feature_sum} the Factor-to-Variable module can be constructed using the
following proposition. 
\begin{proposition}\label{propos:second_layer}
  The operation in \eqref{eq:to_x} can be parameterized by one MPNN
  layer, where the  $\Qcal$ network is identity mapping and the
  $\Mcal$ network consists of a $\Ocal(\max_{c\in\Ccal}\log_2 |\Zcal_c|)$-layer neural network with at most $\Ocal(\max_{c\in\Ccal}|\Zcal_c|^2\log_2 |\Zcal_c|)$ parameters and a linear layer with $\Ocal(\max_{c\in\Ccal}|c|^2 |X|^2)$ parameters. 
\end{proposition}

Using the above two proposition, we achieves the main result of
this section as follows.

\begin{corollary}
  The Max-Product algorithm in \eqref{eq:max_product} can be exactly
  parameterized by the FGNN, where the number of parameters are
  polynomial w.r.t $|X|$, $\max_{c\in\Ccal}|\Zcal_c|$  and $\max_{c\in\Ccal}|c|$.
\end{corollary}

\section{Experiments}
\label{sec:experiment}
In this section, we evaluate our FGNN over a diverse set of tasks. First, we evaluate FGNN's performance for graphical model inference. We create synthetic PGMs in both low order and higher-order settings, and compare FGNN with other PGM inference algorithms. We also conduct experiments on low-density parity check (LDPC) decoding task (a typical PGM inference task). 

Next, we want to study how does FGNN perform on real-world data as against other state-of-the-art models. For this, we evaluate FGNN over three real-world problems where capturing higher-order dependencies is likely useful. We report experiments on the graph matching problem formulated as a PGM inference problem. We performed experiments on handwritten character recognition within words to demonstrate that FGNN is able to exploit sequence information. To validate the effectiveness of FGNN in capturing higher-order information from more general graph structured data, we report results on molecular datasets and compare with other state-of-the-art  $k$-order GNNs that capture higher-order information as well. Finally, we show FGNN is well suited for modeling human motion prediction task.

\subsection{MAP Inference over Synthetic PGMs}
\label{sec:synthetic-data}
We first evaluate FGNN on synthetically constructed datasets on graphical model inference tasks. As FGNN is based on inference algorithms, we test whether FGNN is able to outperform other prominent solvers on these inference problems on multiple datasets. First we describe experiments on chain-structured graphical model and then provide further results on other MRF structures.

\paragraph{Data}

We construct three synthetic datasets (D1, D2 and D3) for this experiment.
All models start with a length-30 chain structure with binary-states
nodes with node potentials randomly sampled from the uniform distribution  over the interval $[0,1]$, $\Ucal[0, 1]$,
and pairwise potentials that encourage two adjacent nodes to take state
$1$, \ie, it gives high value to configuration $(1,1)$ and low value to  other configurations.
In D1, the pairwise potentials are fixed, while in the others, they
are randomly generated. For D1, D2, and D3,  a budget higher-order
potential \citep{martins2015ad} is added at every node; these
potentials allow at most $k$ of the 8 variables  within their scope to
take the state $1$; specifically, we set $k = 5$ in D1 and D2 and set $k$ randomly in D3. %

In this paper, we use the simplest, but possibly most flexible method of defining factors in FGNN: we condition the factors on the input features. Specifically, for the problems in this section, all parameters that are not fixed are provided as input factor features.
We test the ability of the proposed model to find the MAP solutions, and compare the results with a well known graph neural network, MPNN \citep{gilmer2017neural} as well as several MAP inference solvers, namely AD3 \citep{martins2015ad} which solves a linear programming relaxation using subgradient methods, Max-Product Belief Propagation \citep{weiss2001optimality}, implemented by \citep{mooij2010libdai}, and a convergent version of Max-Product -- MPLP \citep{globerson2008fixing}, also based on a linear programming relaxation. The approximate inference algorithms are run with the correct models while the graph neural network models use learned models, trained with exact MAP solutions generated by a branch-and-bound solver that uses AD3 for bounding \citep{martins2015ad}\footnote{When obtaining the ground truth, we let the upper bound to touch the lower bound in the branch-and-bound solver to guarantee the optimality.}.
\begin{table}[t]
  \scriptsize
  \centering
  \setlength{\tabcolsep}{1.5pt}
  
  \caption{\small Results (percentage agreement with MAP and standard
    error) on synthetic datasets with  runtime in microseconds in
    bracket (exact followed by approximate inference runtime for
    AD3). A brief description of the three datasets is as follows. D1:
    random unary potentials + fixed parameter in pairwise and higher
    order potentials; D2: random parameter in unary and pairwise
    potentials + fixed parameter in higher-order potentials; D3:
    random parameter in all potentials.}
  \begin{tabular}{lccccccc}
    \toprule
    & \scriptsize AD3  &
                         {\scriptsize
                         Max-Product} & {\scriptsize MPLP} &
                                                             {\scriptsize
                                                             MPNN}&
                                                                    {\scriptsize Ours}\\
    \midrule
    D1  & 80.7{\scriptsize$\pm$0.0014} (5 / 5)  &57.3{\scriptsize$\pm$0.0020} (6) & 65.8{\scriptsize$\pm$0.0071} (57) & 71.9{\scriptsize$\pm$0.0009} (131) &\textbf{92.5}{\scriptsize$\pm$0.0012} (144) \\
    D2  & 83.8{\scriptsize$\pm$0.0014} (532 / 325) &50.5{\scriptsize$\pm$0.0053} (1228) &68.5{\scriptsize$\pm$0.0074} (55) & 74.3{\scriptsize$\pm$0.0009} (131) &\textbf{89.1}{\scriptsize$\pm$0.0010} (341)\\
    D3  & 88.1{\scriptsize$\pm$0.0006} (91092 / 1059) &53.5{\scriptsize$\pm$0.0081} (4041) &64.2{\scriptsize$\pm$0.0056} (55) &82.1{\scriptsize$\pm$0.0008} (121) &\textbf{93.2}{\scriptsize$\pm$0.0006} (382)\\
    \bottomrule
  \end{tabular}
  
  \label{tab:res_syn}
\end{table}

\paragraph{Architecture and training details}
In this task, we use a factor graph neural network consisting of 8
FGNN layers (see the detail in the Appendix). The model is implemented in pytorch  \citep{paszke2017automatic},
and trained with Adam optimizer~\citep{kingma2014adam} with initial learning rate
$\text{lr}=3\times 10^{-3}$. After each epoch, %
lr
is decreased by multiplying a factor of $0.98$. 
All the models in Table \ref{tab:res_syn} were trained for $50$ epochs after which all of them achieve convergence.

\paragraph{Results} The percentage of agreement with MAP solutions is
shown in Table
\ref{tab:res_syn}. Our model achieves far better results on D1, D2, and
D3 than all others. D4 consists of chain models, where Max-Product works
optimally\footnote{Additional experiment on trees, where Max-Product also works optimally
  can be found in Appendix \ref{sec:exper-tree-struct}
  along with details on all experiments.}. The linear programming
relaxations also perform well. In this case, our method is able to
learn a near-optimal inference algorithm on the chain case as well.

Traditional methods including Max-Product, MPLP perform poorly on D1,
D2 and D3. In these, even though FGNN can emulate the traditional Max-Product, it is
better to learn a different inference algorithm. AD3 have better
performance than others, but is worse than our FGNN. 
The accuracy of FGNN is noticeably higher than that of MPNN as MPNN
does not use the higher-order structural priors that are captured by
FGNN.

\subsubsection{Ablation studies}
In order to study the behaviour of the FGNN model under varying inputs, we conducted the following additional experiments as part of the ablation study using synthetic data.

\paragraph{Effect of wrong graph structures} First we did a small ablation study by modifying the graph structure
inputed to FGNN using the Dataset D1 and D2. Originally D1 and D2 provides chain
structured PGM, with budget higher-order factor formed over every 8
neighbor variables. In this experiment we augment the graph structure by using 4 and 6 variables to form a higher-order factor instead of the correct 8 variables. On D1, the accuracies are 81.7 and 89.9 when 4 and 6 variables are used in place of the correct 8 variables; on  D2, the accuracies are 50.7 and 88.9 respectively. In both cases, the highest accuracies are achieved when the sizes of the HOPs are set correctly.
\paragraph{Generalization to novel graph structures}
In order to further evaluate the generalization of FGNN, we conducted an additional experiment to train the FGNN on fixed length-30
MRFs using the same protocol as Dataset3, and test the algorithm on
60000 random generated chain MRF whose length ranges from 15 to 45
(the potentials are generated using the same protocol as
D3, where all unary, pairwise and higher-order potentials have random parameters). The result is in Table \ref{table:res_var_length}, which
shows that the model trained on fixed size MRF can be generalized to
MRF with different graph structures.
\begin{table}[h]
	\scriptsize
	\centering
	\begin{tabular}[t]{ccc}
		\toprule
		Chain length & AD3 & FGNN \\
		\midrule
		(15, 25) & $88.95$ & $94.31$ \\
		(25, 35) & $88.18$ & $93.64$\\
		(35, 45) & $87.98$ & $91.50$\\
		\bottomrule  
	\end{tabular}
	\vspace{0.25em}
	\captionof{table}{\footnotesize Generalization ability
          (measured by percentage agreement with MAP) of
          algorithms on PGMs with different graph structures. We train our
        FGNN model on the training set of D3 where structure of PGMs
        is fixed and test it on higher-order PGMs with different
        chain-length.  }\label{table:res_var_length}
\end{table}

\paragraph{Effect on low order PGMs such as chains and trees}
\label{sec:exper-tree-struct}

In addition to the higher-order PGM above,
we conducted an additional experiment on chain and tree structured PGMs. For
chain structured PGMs, we use the same protocol as D3 to generate
training and testing data, but with higher-order factors removed. For
tree structured dataset, 
the
training set includes 90000 different PGMs as randomly
generated binary trees whose depths are between 3 and 6. Each node is
associated with a random variable $x_i \in \{0, 1\}$ along with a
log potential $\theta_i(x_i)$ randomly sampled from Gaussian
distribution $\Ncal(0, 1)$. Each edge $(i, j)$ in a tree is
associated with a pairwise log potential $\theta_{ij}(x_i, x_j)$ which
is randomly sampled from Gaussian distribution $\Ncal(0,
1)$. There is also 10000 testing instances which are generated in the
same way as the training set. The experiment result is shown in Table  \ref{tab:exp_tree_pgm}.

\begin{table}[h]
	\scriptsize
	\centering
	\begin{tabular}[t]{lcccccc}
		\toprule
		&AD3&Max-Product&MPLP&MPNN&FGNN
		\\
          \midrule
          \thead{Chain}  & {100}& {100}  & 99.9 & 91.2  &98.0 \\
          \thead{Tree} & 100.0 & 100.0 & 99.97& -- & 98.35
		\\
		\bottomrule
	\end{tabular}
	\caption{Results (percentage agreement with MAP) on chain and tree structured PGMs.}
	\label{tab:exp_tree_pgm}
      \end{table}
      For chain PGMs, our algorithm achieves comparable
      results as Max-Product, which is known to be optimal on chain
      PGMs, and outperforms pairwise MPNN. 
For a tree structured PGM, it is not as easy to shrink the pairwise features to the nodes as an adaptation for MPNN as in the case of chain PGM in Section~\ref{sec:synthetic-data}, so we omit the experiment on MPNN. Still, our Factor Graph Neural Network achieves a good performance even when compared with Max-Product which is optimal on tree PGMs and also with the linear programming relaxations.

\subsection{LDPC Decoding (MAP or Marginal Inference)}
\begin{figure}[t]
  \centering
  \includegraphics[width=\textwidth]{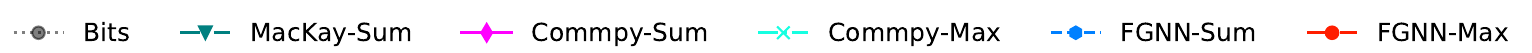}\\
  \subfloat[$\sigma_b=0$]{\includegraphics[width=0.3\textwidth]{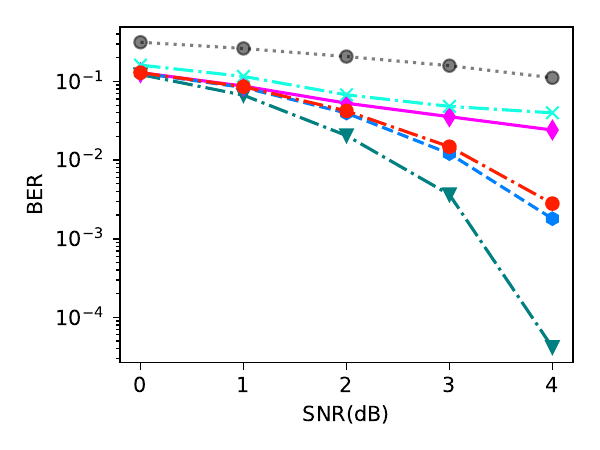}}\hfill
  \subfloat[$\sigma_b=1$]{\includegraphics[width=0.3\textwidth]{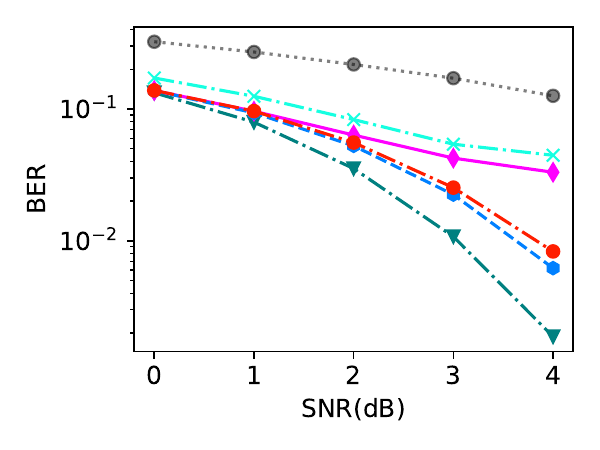}}\hfill
  \subfloat[$\sigma_b=2$]{\includegraphics[width=0.3\textwidth]{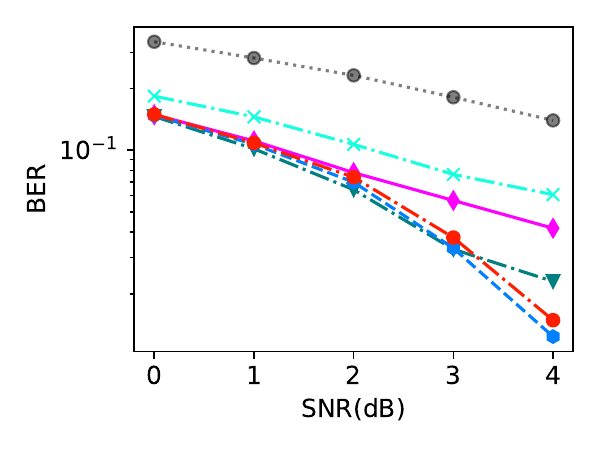}}\\
  
  \subfloat[$\sigma_b=3$]{\includegraphics[width=0.3\textwidth]{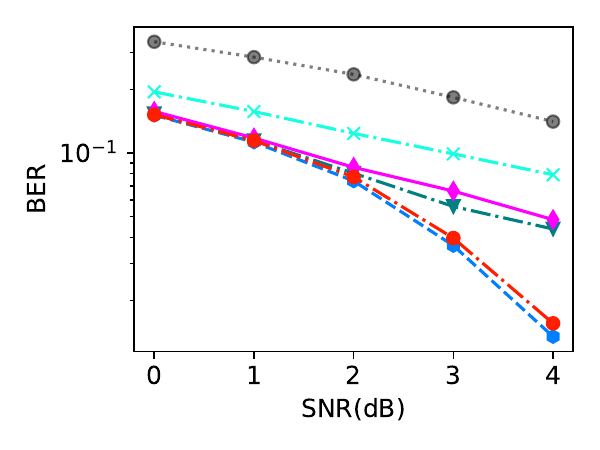}}\hfill
  \subfloat[$\sigma_b=4$]{\includegraphics[width=0.3\textwidth]{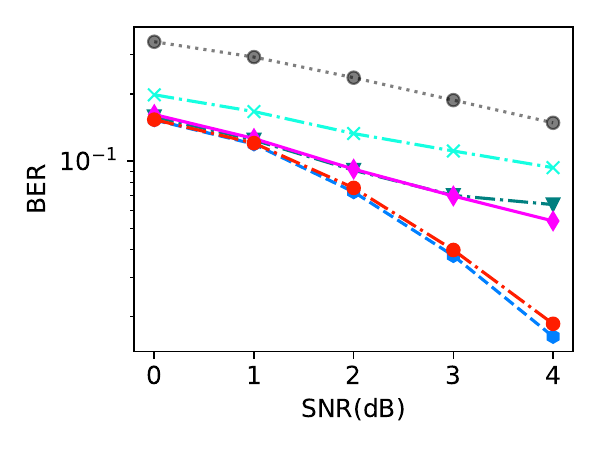}}\hfill
  \subfloat[$\sigma_b=5$]{\includegraphics[width=0.3\textwidth]{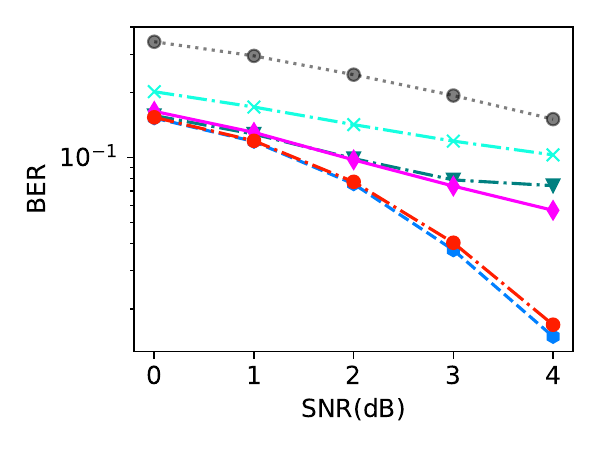}}
  \captionof{figure}{\small Experimental results (Bit Error Rate, BER
    v.s. Signal-to-Noise Ratio, SNR) on LDPC decoding. }
  \label{fig:ldpc_decoding}
\end{figure}

The low-density parity check (LDPC) codes are widely used in wired and wireless communication, where the decoding can be
done by Sum- or Max-Product belief propagation
\citep{zarkeshvari2002implementation}.

\textbf{Data}~~Let $\xb$ be the 48-bit original signal, and $\yb$ be the
96-bit LDPC encoded signal by encoding scheme 
``96.3.963''\citep{mackay2009david}. Then a noisy signal
$\tilde \yb$ is obtained by transferring $\yb$ through a channel with
white Gaussian and burst noise, that is, for each bit $i$, 
$
\tilde y_i = y_i + n_i + p_iz_i,
$
where $n_i \sim \Ncal(0, \sigma^2)$ , $z_i \sim \Ncal(0, \sigma_b^2)$,
and $p_i$ is a Bernoulli random variable \emph{s.t.}
$P(p_i = 1) = \eta; ~P(p_i = 0)= 1-\eta$. In the experiment, we
set $\eta=0.05$ as \citep{kim2018communication} to simulate unexpected
burst noise during transmission. By tuning $\sigma$, we can
get  different signal with $\SNR_{dB} = 20 \log_{10}(1/\sigma) $.

In the experiment, for all learning-based methods, we generate
$\tilde \yb$ from randomly sampled $\xb$ on the fly with
$\text{SNR}_{dB}\in \{0, 1, 2, 3, 4\}$ and $\sigma_b \in
\{0,1,2,3,4,5\}$. For each learning-based method, $10^8$ samples are
generated for training. Meanwhile, for each different $\SNR_{dB}$ and
$\sigma_b$, 1000 samples are generated for validating the performance
of the trained model.

In LDPC decoding, the $\SNR_{dB}$ is usually assumed to be known
and fixed, and the burst noise is often unexpected and its
parameters are unknown to the decoder. Thus, for learning based methods
and traditional LDPC decoding method, the noisy signal $\tilde\yb$ and
the $\SNR_{dB}$ are provided as input. In our experiments, since the
LDPC decoding can be done by both max-product and sum-product, we
train our FGNN with both max and sum aggregation function (see
FGNN-Max and FGNN-Sum in Figure \ref{fig:ldpc_decoding}.). The FGNN is
compared with baselines including two implementation of Sum-Product
(Mackey-Sum \citep{mackay2009david} and Commpy-Sum
\citep{taranalli2015commpy}), the max-product decoder from Commpy
(Commpy-Max \citep{taranalli2015commpy}) and the bit decoding
baseline. 

\textbf{Architecture and training details}~~  In this task, we use a factor graph neural network consisting of 8
FGNN layers (the details are
provided in the Appendix). The model is implemented by using pytorch  \citep{paszke2017automatic}, and
trained with Adam optimizer~\citep{kingma2014adam} with initial learning rate
$\text{lr}=1\times 10^{-2}$. After every 10000 samples, 
lr
is decreased by multiplying a factor of $0.98$. After training over $10^8$
samples, the training loss converges.%

\textbf{Results}~~
We compare FGNN with three public available LDPC decoder
MacKay-Sum \citep{mackay2009david}, Commpy-Sum
\citep{taranalli2015commpy} and Commpy-Max
\citep{taranalli2015commpy}. The first two decoders are using Sum-Product belief
propagation to propagate information between higher-order factors and nodes, but with different belief clipping strategy and
different belief propagation scheduler.  The third decoder are using
Max-Product for as inference algorithm. 
Meanwhile, our FGNN uses a  learned 
factor-variable information propagation scheme, and the other learning
based method, MPNN ignores
the higher-order dependencies.
The decoding accuracy is provided in Figure
\ref{fig:ldpc_decoding}. The Sum-Product based methods (MacKay-Sum and
Commpy-Sum) are known
to be near optimal for Gaussian noise, however the hyper-parameter of
Sum-Product are sensitive. Due to different hyper-parameters,
MacKay-Sum get the best performance when the burst noise level is low
while Commpy-Sum have superior performance than MacKay-Sum with high
burst noise level. Our FGNN (both FGNN-Max and FGNN-Sum) always performs better than the
Commpy-Sum and Commpy-Max, it achieves
comparable but lower performance than the MacKay-Sum 
for low burst noise level($\sigma_b\in [0,2]$), and outperforms all
other methods for high burst noise levels ($\sigma_b\in [3,5]$), and
the results from ``sum'' aggregation function are slightly better than
the results from ``max'' aggregation function.

\subsection{The Graph Matching Problem (MAP Inference)}
The Graph Matching is a fundamental problem by itself and a key step in various
computer vision topics including image registration, tracking, motion
analysis and more. Traditionally, the graph matching problems are often modelled as Quadratic Assignment Problems (QAPs) and they can be also viewed as MAP inference problems over factor graphs \citep{zhang2016pairwise,swoboda2017study}. These problems can be solved by using belief propagations in PGMs \citep{zhang2016pairwise, swoboda2017study},  or using graph neural networks \citep{Wang_2019_ICCV,zhang2019deep}. In this section, we apply our FGNN to graph matching problems and compare with both traditional  methods as well as recent graph neural network based approaches. %
\paragraph{The Problem and Traditional Model} Let $\Pcal=\{\pb_i|i\in[n]\}$,
$\Qcal=\{\qb_i|i\in[n]\}$ be two sets of feature points, where $[n]$
denotes the set $\{1,2,\ldots, n\}$.  The graph matching problem can be formulated as
an MAP inference problem as follows
\begin{align}
  \label{eq:MAP_matching}
  \argmax_{\Xb}\sum_{i,j=1}^{n} x_{ij}S_{\text{n}}(\pb_i, \pb_j)
  + \sum_{i,j,k,l=1}^{n}x_{ik}x_{jl}S_{\text{e}}(\eb_{ij}^{\Pcal}, \eb_{kl}^{\Qcal}) + \phi(\Xb),
\end{align}
where $S_{\text{n}}$ and $S_{\text{e}}$ are user-specified similarity
functions of features, and $\eb_{ij}^{\Pcal}$ and $\eb_{kl}^{\Qcal}$
are edge features extracted or learned from $\Pcal$ and $\Qcal$,
respectively. The variable $\Xb$ is a $n\times n$ matrix with
$x_{ij}$ as its 
entry at $i^{\text{th}}$ row and $j^{\text{th}}$ column.  The higher-order log-potential $\phi(\Xb)$ is used to
enforce the one-to-one constraints as follows,
\begin{align}
  \label{eq:one-to-one}
  \phi(\Xb) = \left \{ \begin{matrix}
      0, & \sum_{i=1}^{n}x_{ij} = 1, \sum_{j=1}^{n}x_{ij}=1, x_{ij}
      \geqslant 0\\
      -\infty, & \text{otherwise}. 
    \end{matrix}
  \right .
\end{align}
Traditionally, both the features and the similarity functions are
handcrafted
\citep{wang2018graph,swoboda2017study,DBLP:journals/ijcv/LiuQYH14,zhou2012factorized,zhang2016pairwise}.
Recently, many approaches were proposed
\citep{wang2021neural,xu2021deep,rolinek2020deep,Fey/etal/2020,wang2020learning,yu2019learning,Wang_2019_ICCV}
to replace the handcrafted feature with learned feature with very promising results.

\begin{figure}
    \centering
    \includegraphics[width=0.4\linewidth]{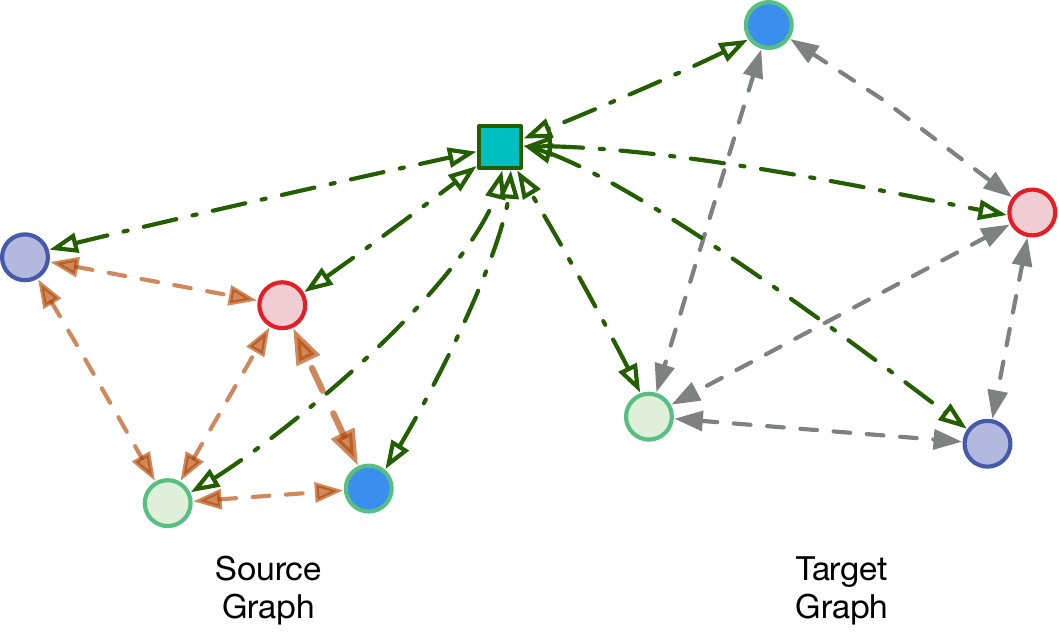}
    \caption{\small Factor Graph Neural Networks for Graph Matching Problems. The pairwise message passing are used to produce better node features so that without the factor it can still work as a metric-learning based methods. With the factor, the performance of matching can be significantly improved as shown in Table \ref{table:matching_res}.}
    \label{fig:FGNN_matching}
\end{figure}

\paragraph{The FGNN Model} In the traditional formulation of graph matching problem \eqref{eq:MAP_matching}, there are $n^2$ binary variables. We instead use a more compact but equivalent formulation where there are $2n$ variables with $n$ states. We formulate the graph matching problem as 
\begin{align}
    \label{eq:fgnn_factor_graph_model}
    \argmax_{\yb, \zb} = \sum_{i=1}^{n}\theta_i(y_i|\Pcal,\Qcal) + \sum_{j=1}^{n}\vartheta_j(z_j|\Pcal,\Qcal) + \varphi(\yb,\zb|\Pcal,\Qcal),
\end{align}
where $y_i = j$ indicates that the $i^{\text{th}}$ node in source graph corresponding to the $j^{th}$ node in the target graph. Similarly, $z_j=i$ indicates that the the $i^{\text{th}}$ node in source graph corresponds the $j^{th}$ node in the target graph. The higher-order term  can enforce that $\yb$ and $\zb$ being consistency, and can absorb the pairwise terms in \eqref{eq:MAP_matching}. As a result, we can extend the graph neural network in \citet{zhang2019deep} to handle higher-order terms more efficiently by using the proposed factor graph neural network shown in Figure \ref{fig:FGNN_matching}, where each node in source graph corresponds to random variable $\yb_i$ and each node in target graph corresponds to random variable $\zb_i$. The pairwise message passing procedure inside the FGNN is only used to produce better node features as \citet{zhang2019deep}. Without factors, the model can still work as a metric-learning based method but the factors can significantly improve the performance of GNN model.

\paragraph{Data} Following the experimental settings of
\citep{Wang_2019_ICCV,yu2019learning,Fey/etal/2020,rolinek2020deep,xu2021deep},
we use the Pascal VOC dataset \citep{everingham2010pascal} with the
keypoints annotated by \citet{bourdev2009poselets} to evaluate the
performance of handcrafted and learning-based graph matching
algorithms. The dataset contains 20 classes of
instances (objects) with manually labeled keypoint locations, and the
instances may vary in scale, view angle and illumination. For each
instance, the number of inliers ranges from 6 to 23. We applied the
same filtering and training/testing split
\citep{Wang_2019_ICCV,yu2019learning,Fey/etal/2020,rolinek2020deep,xu2021deep},
where 7020 annotated images are used for
training and 1682 for testing.

\paragraph{Architecture and training details} In our experiment, we
use the same training protocol as in
\citep{Wang_2019_ICCV,yu2019learning,Fey/etal/2020,rolinek2020deep,xu2021deep},
where the input of the models are two sets of coordinates of key-points
and two images. In our model, as in previous work the two images are
first been fed to the VGG19 \citep{simonyan2014very} net to form
visual features. By using bilinear interpolating as previous work
\citep{Fey/etal/2020}, we can get the visual feature vector of every
node from the outputs of VGG. For each set of key-points, each key-point is connected to its
$k$-nearest neighbors with $k=6$. For each node, the geometric and
visual features are concatenated to form the node feature. For each
edge, the difference of geometric node features of the two nodes connected by the edge serves as the edge
feature. Furthermore, the initial factor features are generated as
follows
\begin{align}
  \label{eq:4}
  \fb = \max_{i\in f}\vb_i,
\end{align}
where $\vb_i$ is the node feature associated with node $i$, and
$\max$ is the entrywise maximization. The above node, edge and factor
features will be fed into a FGNN network composed by three
blocks. In each block, there will be a pairwise message passing module
as \citep{zhang2019deep}, and one factor message passing module as
\ref{algo:FGNN}. Then the FGNN network, along with the VGG network,
will be trained with Adam \citep{kingma2014adam} optimizer with
learning rate $10^{-4}$ ($10^{-6}$ for the VGG part). Our algorithm
has been trained for 200 epochs and in each epoch we random samples
16000 image pairs from the training set.

\paragraph{Results} Our FGNN based algorithm is compared with a bunch
of traditional handcrafted graph matching algorithms, as well as
several state-of-the-arts learned graph matching algorithms. The
results are shown in Table \ref{table:matching_res}, where the results
of handcrafted graph matching methods are from
\citet{wang2020learning}, and the results of learning-based approaches
are from their paper except for MPNN \citep{zhang2019deep}. For the
results of MPNN, their network is identical to ours but with the
factor message passing module removed, and we train the network using
exactly the same protocol as ours.

\begin{table}[t]
  \centering
   \Large
  \resizebox{\textwidth}{!}{ 	\renewcommand{\arraystretch}{1.05}	
  \begin{tabular}{c|cccccccccccccccccccc|c}
    \toprule
     Method & aero & bike &bird & boat& btl& bus& car& cat& chair&
                                                                        cow& table&dog&horse&mbk&prn&plant&shp&sofa&trn&tv&avg\\
    \midrule
    IPFP \citep{leordeanu2009integer}&25.1&26.4&41.4&50.3&43.0&32.9&37.3&32.5&33.6&28.2&26.9&26.1&29.9&32.0&28.8&62.9&28.2&45.0&69.3&33.8&36.6\\
    RRWM \citep{cho2010reweighted}&30.9&40.0&46.4&54.1&52.3&35.6&47.4&37.3&36.3&34.1&28.8&35.0&39.1&36.2&39.5&67.8&38.6&49.4&70.5&41.3&43.0\\
    PSM  \citep{egozi2012probabilistic}&32.6&37.5&49.9&53.2&47.8&34.6&50.1&35.5&37.2&36.3&23.1&32.7&42.4&37.1&38.5&62.3&41.7&54.3&72.6&40.8&43.1\\
    GNCCP \citep{liu2013gnccp}&28.9&37.1&46.2&53.1&48.0&36.3&45.5&34.7&36.3&34.2&25.2&35.3&39.8&39.6&40.7&61.9&37.4&50.5&67.0&34.8&41.6\\
    ABPF \citep{wang2018graph}&30.9&40.4&47.3&54.5&50.8&35.1&46.7&36.3&40.9&38.9&16.3&34.8&39.8&39.6&39.3&63.2&37.9&50.2&70.5&41.3&42.7\\
    \midrule
    GMN \citep{zanfir2018deep} & 31.9& 47.2& 51.9& 40.8& 68.7& 72.2& 53.6& 52.8& 34.6& 48.6& 72.3& 47.7& 54.8& 51.0& 38.6& 75.1& 49.5& 45.0& 83.0& 86.3& 55.3\\
    LCS \citep{wang2020learning}&                                   {46.9}&{{58.0}}&63.6&{{69.9}}&{{87.8}}&{79.8}&{71.8}&60.3&{{44.8}}&{64.3}&{79.4}&57.5&64.4&57.6&{{52.4}}&{{96.1}}&62.9&{{65.8}}&94.4&{{92.0}}&68.5\\
    HNN-HM\citep{liao2021hypergraph} & 39.6& 55.7& 60.7& 76.4& 87.3&
                                                                     86.2& 77.6& 54.2& 50.0& 60.7& 78.8& 51.2& 55.8& 60.2& 52.5& 96.5& 58.7& 68.4& 96.2& 92.8& 68.0\\
    PCA-GM \citep{Wang_2019_ICCV} &40.9&55.0&{{65.8}}&47.9&76.9&77.9&63.5&{67.4}&33.7&65.5&63.6&{61.3}&{68.9}&{{62.8}}&44.9&77.5&67.4&57.5&86.7&{90.9}&63.8\\
    CIE-H \citep{yu2019learning} & 51.2& 69.2& 70.1& 55.0& 82.8& 72.8& 69.0 &74.2& 39.6& 68.8& 71.8& 70.0& 71.8& 66.8& 44.8& 85.2& 69.9& 65.4& 85.2& 92.4& 68.9\\
    DGMC \citep{Fey/etal/2020} & 47.0& 65.7& 56.8& 67.6& 86.9& 87.7
                                                &85.3 &72.6& 42.9&
                                                                   69.1& 84.5& 63.8& 78.1& 55.6& 58.4& 98.0& 68.4& 92.2& 94.5& 85.5& 73.0\\
        BBGM \citep{rolinek2020deep} & 61.5 &75.0& 78.1& 80.0& 87.4& 93.0& 89.1& 80.2& 58.1& 77.6&76.5&79.3&78.6& 78.8& 66.7& 97.4& 76.4& 77.5& 97.7& 94.4& 80.1\\
    NGMv2 \citep{wang2021neural} & 61.8&  71.2& 77.6& 78.8& 87.3& 93.6& 87.7& 79.8& 55.4& 77.8& 89.5& 78.8& 80.1& 79.2& 62.6& 97.7& 77.7& 75.7& 96.7& 93.2& 80.1\\
    NHGMv2\citep{wang2021neural} & 59.9&  71.5& 77.2& 79.0& 87.7& 94.6& 89.0& 81.8& 60.0& 81.3&
                                                                                                     87.0& 78.1& 76.5& 77.5& 64.4& 98.7& 77.8& 75.4& 97.9& 92.8& \textbf{80.4}\\
    MPNN \citep{zhang2019deep} & 57.8&69.1&74.4&77.7&89.2&90.4&90.4&77.4&73.1&81.9&90.4&76.5&78.6&76.5&54.4&97.9&78.2&70.0&97.3&94.9&79.8\\
    Ours &57.3&69.0&75.9&78.3&93.8&91.6&90.8&76.9&73.9&82.5&89.9&77.0&79.9&77.8&54.4&98.2&78.2&74.9&97.7&94.7&\underline{\textbf{80.6}}\\
    \bottomrule
  \end{tabular}
  }
  \caption{Accuracy on the Pascal VOC Keypoint dataset. \textbf{Top}: Results of traditional hand-crafted solver. \textbf{Bottom:}  Result from methods using learned feature. All the learning based approaches are using VGG19 \citep{simonyan2014very} as backbone to extract visual feature, but the graph neural network architectures are different. }\label{table:matching_res}
\end{table}

Our algorithm outperforms the previous methods because our factor
message passing module can handle higher-order information
better. Particularly the performance of previous pairwise network based
state-of-the-arts methods NGMv2 \citep{wang2021neural}  can be
improved by introducing higher-order terms to form NHGMv2
\citep{wang2021neural} to get an improvement of $0.3\%$.
Meanwhile, compare to our pairwise counter-part MPNN
\citep{zhang2019deep}, we got a performance improvement of $0.8\%$ and
our average performance outperforms all previous methods. 

\subsection{Handwritten character sequence recognition (Marginal Inference)}\label{sec:handwriting-char-recog}
In this experiment, we explore how useful higher-order modeling with FGNN is for structured prediction tasks with sequence data. A sequence is one of the simplest graph structures where nodes are connected in the form of a linear chain. FGNN should be able to capture strong higher-order dependencies in such datasets. Therefore, we explore the effect of order of factors on the task of handwritten character recognition. 

\textbf{Modeling}~~In handwritten character recognition sequence, the adjacent few characters of a node are likely to contain useful information for predicting a character. We consider one of the simplest higher-order model: a $k-$order factor for each node in the sequence. Formally,
let $x = [x_1,\dots,x_{|x|}]$ denote an input sequence of length $|x|$ with each $x_i$ associated with a label $y_i \in \{a,b, \dots,z\}$. 
If $x_{i,k} = [x_{i-k},\dots ,x_{i+k}]$ is an order-$k$ segment of $x$ centered at $x_i$ and $y_{i,k}$ is the corresponding label segment of $x_{i,k}$, we define $f_i(y_{i,k}|x_{i,k}:\theta)$ as the $k-$order factor encoding dependencies in $y_{i,k}$. This gives us a conditional random field with unnormalized probability as $P(y|x) = \prod_{i=1}^{|x|}f_i(y_{i,k}|x_{i,k};\theta)$. 
This corresponds to a factor $f_i$ for every node $i$ connecting its adjacent $k$ nodes in the sequence. %

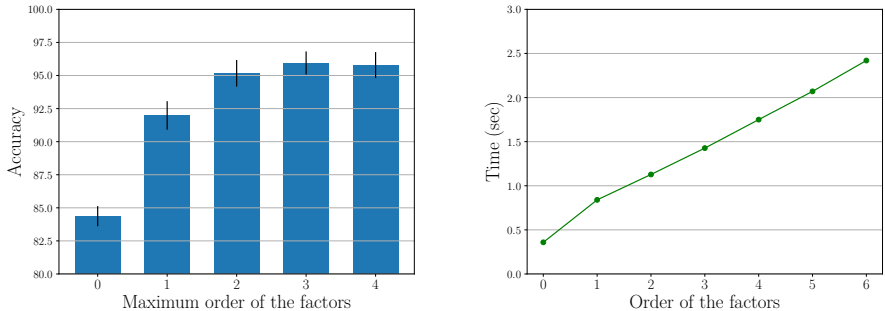
\begin{figure}%
  \centering
  \begin{minipage}{0.4\textwidth}
    \centering
    \scalebox{0.3}{\input{./figure/ocr_a.pgf}}
  \end{minipage}
  \begin{minipage}{0.4\textwidth}
    \centering
    \scalebox{0.3}{\input{./figure/ocr_b.pgf}}
  \end{minipage}
  \caption{Handwriting character Recognition}\label{fig:graph}	
\end{figure}
\textbf{Data}~~We study the properties of FGNN network with the handwriting recognition data set from~\citep{taskar2004max}, originally collected by ~\citep{kassel1995comparison}. 
The dataset consists of a subset of $\sim 6100$ handwritten words with the average length of $\sim 8$ characters. The words are segmented into characters and each character is rasterized into an image of 16 by 8 binary pixels. The dataset is available in 10 folds with each fold containing $\sim 6000$ characters. The task is to recognise characters by leveraging the neighbourhood of characters within the context of words. Since the words come from a small vocabulary, there is a strong higher-order correlation present in the dataset. In our framework, depending on the order, each character node $x_i$ can share a factor with other character nodes $x_j$ within the same word. We follow a 10-fold cross-validation setup with 1 fold for training and 9 folds for testing and report averaged results. We evaluate the performance of FGNN by varying the order and rank of the factors.

\textbf{Architecture and training details}~~ We use 3 standard convolution and a fully connected layer as the base feature extractor to get the zero-th order feature of 512 dimensions. We then run 3 iterations of higher-order message passing followed by a classifier. We fix the rank of all factors at 1024 and share parameters between factors with the same order. We train for 50 epochs with a learning rate of 0.001.

\textbf{Results}~~In this experiment we study the behaviour of the model in terms of accuracy and training time when order and rank of the factors are varied. Results as shown in Figure~(\ref{fig:graph}) suggest the FGNN is able to capture higher-order information. The model shows strong improvements as maximum order of factors used is successively increased before saturating at 4th-order and above. To evaluate the efficiency of FGNN with higher-order factors, we analysed the computation time as we vary the order of factors used. Figure~(\ref{fig:graph}) shows that the training time per epoch grows almost linearly with the order of factors.

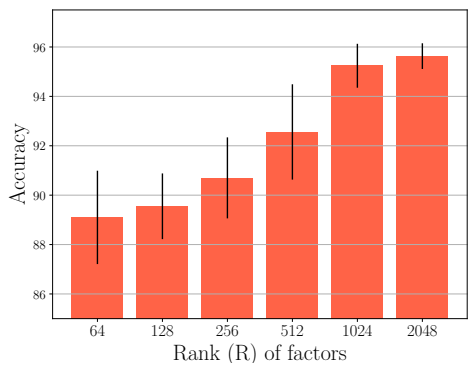
\begin{figure}[h]%
  \centering
  \begin{minipage}{0.4\textwidth}
    \centering
    \scalebox{0.35}{\input{./figure/ocr_c.pgf}}
  \end{minipage}
  \caption{Effect of rank of factor parameters for Handwriting
    character Recognition. The accuracy of the model increases with increasing rank and saturates after rank-$1024$,  whereas a full rank 3rd order factor would need at least $17576$ components for exact representation. %
    }\label{fig:ocr_rank_fig}
\end{figure}

\paragraph{Effect of Rank of factor parameters} One important hyperparameter of the FGNN model is the rank of the higher-order factors which is given by the out dimension of transformation matrix $\mathbf{W}$. Since, the effect of higher-order context is clear from the results, we analyse the sensitivity of the model performance with the rank of the higher-order factors. To evaluate the effect of varying rank, we fixed the zero-th order feature dimension to 64 and used factors up to order-3. We then ran three message passing iterations by varying the rank of the factors from 64 to 2048. Results in Figure~(\ref{fig:ocr_rank_fig}) show consistent improvements in performance with the increasing rank before saturating at 1024 and above.
Note that for characters with the class size of $26$ for each character, a full-rank tensor representation for order-$3$ potential function needs at least $26^3=17576$ components. Clearly, $1024 << 26^3$ shows that the underlying high-order potential is well approximated by the low-rank representation.

\subsection{Molecular data (Graph Regression)}\label{sec:molecular_data}
In the following set of experiments, we use FGNN purely as graph neural network operating on graph structured data. Note that FGNN only provides inductive bias of graphical model inference algorithms but is free to learn a richer algorithm  suitable for the task. With the molecular data, we show that FGNN can work purely as a graph neural network on variable sized graphs and provides additional flexibility in modeling typed nodes and edges.

Molecular data has several properties needed for an effective study of higher-order representation learning models of graphs. A molecule can be represented as a graph with atoms as nodes and bonds that exist between atoms as edges. Higher-order information is present in the form of valence constraints of atoms which determine the number of bonds they can form. In addition, molecules often contain subgraphs, also called functional groups (\eg, OH, CHO etc.), which are identifiable and significant in determining their functionality and geometry. %
Any relational learning model should be powerful enough to capture such higher-order structures and dependencies in order to learn highly discriminative representations. %
Furthermore, molecules come with varying number of nodes and hence learning higher-order functions with shareable parameters is necessary. This makes FGNN suitable for statistical learning in such datasets. We now focus on molecular data to study the effectiveness of the proposed model and show its modeling flexibility in incorporating domain knowledge in constructing the factors. 

\textbf{Modeling}~~In molecular data, there is an input graph with node and edge features. To use FGNN, we need to define a suitable factor graph model based on the graph structure. In FGNN,  $\Qcal(\tb_{ci})$ %
is conditioned on edge features and hence separate for all variables connected to the factor. %
This gives much freedom in modeling to leverage domain knowledge in order to share some of the factor parameters and be able to work with large graphs with typed edges as well. Given a molecular graph, we discuss possible ways of constructing higher-order factors, conditioning and sharing of parameters. 

One way to capture higher-order information is to add a factor for every node in the molecule connecting that node (we will call this node \textit{central atom} in that factor) and its neighbours to the factor. Then weights of the potential for the factor can be shared by conditioning on the  following: 
\begin{itemize}[{leftmargin=*}]
\item \textbf{Central atom type (CAT):} Weights within the factor are shared but different factors share parameters only if they have the same central atom type. %
\item \textbf{Bond type (BT):} Weights are shared if the bond type between the central atom and its neighbour is same. %
\item \textbf{Central atom and bond type (CABT):} Weights are shared if both the central atom type and bond type are same. %
\item \textbf{Central atom, bond and neighbour atom type (CABTA):} Weights are shared if the bond type and the atom types of atoms sharing the bond are same. 
\end{itemize}
Do note that most molecular datasets have small number of atom types and bond types. %
This shows the proposed model of message passing is flexible and provides sufficient freedom in modeling.%

\begin{figure}%
  \centering
  \includegraphics[width=0.65\linewidth]{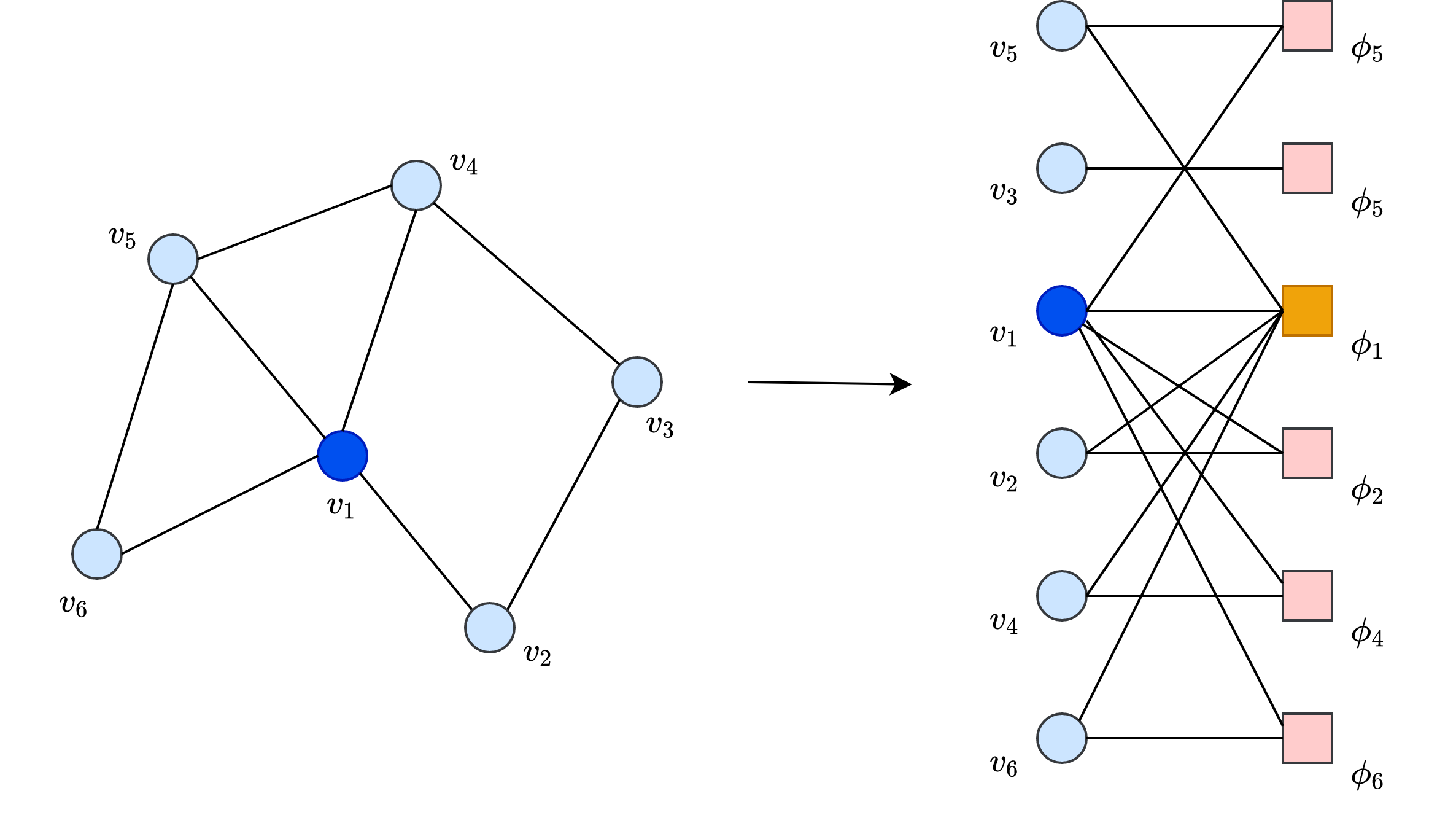}
  \caption{Factor Graph model for molecular graphs.}
    \label{fig:factor_graph_moleculargraph}
\end{figure}

\textbf{Data} ~~ We evaluate our model on two large scale datasets on quantum chemistry, QM9~\citep{ruddigkeit2012enumeration,ramakrishnan2014quantum} and Alchemy~\citep{chen2019alchemy} datasets on the task of regression on 12 quantum-mechanical properties. QM9 is composed of 134K drug-like organic molecules with sizes varying from 4 to 29 atoms per molecule. Atoms belong to one of the 5 types, Hydrogen (H), Carbon (C), Oxygen (O), Nitrogen(N), and Fluorine (F) and each molecule may contain up to 9 heavy (non-Hydrogen) atoms. Nodes come with discrete features along with their 3D positions. We follow the standard 80:10:10 random split for training, validation, and testing. 

Alchemy~\citep{chen2019alchemy} is a recently released more challenging dataset of 119K organic molecules comprising of 7 atom-types H, C, N, O, F, S (Sulphur) and Cl (Chlorine) with up to 14 heavy atoms. These molecules are screened for being more likely to be useful for medicinal chemistry based on the functional group and complexity. Furthermore, we follow the split based on molecule size where almost all training set contains molecules with 9 or 10 heavy atoms while the validation and test set contain molecules with 11 or 12 heavy atoms. As quantum-mechanical properties depend on the molecule size, this split tests how well the model can generalize to heavier molecules. The regression targets are the same as in QM9 dataset.

\textbf{Architecture and training details} ~~ We run our message passing scheme on features initialized on MPNN network with the  standard implementation provided by Pytorch-geometric~\citep{fey2019fast}. The MPNN implementation has Edge-conv~\citep{gilmer2017neural} as message function, GRU~\citep{chung2014empirical} as update function followed by a Set2set~\citep{vinyals2015order} function as readout for the whole graph. Readout function is required as the task is a prediction on graphs and the readout function takes in node features and outputs a vector which is used for final regression. In our implementation, we run 3 iterations of MPNN message passing scheme on input graph followed by 3 iterations of higher-order FGNN message passing described in Algorithm \ref{algo:FGNN}. Learning node marginals is likely to be helpful in this case as we want good node representations to be combined in the readout function. We use max function as the aggregator in VF module and summation as aggregator in FV module. We select this combination as we found it useful while being numerically stable. We then combine the MPNN output from the third iteration and the FGNN output with concatenation followed by the set2set readout function. For FGNN module, we set the hidden vector dimension to 64. The projection dimension of VF module is set to to 512. We use Adam optimizer initialized with learning rate of $1e^{-3}$. Since we want to show improvements over MPNN, all other hyperparameters are maintained as is provided by Pytorch-geometric implementation of MPNN for a fair comparison. All targets are normalized and are trained with the absolute error loss for 200 epochs with batch size of 64.

\begin{table}[t]
  \scriptsize
  \caption{Graph Regression results on QM9 dataset. The error rates of MPNN, 123-GNN, PPGNN and NestedGNN are reported in different units in the respective papers. We have converted the reported numbers to units below according to multiples mentioned in Pytorch-Geometric~\citep{fey2019fast}. MPNN is a general GNN model and all the other baselines are learnable GNN models designed to be more expressive than MPNN. The backbone neural model of FGNN is MPNN.}
  \label{table-qm9}
  \centering
  \resizebox{\linewidth}{!}{ 	\renewcommand{\arraystretch}{1.05}	
  \begin{tabular}{lrrrrrrrrrrr}
    \toprule
    & & \multicolumn{5}{c}{Joint training of targets}  & \multicolumn{5}{c}{Separate training of targets} \\
    \cmidrule(lr){3-7} \cmidrule(lr){8-12}
    Target & Units & MPNN & 123-GNN & PPGNN & FGNN & Gain(\%) & 123-GNN & NestedGNN & PPGNN & FGNN & Gain(\%)  \\
    \midrule
    $\mu$ & D &  0.3580 & 0.4070 &  0.2310 &  \textbf{0.0920} & 60.19 &    0.4760 &  0.4330   & 0.0934 &     \textbf{0.0688} &  26.33\\
    $\alpha$ & {$a_0^3$} &   0.8900  &   0.3340 &   0.3820 & \textbf{0.1830} & 45.20  &     0.2700 &    0.2650 & 0.3180 &     \textbf{0.1403} &  47.05 \\
    $\epsilon_{homo}$ & meV & 147.21 &  2124  &  75.10  &  \textbf{54.15} & 27.89 &   100.68 & 75.91 & 47.34 &    \textbf{45.71} &  2.98 \\
    $\epsilon_{lumo}$ & meV & 169.52 & 2310.79   & 78.09 &   \textbf{55.78} & 28.57 &  95.51 & 75.10 & 57.14 &  \textbf{43.53} & 22.95 \\
    $\Delta_\epsilon$ & meV & 179.59 &  2976.65 &  110.47 &   \textbf{75.91} & 31.28 & 130.61 & 106.13 & 78.91 &   \textbf{65.30} & 15.51\\
    $\langle R^2\rangle$& {$a_0^2$}  & 28.50  &  22.83  &  16.07  & \textbf{2.81} &  82.53 &  22.90 &  20.10 & 3.78   &  \textbf{1.41} & 62.69\\
    $ZPVE$ &  {meV}  & 58.77 &  307.21  &  17.41  &   \textbf{4.898}  &  71.87 &  5.170 & 4.08  & 10.85 &     \textbf{2.72} & 33.33  \\
    $U_0$ &  {meV}  & 55783 &  242453 & 6367 &  \textbf{2468} & 61.25 & 1161 & 5578 & 598 &  \textbf{443} & 25.90  \\
    $U$&  {meV}  & 54422 & 242453 &  6367 &  \textbf{2468} & 61.23 & 3020 & 5442 & 1371 &   \textbf{378} & 72.42  \\
    $H$&  {meV}  & 54967 & 242453 &  6231 &  \textbf{2465} & 60.42  &  1140 & 5775 & 800 &   \textbf{476} & 40.47  \\
    $G$ &  {meV}  & 54967 & 242453 &  6476 &  \textbf{2468} & 61.91 &  1276 & 6884 & 653 &   \textbf{364} & 44.16 \\ 
    $C_v$ & {$\frac{\text{cal}}{\text{mol K}}$} &    0.42  &   0.1184  &   0.184  &   \textbf{0.0840}  & 29.05     &   0.0944 & 0.081 & 0.144 & \textbf{0.0552} & 32.09\\
    \bottomrule
  \end{tabular}
   }
\end{table}
\begin{table}[t]%
  \scriptsize
  \centering
  \caption{Graph Regression results on Alchemy dataset. The backbone neural model of FGNN is MPNN, which is also shown for comparison with FGNN.}
  \label{table-alchemy}
  \resizebox{0.75\linewidth}{!}{ \renewcommand{\arraystretch}{1.05}	
  \begin{tabular}{lrrrrrrr}
    \toprule
    \multirow{2}{*}{Target} & \multirow{2}{*}{MPNN*} & \multirow{2}{*}{FGNN} & \multirow{2}{*}{Gain(\%)} & \multicolumn{4}{c}{FGNN Network Ablation Models} \\
    \cmidrule(lr){5-8} %
                            &  &  & & CAT & BT & CABT & CABTA \\
    \midrule
    $\mu$& \textbf{0.1026}  & 0.1041 & -1.41   & 0.1091  & 0.1041  & 0.1092  & 0.1233 \\
    $\alpha$& 0.0557 & \textbf{0.0451}  & 19.05  & 0.0473  & 0.0451 & \textbf{0.0446}  & 0.0449 \\
    $\epsilon_{homo}$& 0.1151  & \textbf{0.1004} & 12.74   & 0.1065   & 0.1004  & \textbf{0.1001}  & 0.1007  \\
    $\epsilon_{lumo}$& 0.0817  & \textbf{0.0664} & 18.74 & 0.0712 & \textbf{0.0664}  & 0.0685  & 0.0696  \\
    $\Delta_\epsilon$& 0.0832   & \textbf{0.0691} & 16.88 & 0.0739   & \textbf{0.0691} & 0.0703   & 0.0720    \\
    $\langle R^2 \rangle$& 0.0271   & \textbf{0.0099} 	& 63.47   & 0.0099     & 0.0099  & \textbf{0.0094}   & 0.0120 \\
    $ZPVE$& 0.0259   & \textbf{0.0115} 	& 55.42 & 0.0116   & 0.0115      & \textbf{0.0108}  & 0.0140   \\
    $U_0$& 0.0131    & \textbf{0.0044}	& 65.80   & \textbf{0.0042}  & 0.0044    & 0.0046    & 0.0054    \\
    $U$& 0.0131    & \textbf{0.0044} 	& 65.90   & \textbf{0.0041}  & 0.0044    & 0.0046   & 0.0054   \\
    $H$& 0.0130     & \textbf{0.0044}	& 65.77  & \textbf{0.0042}     & 0.0044    & 0.0046    & 0.0054 \\
    $G$& 0.0130     & \textbf{0.0044}	& 65.77  & \textbf{0.0042}     & 0.0044  & 0.0046   & 0.0054  \\
    $C_v$& 0.0559  & \textbf{0.0481}	&13.93    & \textbf{0.0472}  & 0.0481   & 0.0488   & 0.0502  \\
    \midrule
    MAE & 0.0499  &	\textbf{0.0394}		& 21.18   & 0.04115  & \textbf{0.0394}  & 0.0400  & 0.0424     \\
    \bottomrule
  \end{tabular}
   }
\end{table}

\textbf{Results} ~~ For QM9 dataset, following~\citep{maron2019provably} we report Mean Absolute Error (MAE) in two settings, one where all targets are jointly trained and the other where each target is separately trained. Factors are constructed as described in modeling, with factor weights conditioned on the central atom and bond type (CABT). The baselines we compare with are MPNN~\citep{gilmer2017neural}, 123-GNN~\citep{morris2019weisfeiler}, PPGNN~\citep{maron2019provably} and NestedGNN~\citep{zhang2021nested}. MPNN is a generalized GNN similar to that of~\citet{battaglia2018relational} with good performance on QM9. 123-GNN and  PPGNN are the $k$-order methods which capture higher-order information. NestedGNN is another more expressive GNN which passes messages on rooted subgraphs instead of trees and is shown to do well on QM9 dataset. %
Table~\ref{table-qm9} shows that FGNN outperforms MPNN, 123-GNN, NestedGNN and PPGNN by a significant margin in all the targets under both the settings. Furthermore, the margin of improvement indicates that much of the higher-order information was not sufficiently captured by the $k$-order GNNs.%

For Alchemy dataset, following~\citep{chen2019alchemy} we report MAE on jointly trained normalized targets and compare with MPNN which was the best performing model in the benchmark results~\citep{chen2019alchemy}. We use the validation set to select among the models in Section~\ref{sec:molecular_data}. As FGNN is built on MPNN as the backend network, the margin of improvement in Table~\ref{table-alchemy} is mainly because of higher-order message passing module. We also did an ablation study using the different sharing configurations. Results indicate that conditioning of parameters on either central atom or edge type helps most. Conditioning in these ways helps capture most of the higher-order information centered around an atom (node). For the CABTA method, there is a slight decrease in performance which is likely caused by the large parameter size in the model.
Collectively, the ablation results suggest that major improvements are coming from the higher-order message passing scheme itself since conditioning on only bond types (BT) seems to be sufficient for a better performance.
\begin{figure}[th]%
  \centering
    \includegraphics[width=0.5\linewidth]{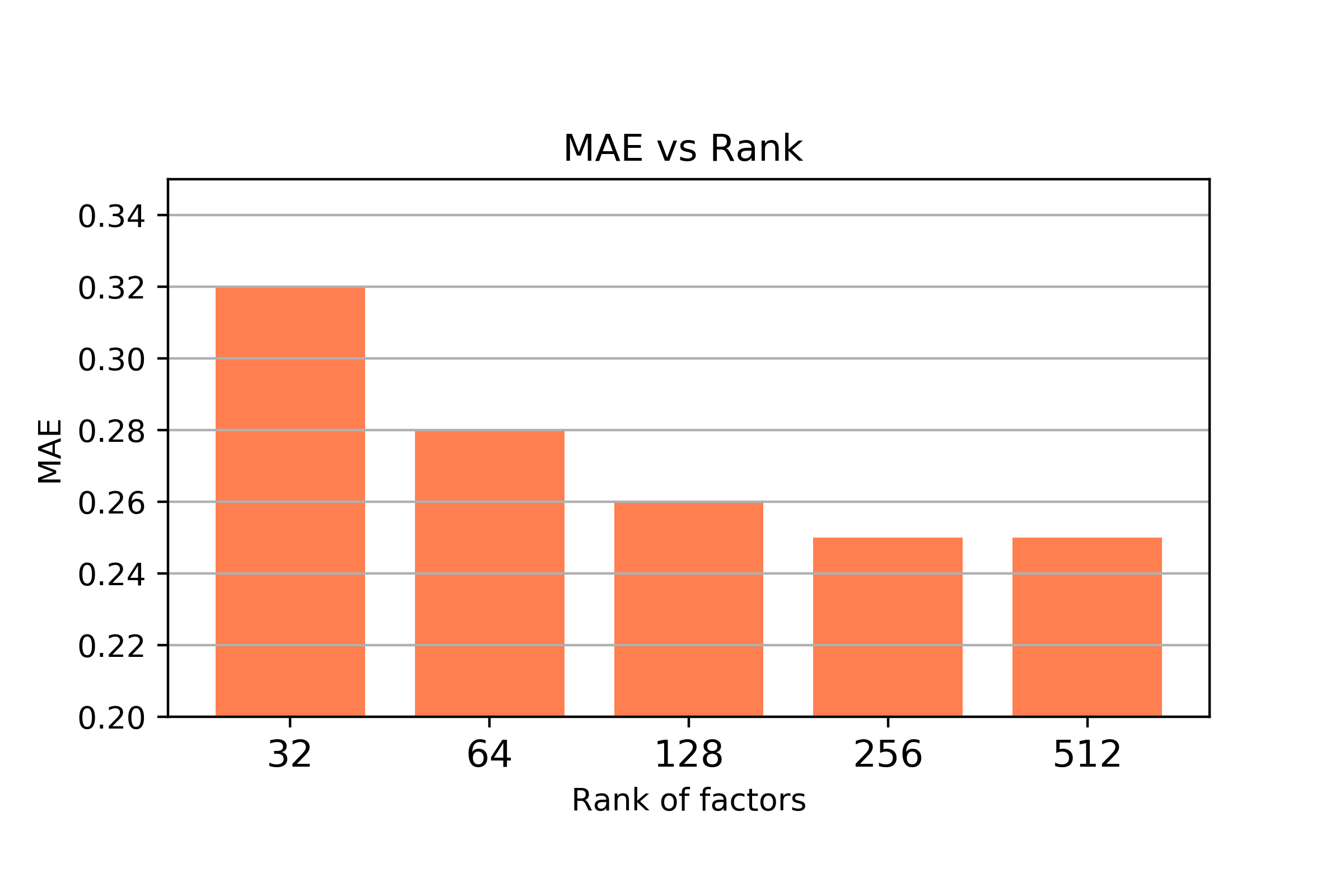}
    \caption{Effect of rank of factor parameters on QM9
      dataset. Mean absolute error of the model consistently decreases with the increasing rank and saturates after rank-$256$. In contrast, an exact representation of the full-rank tensor of the order-$5$ potential would need at least $3125$ components.  
      }\label{fig:qm9_mae-rank}	
\end{figure}
\paragraph{Effect of rank of factor parameters} We now study the sensitivity of the model performance MAE with the variation in the rank of the factor parameters on the molecular data as well. For this, we fix all the model hyperparameters described in architecture details and only vary the factor rank.  Figure~\ref{fig:qm9_mae-rank} shows the variation of MAE with the increasing factor rank on the QM9 dataset. Clearly, there is improvement in model performance with the increasing rank and saturates above 256. The rank of the higher-order factors in this task is considerably lower compared to the rank in the character recognition ablation in Figure~\ref{fig:ocr_rank_fig}. This can be explained with the different class sizes of the variables in both the tasks. The variables in the molecular data belong to $5$ atom types only, whereas the variables in character-recognition had a class size of $26$. Considering a factor of order-$5$ (for Carbon atom),  a full-rank tensor representation for the potential function would need at least $5^5=3125$ components for exact representation. Both of our ablation experiments on handwriting recognition and molecular data provide consistent evidence that the higher-order factors can be approximated with a mixture of relatively small number of rank-1 tensors.

\subsubsection{QM9 with positional information}
The QM9 dataset comes with additional atom coordinate features which can provide the positional as well as directional information. These directional features have been shown to be useful in Directional message passing network (Dimenet)~\citep{klicpera2020directional}, which was further confirmed in subsequent works of ALIGNN~\citep{choudhary2021atomistic},  SphereNet~\citep{liu2022spherical} and GEM~\citep{fang2022geometry}. The directional features, \ie positional and angular information, used in Dimenet are known to help substantially in QM9 dataset. Specifically, Dimenet extracts the features from triplets of nodes with each triplet feature constructed from the angular information within the triplet encapsulated within a basis function. In order to incorporate this informative directional information, we augment our FGNN model with the additional factors along with node positional coordinates. We conduct additional experiments with the added factors and compare with the Dimenet and other recent baselines in performance.

\begin{table}[t]
	\scriptsize
	\centering
	\caption{Comparison of FGNN with position coordinate factors with recent baselines. The backbone model of FGNN is MPNN. All the baselines compared are more powerful models than MPNN specifically designed for exploiting 3D geometric information of molecules.}
	\label{tab:qm9_exp2}
	\resizebox{\textwidth}{!}{
\begin{tabular}{llcccccccc}
	\toprule
	\multirow{2}{*}{Target} & \multirow{2}{*}{Units} &  \multicolumn{2}{c}{Joint training of targets} &
	\multicolumn{6}{c}{Seperate training of targets} \\
	\cmidrule(lr){3-4} \cmidrule(lr){5-10} 
	& & Dimenet & FGNN & Dimenet & MGCN &  EGNN & ALIGNN & SphereNet &FGNN \\
	\midrule
	$\mu$                        &                                      D &  0.0775 &       \textbf{0.0549} &  {0.0286} & 0.056 &   0.029 & \textbf{0.0146} & 0.0245 & {0.0373}\\
	$\alpha$                     &                                     {$a_0^3$} &  \textbf{0.0616} & {0.1364} &  {0.0469} & \textbf{0.030} &  0.071 & 0.0561 & 0.0449 & {0.0868}\\
	$\epsilon_\text{HOMO}$       &                         {meV} &    {45.1} &         \textbf{38.7} &    {27.8}  &  42.1 &  29 & \textbf{21.4} & 22.8 &{33.68} \\
	$\epsilon_\text{LUMO}$       &                         {meV} &    {41.1} &         \textbf{37.4} &    {19.7}&  57.4 &  25 & 19.5 & \textbf{18.9} & {31.26} \\
	$\Delta\epsilon$             &                         {meV} &    {59.2} &         \textbf{53.7} &    {34.8}  & 64.2 &  48 & 38.1 & \textbf{31.1} & {48.7} \\
	$\left< R^2 \right>$         &                                     {$a_0^2$} &   \textbf{0.345} &        {1.40} &   {0.331}  &    0.11 &  \textbf{0.106} & 0.543 & 0.268 &{0.365}\\
	ZPVE                         &                         {meV} &    \textbf{2.87} &         {3.51} &    {1.29}  &  \textbf{1.12} &  1.55  & 3.1 & \textbf{1.12} & {1.8}\\
	$U_0$                        &                         {meV} &    \textbf{12.9} &         {36.3} &    {8.02} & 12.9 &  11 & 15.3 & \textbf{6.26} & {21.5} \\
	$U$                          &                         {meV} &    \textbf{13.0} &         {36.6} &    {7.89} & 14.4 &  12 & 14.4 & \textbf{6.36} & {20.6}\\
	$H$                          &                         {meV} &    \textbf{13.0} &         {36.7} &    {8.11}  & 16.2 &  12 & 14.7 & \textbf{6.33} & {27.4} \\
	$G$                          &                         {meV} &    \textbf{13.8} &         {36.8} &    {8.98} & 14.6  &  12 & 14.4 & \textbf{7.78} & {26.9} \\
	$c_\text{v}$ \vspace{1pt}    &  {$\frac{\text{cal}}{\text{mol K}}$} &  \textbf{0.0309} &       {0.057} &  {0.0249}  & 0.038 &  0.031 & NA & \textbf{0.0215} & {0.0249}\\
	\bottomrule
\end{tabular}
}
\end{table}

\textbf{Modeling and Architecture} ~~ We augment the above FGNN model for molecular graphs with edge factors with positional coordinates. The edge factor connects two adjacent nodes in the molecular graph with elementwise multiplication as the aggregator function.
\begin{equation}
\tilde \gb_{ij} = \Mcal(\fb_i|\Theta_{\text{VF$_i$}}) \odot \Mcal(\fb_j|\Theta_{\text{VF$_j$}})
\end{equation}
where $\fb_i$ is the node position coordinate feature and $\Mcal()$ is an MLP. Note that unlike Dimenet, our model does not include features for each triplet of nodes.

\textbf{Results}
Table~\ref{tab:qm9_exp2} shows the results of the augmented FGNN model on QM9 dataset in both the settings \ie joint training of 12 targets and separate training of each target. Results suggest that FGNN performs competitively in most targets and under the setting of joint training of targets, FGNN is able to beat the Dimenet in four of the twelve targets. Note that FGNN uses only node position features without explicit angular features encapsulated in basis functions as in DImenet. Furthermore, for completeness, we compare with other models including MGCN~\citep{shui2020heterogeneous} and more recent baselines EGNN~\citep{satorras2021n}, ALIGNN\citep{choudhary2021atomistic}, SphereNet\citep{liu2022spherical} which all use the 3D coordinate positional information. Furthermore, note that these models, including Dimenet, are specifically designed for molecular graph structures with domain-specific inductive biases while FGNN is a general model which can be used over any graph structure. Hence, FGNN is not able to outperform these models, however, can perform competitively. 

\begin{table}[t]
  \scriptsize
  \centering
  \caption{Long-term prediction error (the smaller the better) of joint angles (top) and 3D joint positions (bottom) on H3.6M. Our model share the same backbone as \citet{mao2019learning}, and we replace the last two layer of GNN in \citet{mao2019learning} with our FGNN model.}
  \label{tab:hmp_long}
  \begin{tabularx}{\textwidth}{ccccccccccc}
    \toprule
    &   \multicolumn{2}{c}{Walk} & \multicolumn{2}{c}{Eating}&
                                                               \multicolumn{2}{c}{Smoking}
    & \multicolumn{2}{c}{Discussion} & \multicolumn{2}{c}{Average}\\
    milliseconds & 560 & 1000& 560& 1000& 560& 1000 & 560& 1000& 560 & 1000\\
    \midrule
    convSeq2Seq\citep{li2018convolutional}  & N/A&0.92&N/A&1.24&N/A&1.62&N/A&1.86&N/A&1.41 \\
    GNN\citep{mao2019learning} &
                                 \textbf{0.65} &0.67&0.76& \textbf{1.12}&0.87&1.57&\textbf{1.33}&1.70&0.90&1.27 \\
    DMGNN \citep{li2020dynamic} & 0.66 & 0.75 & \textbf{0.14} & 1.14 & \textbf{0.83} & \textbf{1.52} & \textbf{1.33} &
                                                                    \textbf{1.45}
                             & \textbf{0.89} & \textbf{1.21}\\
    Ours & 0.67 & \textbf{0.70} & 0.76 & \textbf{1.12} & 0.88 & 1.57 & 1.35 & 1.70 &
                                                                   0.91&
                                                                         1.27\\
    \toprule
    convSeq2Seq\citep{li2018convolutional} &69.2&81.5&71.8&91.4&50.3&85.2&101.0&143.0&73.1&100.3 \\
    GNN\citep{mao2019learning} & 55.0 &
                                        60.8&68.1&79.5&42.2&70.6&93.8&119.7&64.8&82.6\\
    Hist-Attention \citep{mao2020history} & 47.4 & 58.1 & \textbf{50.0} & 75.7 & 47.6 & 69.5 & \textbf{86.6} & 119.8 &
                                                                     57.9
                                        & 80.7\\
    Ours & \textbf{44.1} & \textbf{53.5} & 59.5& \textbf{73.0} & \textbf{33.0} & \textbf{61.9} & 86.9 & \textbf{113.5} & \textbf{55.9} & \textbf{75.5}\\
    \bottomrule
  \end{tabularx}
\end{table}

\subsection{Human Motion Prediction (Sequential Prediction)}
\label{sec:human-moti-pred}
The human motion prediction aims at predicting the future motion of a
human given a history motion sequence. As there are obviously higher
order dependencies between joints, the factor graph neural network may
help to improve the performance of the predictor. In this section, we
consider the human motion prediction problem for the skeleton data, where
the angle and 3d position of each joint are predicted. We build a factor
graph neural network model for the skeleton data and compare the FGNN
model with the state-of-the-art model based on GNN. %

\textbf{Data and Modeling} ~~ For human motion prediction, we are using the Human 3.6M (H3.6M) dataset. In this experiment, we replace the last two GNN layer in
\citep{mao2019learning}'s model with FGNN layer with the same number
of output channels. The H3.6M dataset includes seven actors performing 15 varied activities such as walking, smoking \etc. The poses of the actors are represented as an
exponential map of joints, and a special pre-processing of global
translation and rotation. In our experiments, as in previous
work\citep{li2018convolutional,mao2019learning}, we only predict the
exponential map of joints. That is, for each joints, we need to
predict a 3-dimensional feature vector. Thus we add a factor for the 3
variable for each joint \footnote{In practice, those angles with very
	small variance are ignored, and these variables are not added to the
	factor graph}. Also for two adjacent joint, a factor of 6 variables
are created. The factor node feature are created by put all its
variable node feature together. For the edge feature, we simply use
one hot vector to represent different factor-to-variable edge. For
evaluation, we compared 4 commonly used action --- walk, eating,
smoking and discussion. The result of GNN and convSeq2Seq are taken
from \citep{mao2019learning}, and our FGNN model also strictly
followed the training protocol of \citep{mao2019learning}. 
	
\textbf{Architecture and training details} ~~ We train our model on
the Human3.6M  dataset using the standard training-val-test split as
previous works \citep{mao2019learning, li2018convolutional,
  martinez2017human}, and we train and evaluate our model using the same protocol
as \citep{mao2019learning} (For details, see the Appendix). 

\textbf{Results} ~~ The results are provided in Table
\ref{tab:hmp_long}. For angle error, our FGNN model achieves similar
results compared to the previous state-of-the-art GNN-based method
\citep{mao2019learning,li2020dynamic}, while for 3D position error, our model
achieves superior performance than the state-of-the-art models \citep{mao2019learning,mao2020history}. Furthermore, since the backbone of our model is same as that of~\citet{mao2019learning}, the performs improvement suggests that the higher-order modeling with FGNN is able to capture better higher-order structural priors compared to the pairwise GNN. %

\section{Conclusion}
\label{sec:conclusion}
In this paper, we derive an efficient Low-rank Sum-Product Loopy Belief Propagation procedure for inference in factor graphs. %
The derived update functions are \emph{simple}---need only matrix multiplication and Hadamard product operations, and \emph{efficient}---the complexity of message updates grows linearly with the number of variables in the factor. In order to learn better node representations with end-to-end training, we neuralize the message passing updates to give Factor Graph Neural Network (FGNN) allowing the network to capture higher-order dependencies among the variables. We then showed FGNN can also represent the execution of the Max-Product inference algorithm on probabilistic graphical models, providing a graph neural network architecture that can represent both the Sum and Max-Product belief propagation inference algorithms. %

Furthermore, we showed multiple ways of modeling higher-order factors with graph structured input data. This gives us a fairly simple, flexible, powerful, and efficient message passing scheme for representation learning of graph data where higher-order information is present. We evaluated the proposed model with extensive experiments on various tasks and domains including inference in PGMs, molecular and vision datasets where it either outperforms other state-of-the-art models substantially or is at least competitive enough. The FGNN provides a convenient method of capturing arbitrary dependencies in graphs and hypergraphs, including those with typed or conditioned nodes and edges, opening up new opportunities for adding higher-order inductive biases into learning and inference problems. More importantly, it provides a deeper theoretical understanding of the relationship between graph neural networks and inference algorithms on graphical models.

\acks{This work was supported by the National Research Foundation Singapore
under its AI Singapore Program (Award Number: AISGRP- 2018-006). Any
opinions, findings and conclusions or recommendations expressed in
this material are those of the author(s) and do not reflect the views
of National Research Foundation, Singapore. Zhen Zhang's participation
was partially supported by the Australian Research Council Grant
DP160100703. Zhen Zhang and Javen Shi's participation were partially supported by Centre for Augmented Reasoning at the Australian Institute for Machine Learning.}

\appendix
\onecolumn
\section{Proof of propositions}
\subsection{Propositions for decomposing higher-order potentials}
First we provide Lemma \ref{lemma:lossless_agg}, which will be used in
the proof of Proposition \ref{propos:decomposition} and \ref{propos:feature_sum}. 
\begin{lemma}\label{lemma:lossless_agg}
  Given $l$ non-negative feature vectors $\fb_i=[f_{i0}, f_{i1}, \ldots,
  f_{ik}]$, where $ i=1,\ldots, l$, there
  exists $l$ matrices $\Qb_i$  with shape $lk\times k$ and $l$ vector
  $\hat \fb_i = \Qb_i\fb_i^{T}$, s.t.
  \begin{align*}
    , \qquad [\fb_1, \fb_2, \ldots, \fb_l] =
    [\max_{i}\hat f_{i0},\max_{i}\hat f_{i1},\ldots, \max_{i}\hat f_{i,kl}].
  \end{align*}
\end{lemma}
\begin{proof}
  Let
  \begin{align}
    \label{eq:6}
    \Qb_i = \left[\underbrace{\mathbf{0}^{k\times k}, \ldots, 
    \mathbf{0}^{k\times k}}_{i-1\text{ matrices}}, \mathbf{I},
    \underbrace{\mathbf{0}^{k\times k},\ldots,
    \mathbf{0}^{k\times k}}_{n - i  \text{ matrices}}\right]^{\top},
  \end{align}
  then we have that
  \begin{align*}
    \hat \fb_i = \Qb_i\fb_i^T = \left[\underbrace{0, \ldots, 0}_{(i-1)k
    \text{ zeros}},f_{i0}, f_{i1}, \ldots, f_{ik}, \underbrace{0,
    \ldots, 0}_{(n - i)k \text{ zeros}}  \right]^{\top}.
  \end{align*}
  By the fact that all feature vectors are non-negative, obviously we
  have that
  $[\fb_1, \fb_2, \ldots, \fb_l] =
  [\max_{i}\hat f_{i0},\max_{i}\hat f_{i1},\ldots, \max_{i}\hat f_{i,kl}]$.  
\end{proof}
Lemma \eqref{lemma:lossless_agg} suggests that for a group of feature
vectors, we can use the $\Qcal$ operator to produce several $\Qb$
matrices to map different vector to
different sub-spaces of a high-dimensional spaces, and then our
maximization aggregation can sufficiently gather information from
the feature groups.

\begin{repproposition}{propos:decomposition}
  A factor graph $\Gcal = (\Vcal,\Ccal,\Ecal)$ with variable log potentials $\theta_i(x_i)$ and 
  factor log potentials $\varphi_c(\xb_c)$ can be converted to a factor graph $\Gcal'$ with the same variable potentials and the decomposed  log-potentials $\varphi_{ic}(x_{i}, z_c)$ using a one-layer FGNN. 
\end{repproposition}
\begin{proof}
  Without loss of generality, we assume that $\log\phi_c(\xb_c)\geqslant 1$. Then let
  \begin{align}\label{eq:decomposition_to_node}
    \theta_{ic}(x_i, z_c) = \left\{ \begin{array}{ll}
                                      \frac{1}{|s(c)|}\log\phi_c(\xb_c^{z_c}),     & \text{if } \hat x_i = x_i^{z_c}, \\
                                      -c_{x_i, z_c},     &  \text{otherwise,}
                                    \end{array}
                                                           \right.
  \end{align}
  where $c_{x_i, z_c}$ can be arbitrary real number which is larger than
  $\max_{\xb_c}\theta_c(\xb_c)$. 
  Obviously we will have
  \begin{align}
    \log\phi_c(\xb_c) = \max_{z_c}\sum_{i\in s(c)}\theta_{ic}(x_i, z_c)
  \end{align}
  Assume that we have a factor $c={1,2,\ldots n}$, and each nodes can take $|X|$ states. Then $\xb_c$ can be sorted as 
  \begin{align*} 
    [&\xb_c^0=[x_1=0,x_2=0, \ldots, x_n=0],\\
     &\xb_c^1=[x_1=1,x_2=0, \ldots, x_n=0],\\
     &\ldots, \\
     &\xb_c^{|X|^n-1}=[x_1=|X|,x_2=|X|, \ldots, x_n=|X|]
       ],
  \end{align*}
  and higher-order potential can be organized as vector $\gb_c =
  [\log \phi_c(\xb_c^0), \log \phi_c(\xb_c^1), \ldots, \log
  \phi_c(\xb_c^{|X|^n - 1})]$. Then for each $i$ the item
  $\theta_{ic}(x_i, z_c)$ in \eqref{eq:decomposition_to_node} have
  $|X|^{n+1}$ entries, and each entry is either a scaled entry of the
  vector $\gb_c$ or arbitrary negative number less than
  $\max_{\xb_c}\theta_c(\xb_c)$.
  
  Thus if we organize $\theta_{ic}(x_i, z_c)$ as a length-$|X|^{n+1}$
  vector $\fb_{ic}$, then we define a $|X|^{n+1}\times |X|^n$ matrix $\Qb_{ci}$, where
  if and only if the \zzth{l} entry of $\fb_{ic}$ is set to the \zzth{m} entry of
  $\gb_c$ multiplied by $1/|s(c)|$, the entry of $\Qb_{ci}$ in \zzth{l} row,
  \zzth{m} column will be set to $1/|s(c)|$; all the other entries of
  $\Qb_{ci}$ is set to some negative number smaller than
  $-\max_{\xb_c}\theta_c(\xb_c)$. Due to the assumption that
  $\log\phi_c(\xb_c)\geqslant 1$, the matrix multiplication $\Qb_{ci} \gb_c$
  must produce a legal $\theta_{ic}(x_i, z_c)$. 
  
  If we directly define a $\Qcal$-network which produces the above matrices
  $\Qb_{ci}$, then in the aggregating part of our network there might be
  information loss. However, by Lemma \ref{lemma:lossless_agg} there
  must exists a group of $\tilde \Qb_{ci}$ such that the maximization
  aggregation over features $\tilde\Qb_{ci}\Qb_{ci}\gb_c$ will produce exactly
  a vector representation of $\theta_{ic}(x_i, z_c), i\in s(c)$. Thus if
  every $t_{ci}$ is a different one-hot vector, we can easily using one
  single linear layer $\Qcal$-network to produce all
  $\tilde\Qb_{ci}\Qb_{ci}$, and with a $\Mcal$-network which always
  output factor feature, we are able to output a vector representation
  of $\theta_{ic}(x_i, z_c), i\in s(c)$ at each factor node $c$. 
\end{proof}

\subsection{Derivation of decomposed max-product belief propagation}
In this section, we reformulate the \eqref{max_product_factor} using
the decomposed higher-order-log potentials. We use
$m_{c\rightarrow i}(x_i)$ and $b_i(x_i)$ for the previous messages
and beliefs, and use $m'_{c\rightarrow i}(x_i)$ and $b'_i(x_i)$ for
the updated messages and beliefs. Then message updating step in the max product belief
propagation \eqref{max_product_factor} can be reformulated as
\allowdisplaybreaks
\begin{subequations}
  \label{eq:updated_mp_iter}
  \begin{align}
    n_{i\rightarrow c}(x_i) = & \theta_i(x_i) +  \sum_{d: d\neq c, i \in
                                s(d)}m_{d\rightarrow i}(x_i), \notag\\
    = & \theta_i(x_i) +  \sum_{d: i \in
                                s(d)}m_{d\rightarrow i}(x_i) -
        m_{c\rightarrow i}(x_i)\notag \\
    = & b_i(x_i) - m_{c\rightarrow i}(x_i)\\
    m'_{c\rightarrow i}(x_i) =& \max_{\xb_c\setminus
    x_i}\left[\theta_c(\xb_c) + \sum_{{j}\in
                              s(c), {j}\neq i}n_{{j}\rightarrow
                                c} (x_{{j}}) \right]\notag\\
    = & \max_{\xb_c\setminus
    x_i}\left\{\max_{z_c}\left[\sum_{{j} \in s(c), {j}\neq i}\varphi_{{j}c}(x_{{j}}, z_c) + \varphi_{ic}(x_i, z_c)\right]+ \sum_{{j}\in
                              s(c), {j}\neq i}n_{{j}\rightarrow
                                c} (x_{{j}})  
        \right\}        \notag\\
    = & \max_{\xb_c\setminus
    x_i}\left\{\max_{z_c}\left[\sum_{{j} \in s(c), {j}\neq i}\varphi_{{j}c}(x_{{j}}, z_c) + \varphi_{ic}(x_i, z_c)\right]+ \sum_{{j}\in
                              s(c, {j}\neq i)}\left[b_{{j}}(x_{{j}}) -
        m_{c\rightarrow {j}}(x_{{j}})\right]       \right\} \notag \\
    = & \max_{z_c}\max_{\xb_c\setminus x_i}\left\{\left[\sum_{{j} \in s(c), {j}\neq i}\varphi_{{j}c}(x_{{j}}, z_c) + \varphi_{ic}(x_i, z_c)\right]+ \sum_{{j}\in
    s(c, {j}\neq i)}\left[b_{{j}}(x_{{j}}) -
m_{c\rightarrow {j}}(x_{{j}})\right] \right\}\notag \\
    = & \max_{z_c}\left\{
      \sum_{{j}\in s(c), {j}\neq i}\max_{x_{{j}}}\left[\varphi_{{j}c}(x_{{j}}) - m_{c\rightarrow {j}}(x_{{j}}) + b_{{j}}(x_{{j}})
       \right] + \varphi_{ic}(x_i, z_c)
      \right\} \label{eq:deriv_msg_update}
  \end{align}
\end{subequations}
Here for simplying the notation we define 
\begin{align*}
  b_{c\rightarrow i}(z_c) = & \sum_{{j}\in s(c),{j}\neq i}\max_{x_{j}}\left[\varphi_{{j}c}(x_{{j}}, z_c) - m_{c\rightarrow {j}}(x_{{j}}) + b_{{j}}(x_{{j}})\right], \forall c, i\in s(c)\\
\end{align*}
and then the updating rule for beliefs can be reformulated as 
\begin{align*} 
  b'_i(x_i) &=   \theta_i(x_i) + \sum_{c: i \in s(c)}m'_{c\rightarrow
  i}(x_i) \notag \\
  &= \theta_i(x_i) + \sum_{c: i \in s(c)}\max_{z_c}\left[b_{c\rightarrow i}(z_c) + \varphi_{ic}(x_i, z_c) \right].
\end{align*}
Thus finally the max-product updating rules are 
\begin{align*}
  b_{c\rightarrow i}(z_c) \leftarrow & \sum_{{j}\in s(c),{j}\neq i}\max_{x_{{j}}}\left[\varphi_{{j}c}(x_{{j}}, z_c) - m_{c\rightarrow {j}}(x_{{j}}) + b_{{j}}(x_{{j}})\right], \\
  m_{c\rightarrow i} (x_i) \leftarrow & \max_{z_c}\left[b_{c\rightarrow i}(z_c) + \varphi_{ic}(x_i, z_c) \right],\\
  b_{i}(x_i) \leftarrow & \theta_i(x_i) +  \sum_{c: i \in s(c)} m_{c\rightarrow i} (x_i) 
\end{align*}

\subsection{Recovering decomposed max-product belief propagation using FGNN}
Given the log potentials represented as a set of rank-1 tensors at each factor node, we need to show that each iteration of the Max Product message passing update can be represented by a Variable-to-Factor layer followed by a Factor-to-Variable layer (forming a FGNN layer). We reproduce the update equations here. 
\begin{subequations}
  \begin{align}\label{eq:bp_supp}
    b_{c\rightarrow i}(z_c) \leftarrow & \sum_{{j}\in s(c),{j}\neq i}\max_{x_{j}}\left[\varphi_{{j}c}(x_{{j}}, z_c) - m_{c\rightarrow {j}}(x_{{j}}) + b_{{j}}(x_{{j}})\right], \\
  m_{c\rightarrow {j}(x_i)} \leftarrow &
                                        \max_{z_c}\left[b_{c\rightarrow
                                        i}(z_c) + \varphi_{ic}(x_i,
                                        z_c) \right], \quad 
  b_{i}(x_i) \leftarrow \theta_i(x_i) +  \sum_{c: i \in s(c)} m_{c\rightarrow {j}(x_i)} 
  \end{align}
\end{subequations}
In the Max-Product updating procedure, we should keep all the decomposed
$\varphi_{{j}c}(x_{{j}}, z_c) = \log\phi_{{j}c}(x_{{j}}, z_c)$  and all the unary potential
$\theta_i(x_i)$ for use at the next layer. That requires the FGNN to have the ability to
fit the identity mapping. Consider letting the $\Qcal$ network to always output
identity matrix, $\Mcal([\gb_c, f_i]|\Theta_{\text{VF}})$ to always output
$\gb_c$, and $\Mcal([\gb_c, f_i]|\Theta_{\text{FV}})$ to always output
$f_i$. Then the FGNN will be an identity mapping. As $\Qcal$ always
output a matrix and $\Mcal$ output a vector, we can use part of their
blocks as the identity mapping to keep $\log\phi_{{j}c}(x_{{j}}, z_c)$ and
$\theta_i(x_i)$. The other blocks are used to updating
$b_{c\rightarrow i}(z_c)$, 
messages $m_{c\rightarrow {j}}(x_{{j}})$, and $b_i(x_i)$.

First we show that $\Mcal$ operators in the Variable-to-Factor layer can be used to construct the computational graph for the max-marginal operations.
\begin{repproposition}{propos:matrix_max}
  For arbitrary real valued feature
  matrix $\mathbf{X} \in \mathbb{R}^{k\times l}$ with $x_{ij}$ as its
  entry in the $i^{\text{th}}$ row and $j^{\text{th}}$ column, the feature mapping operation
  $
  \hat \xb = [\max_{j}x_{ij}]_{i=1}^k
  $
  can be exactly parameterized with a 2$\log_2 l$-layer neural network with RELU as activation function and at most $2n$ hidden units. 
\end{repproposition}

\begin{proof}
  Without loss of generality we assume that $k=1$, and then we use $x_i$ to denote $x_{1i}$.
  When $l=2$, it is obvious that 
  \[
    \max(x_1,x_2) = \textbf{Relu}(x_1-x_2) + x_2 = \textbf{Relu}(x_1-x_2) + \textbf{Relu}(x_2) - \textbf{Relu}(-x_2)
  \]
  and the maximization can be parameterized by a two layer neural network with 3 hidden units, which satisfied the proposition. 
  
  Assume that when $l=2^a l$ for some integer $a>=1$, the proposition is satisfied \footnote{For any $l$ we can always padding the vector to get an $l'=2^a$ for some interger $a$}. Then for $l=2^{a+1}$, we can find $\max(x_1, \ldots, x_{2^a})$ and $\max(x_{2^a+1}, \ldots, x_{2^{a+1}})$ using two network with $2a$ layers and at most $2^{a+1}$ hidden units. Stacking the two neural network together would results in a network with $2i$ layers and at most $2^{i+2}$ parameters. Then we can add another 2 layer network with 3 hidden units to find $\max(\max(x_1, \ldots, x_{2^a}), \max(x_{2^a+1}, \ldots, x_{2^{a+1}}))$. Thus by mathematical induction the proposition is proved.
\end{proof}

The update equations contain summations of columns of a matrix after
the max-marginal operations. However, the VF and FV layers use max
operators to aggregate features produced by $\Mcal$ and $\Qcal$
operator. Assume that the $\Mcal$ operator has produced the
max-marginals, then we  use the $\Qcal$ to produce several weight
matrix. The max-marginals are multiplied by the weight matrices to
produce new feature vectors, and the maximization aggregating function
are used to aggregating information from the new feature vectors. We
use the following propagation to show that the summations of
max-marginals can be implemented by one MPNN layer plus one
linear layer. Thus we can use the VF layer plus a linear layer to produce 
$b_{c\rightarrow i}(z_c)$ and use the FV layer plus another linear layer 
to produce $b_i(x_i)$. 
Hence to do $k$ iterations of Max Product, we need $k$ FGNN layers followed by a linear layer. 
\begin{repproposition}{propos:feature_sum}
  For arbitrary non-negative valued feature
  matrix $\mathbf{X} \in \mathbb{R}_{\geqslant 0}^{k\times l}$ with $x_{ij}$ as its
  entry in the $i^{\text{th}}$ row and $j^{\text{th}}$ column, there
  exists a constant tensor $\mathbf{W} \in
  \mathbb{R}^{k \times l \times kl}$ that can be used to transform $\mathbf{X}$ into an intermediate representation $y_{ir} = \sum_{ij}x_{ij}w_{ijr}$, such that after maximization operations are done to obtain $\hat y_r = \max_{i}y_{ir}$, we can use another constant matrix $\mathbf{Q}\in \mathbb{R}^{l
    \times kl }$ to obtain 
  \begin{equation}
    [\sum_ix_{ij}]_{j=1}^{l} = \Qb[\hat y_r]_{r=1}^{kl}.
  \end{equation}
\end{repproposition}

\begin{proof}
  The proposition is a simple corollary of Lemma
  \ref{lemma:lossless_agg}. The tensor $\Wb$ serves as the same role as
  the matrices $\Qb_i$ in  Lemma
  \ref{lemma:lossless_agg}, which can convert the feature matrix $\Xb$
  as a vector, then a simple linear operator can be used to produce the
  sum of rows of $\Xb$, which completes the proof.
\end{proof}

In Lemma \ref{lemma:lossless_agg} and Proposition
\ref{propos:feature_sum}, only non-negative features are considered, while
in log-potentials, there can be negative entries. However, for the MAP
inference problem in \eqref{eq:map_def}, the transformation as follows
would make the log-potentials non-negative without changing the final
MAP assignment,
\begin{align}
  \tilde \theta_i(x_i) = \theta_i(x_i) - \min_{x_i}\theta_i(x_i),
  \qquad \tilde \theta_c(\xb_c) = \theta_c(\xb_c) - \min_{\xb_c}\theta_c(\xb_c). 
\end{align}
As a result, for arbitary PGM we can first apply the above transformation to make the log-potentials non-negative, and then our FGNN can exactly do Max-Product Belief Propagation on the transformed non-negative log-potentials.

\subsection{A Factor Graph Neural Network Module Recovering the Belief Propagation}
\label{sec:factor-graph-neural}
\begin{figure}[H]
  \centering
  \includegraphics[width=0.8\textwidth]{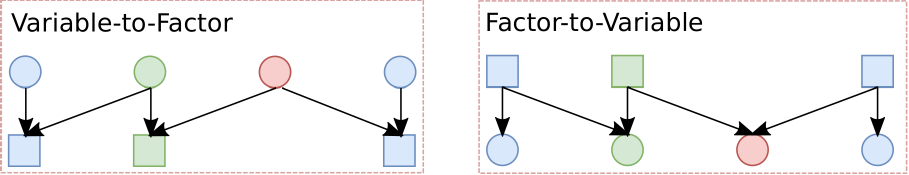}
  \caption{\small Directed bipartite graph for constructing FGNN layers. In
    the Variable-to-Factor sub-graph, each factor receives the
    messages from the same number of nodes. On the other hand, for
    each Factor-to-Variable sub-graph, each nodes may receives
    messages from different number of factors.}
  \label{fig:DBG}
\end{figure}

In this section, we give the proofs of Proposition
\ref{propos:first_layer} and \ref{propos:second_layer} by constructing
two FGNN layers which exactly recover the belief propagation
operation. As lower order factors can always shrank by higher-order
factors, we will construct the FGNN layers on an factor graph $\Hcal =
(\Vcal, \Fcal, \hat \Ecal)$, which satisfies the following condition
\begin{enumerate}
\item $\forall i \in \Vcal$, the associated $\theta_i(x_i)$ satisfies
  that $\theta_i(x_i) > 0 \forall x_i \in X$;
\item $\forall f_1, f_2 \in \Fcal$, $|f_1| = |f_2|$;
\item $\forall f \in \Fcal$, the corresponding $\varphi_f(\xb_f)$ can
  be decomposed as
  \begin{align}
    \label{eq:7}
    \varphi_f(\xb_f) = \max_{z_f\in \Zcal}\sum_{i\in
    f}\varphi_{fi}(x_i, z_f),
  \end{align}
  and $\forall i \in f, \varphi_{fi}(x_i, z_f)$ satisfies that $\varphi_{fi}(x_i,
  z_f) > 0.$
\end{enumerate}
On factor graph $\Hcal$, we construct a FGNN layer on the directed
bipartite graph in Figure \ref{fig:DBG}.

\paragraph{FGNN Layer to recover \eqref{eq:to_z}} Here we construct an
FGNN layer to produce all $b_{f\rightarrow i}(z_f)$. First we
reformulate \eqref{eq:to_z} as
\begin{equation}
  \label{eq:bfi_reform}
  \begin{split}
    b_{f\rightarrow i}(z_f) &\leftarrow \tilde \varphi_f(z_f) -
    \max_{x_i}[\varphi_{if}(x_i, z_f) - m_{f\rightarrow i}(x_i)+ b_i(x_i)],\\
    \tilde \varphi_f(z_f) &\leftarrow \sum_{i \in
      f}\max_{x_i}[\varphi_{if}(x_i, z_f) - m_{f\rightarrow i}(x_i) + b_i(x_i)].
  \end{split}
\end{equation}
Here we use the Variable-to-Factor sub-graph to implement
\eqref{eq:bfi_reform}. For each variable node $i$, we associated it with an
length-$|X|$ vector $[b_i(x_i)]_{x\in X}$ (Initially $b_i(x_i)=\theta_i(x_i)$). For each edge in the
sub-graph,  assume that $f=[i_1, i_2, \ldots, i_{|f|}]$, then for some
$i_j\in f$, the associated feature vector is as length-$|f|$ one-hot
vector as follows
\[
  [0, 0, \ldots, \underbrace{1}_{\text{The }j^{\text{th}}\text{
      entry.}}, \ldots, 0].
\]
For each factor node $f=[i_1, i_2, \ldots, i_{|f|}]$ in the sub-graph, it is associated with an
$|f|\times |X||Z|$ feature matrix as follows
\[
  \left[
    \begin{matrix}
      [\varphi_{fi}(x_{i_1}, z_f)- m_{f\rightarrow i}(x_i)]_{x_{i_1}=1, z_f=1}^{x_{i_1}=|X|, z_f=|Z|}\\
      [\varphi_{fi}(x_{i_2}, z_f)- m_{f\rightarrow i}(x_i)]_{x_{i_2}=1, z_f=1}^{x_{i_2}=|X|, z_f=|Z|}\\
      \ldots \\
      [\varphi_{fi}(x_{i_{|f|}}, z_f)- m_{f\rightarrow i}(x_i)]_{x_{i_{|f|}}=1, z_f=1}^{x_{i_{|f|}}=|X|,z_f=|Z|}
    \end{matrix}
  \right].
\]
Then we construct an MPNN

\begin{align}
  \label{eq:MPNN_spec_app}
  \tilde \fb_{i} = \max_{i \in f}\Qcal(\eb_{f\rightarrow i})\Mcal(\fb_i, \fb_f)
  ,
\end{align}
as follows. The $\Qcal(\eb_{f\rightarrow i})$ is an identity
mapping. The $\Mcal(\fb_i, \fb_{f})$ consists of $|f|$ addition
networks, where the $i_{j}^{\text{th}}$ networks will have an
$|f|\times |X||Z|$ parameter
\[
  \left[
    \begin{matrix}
      -\infty\\
      -\infty\\
      \ldots \\
      [\varphi_{fi}(x_{i_j}, z_f)- m_{f\rightarrow i}(x_i)]_{x_{i_{j}}=1,
        z_f=1}^{x_{i_{j}}=|X|,z_f=|Z|}\\
      \ldots\\
      -\infty
    \end{matrix}
  \right].
\]
In the $\Mcal$-network, the $|f|\times |X||Z|$ parameter will be added
to the $|f|\times |X||Z|$ and then the result will be reshaped to an
$|f|\times |X|\times |Z|$ tensor. After that the tensor will be added
to the length-$|X|$ feature vector of each nodes (reshaped to $1\times
1\times |X| \times 1$ tensor). In that case, for each $i_j \in f$, the
$i_{k}^{\text{th}}$ will produce
\[
  \left[
    \begin{matrix}
      -\infty\\
      -\infty\\
      \ldots \\
      [\varphi_{fi}(x_{i_k}, z_f) - m_{f\rightarrow i}(x_i) + b_{i_j}(x_{i_j})]_{x_{i_{k}}=x_{i_{j}}=1,
        z_f=1}^{x_{i_{k}}=x_{i_{j}}=|X|,z_f=|Z|}\\
      \ldots\\
      -\infty
    \end{matrix}
  \right].
\]
The $|f|$  $|f|\times |X|\times|Z|$ tensors will be stacked into an
$|f|\times|f| \times |X|\times |Z|$ tensor, and it will be multiplied
by the length-$|f|$ one-hot edge feature vector. That will produce
\[
  \left[
    \begin{matrix}
      -\infty\\
      -\infty\\
      \ldots \\
      [\varphi_{fi}(x_{i_j}, z_f) - m_{f\rightarrow i_j}(x_{i_j}) + b_{i_j}(x_{i_j})]_{x_{i_{j}}=1,
        z_f=1}^{x_{i_{j}}=|X|,z_f=|Z|}\\
      \ldots\\
      -\infty
    \end{matrix}
  \right].
\]
Then the $\max$ operation over all $i\in f$ will produce edge feature
matrix
\[\left[
    \begin{matrix}
      [\varphi_{f{i_1}}(x_{i_1}, z_f) - m_{f\rightarrow i_1}(x_{i_1})+ b_{i_{1}}(x_{i_{1}})]_{x_{i_1}=1, z_f=1}^{x_{i_1}=|X|, z_f=|Z|}\\
      [\varphi_{f{i_2}}(x_{i_2}, z_f) - m_{f\rightarrow i_2}(x_{i_2}) + b_{i_{2}}(x_{i_{2}})]_{x_{i_2}=1, z_f=1}^{x_{i_2}=|X|, z_f=|Z|}\\
      \ldots \\
      [\varphi_{f{i_{|f|}}}(x_{i_2}, z_f) - m_{f\rightarrow
        i_n}(x_{i_n}) +  b_{i_{|f|}}(x_{i_{|f|}})]_{x_{i_{|f|}}=1, z_f=1}^{x_{i_{|f|}}=|X|,z_f=|Z|}
    \end{matrix}
  \right].
\]
Then by Proposition \ref{propos:matrix_max}, we can recover the
maximization operation in \eqref{eq:bfi_reform} using an $\Ocal(\log_2|X|)$-layer neural network with at most $\Ocal(|X|^2\log_2|X|)$ hidden
units. After that, all the other operations are simple linear
operations, and they can be easily encoded in a neural-network
without adding any parameter. Thus we can construct an FGNN layer,
which produces factor features for each factor $f$ as follows
\[
  \left[
    \begin{matrix}
      [b_{f\rightarrow i_{1}}(z_f)]_{z_f=1}^{z_f=|Z|}\\
      [b_{f\rightarrow i_{2}}(z_f)]_{z_f=1}^{z_f=|Z|}\\
      \ldots \\
      [b_{f\rightarrow i_{|f|}}(z_f)]_{z_f=1}^{z_f=|Z|}
    \end{matrix}
  \right].
\]

Finally we constructed an FGNN to parameterize the operation in
\eqref{eq:to_z}, and this construction also proves Proposition
\ref{propos:first_layer} as follows. 

\begin{repproposition}{propos:first_layer}
  The operation in \eqref{eq:to_z} can be parameterized by one MPNN layer
  with $\Ocal(|X| \max_{c\in\Ccal}|\Zcal_c|$ hidden units followed by a
  $\Ocal(\log_2|X|)$-layer neural network with at most $\Ocal(|X|^2\log_2|X|)$ hidden
  units. 
\end{repproposition}

\paragraph{FGNN Layer to recover \eqref{eq:to_x}} Here we construct an
FGNN layer to parameterize \eqref{eq:reparam_max_z} and \eqref{eq:to_x} in order to prove
Proposition \ref{propos:second_layer}. Using the notation in this
section the operation in
\eqref{eq:to_x} can be reformulated as 
\begin{align*}
  m_{f\rightarrow i}(x_i) &\leftarrow \max_z[\varphi_{if}(x_i, z_f) +
  b_{c\rightarrow i}(z_f)]\notag\\
  b_i(x_i) &\leftarrow \theta_i(x_i)  + \sum_{f: i\in
              f}\max_z[\varphi_{if}(x_i, z_f) + b_{f\rightarrow i}(z_f)].
\end{align*}
In previous paragraph, the new factor feature 
\[
  \left[
    \begin{matrix}
      [b_{f\rightarrow i_{1}}(z_f)]_{z_f=1}^{z_f=|Z|}\\
      [b_{f\rightarrow i_{2}}(z_f)]_{z_f=1}^{z_f=|Z|}\\
      \ldots \\
      [b_{f\rightarrow i_{|f|}}(z_f)]_{z_f=1}^{z_f=|Z|}
    \end{matrix}
  \right].
\]
Considering the old factor feature
\[
  \left[
    \begin{matrix}
      [\varphi_{fi}(x_{i_1}, z_f)]_{x_{i_1}=1, z_f=1}^{x_{i_1}=|X|, z_f=|Z|}\\
      [\varphi_{fi}(x_{i_2}, z_f)]_{x_{i_2}=1, z_f=1}^{x_{i_2}=|X|, z_f=|Z|}\\
      \ldots \\
      [\varphi_{fi}(x_{i_{|f|}}, z_f)]_{x_{i_{|f|}}=1, z_f=1}^{x_{i_{|f|}}=|X|,z_f=|Z|}
    \end{matrix}
  \right],
\]
we can use \emph{broadcasted} addition between these two features to
get
\[
  \left[
    \begin{matrix}
      [b_{f\rightarrow i_{1}}(z_f) + \varphi_{fi}(x_{i_1}, z_f)]_{x_{i_1}=1, z_f=1}^{x_{i_1}=|X|, z_f=|Z|}\\
      [b_{f\rightarrow i_{2}}(z_f) + \varphi_{fi}(x_{i_2}, z_f)]_{x_{i_2}=1, z_f=1}^{x_{i_2}=|X|, z_f=|Z|}\\
      \ldots \\
      [b_{f\rightarrow i_{|f|}}(z_f) + \varphi_{fi}(x_{i_{|f|}}, z_f)]_{x_{i_{|f|}}=1, z_f=1}^{x_{i_{|f|}}=|X|,z_f=|Z|}
    \end{matrix}
  \right].
\]
After that we have an $|f|\times |X| \times |Z|$ feature tensor for
each factor $f\in \Fcal$. By \ref{propos:matrix_max}, a
$\Ocal(\log_2 |\Zcal|)$-layer neural network with at most
$\Ocal(|\Zcal|^2\log_2 |\Zcal|)$ parameters can be used to convert
the above feature to
\[
  \left[
      \begin{matrix}
        [m_{f\rightarrow i_1}(x_{i_1})]_{x_{i_1}=1}^{x_{i_1}=|X|}\\
        [m_{f\rightarrow i_2}(x_{i_2})]_{x_{i_2}=1}^{x_{i_2}=|X|}\\
        \cdots\\
        [m_{f\rightarrow i_{|f|}}(x_{i_{|f|}})]_{x_{i_{|f|}}=1}^{x_{i_{|f|}}=|X|}\\
        \end{matrix}
      \right]
    \leftarrow
  \left[
    \begin{matrix}
      [\max_{z_f}[b_{f\rightarrow i_{1}}(z_f) +
      \varphi_{fi}(x_{i_1}, z_f)] ]_{x_{i_1}=1}^{x_{i_1}=|X|}\\
      [\max_{z_f}[b_{f\rightarrow i_{2}}(z_f) + \varphi_{fi}(x_{i_2}, z_f)]]_{x_{i_2}=1}^{x_{i_2}=|X|}\\
      \ldots \\
      [\max_{z_f}[b_{f\rightarrow i_{|f|}}(z_f) + \varphi_{fi}(x_{i_{|f|}}, z_f)]]_{x_{i_{|f|}}=1}^{x_{i_{|f|}}=|X|}
    \end{matrix}
  \right].
\]
We will use this as the first part of our $\Mcal$ network. For the
second part, as  we need to parameterize the $\sum_{f: i\in
  f}\max_z[\varphi_{if}(x_i, z_f) + b_{c\rightarrow
  i}(z_f)]$ from feature $\max_z[\varphi_{if}(x_i, z_f) + b_{c\rightarrow
  i}(z_f)$, by Proposition \ref{propos:feature_sum}, it
will require another linear layer with $\Ocal(\max_{i
  \in\Vcal}\text{deg(i)}^2 |X|^2)$, where $\text{deg}(i) =
|\{f|f\in\Fcal, i \in f\}|$. After that, the $\Qcal$
network can be a simple identity mapping, and the FGNN
would produce updated messages  $m_{f\rightarrow i}(x_i) = \max_z[\varphi_{if}(x_i, z_f) + b_{c\rightarrow
  i}(z_f)]$ for each node. Adding these feature with the
initial node feature would results new node feature
$b_i(x_i)$. Thus by constructing a FGNN layer to parameterize
\eqref{eq:reparam_max_z} and \eqref{eq:to_x} we complete the proof of Proposition
\ref{propos:second_layer}.

\subsection{Example of Recovering Max Product Belief Propagation}
We provide a simple example that uses the proposed FGNN to recover Max Product Belief Propagation. Since the Max Product Belief Propagation can be viewed as a continuous mapping between the input log-potentials and the output ``beliefs'', and our FGNN is acutally a universal approximator for such mapping. In this part, we provide one parametrization of FGNN that can exactly recover Max Product Belief Propagation, but it may not be the only one or the optimal one. Our goal is to design a network that is capabale of recovering traditional inference procedures such as belief propagation, but by learning from data our approach may learn a better inference approach.

\begin{figure}[t]
  \centering
  \includegraphics[width=\textwidth]{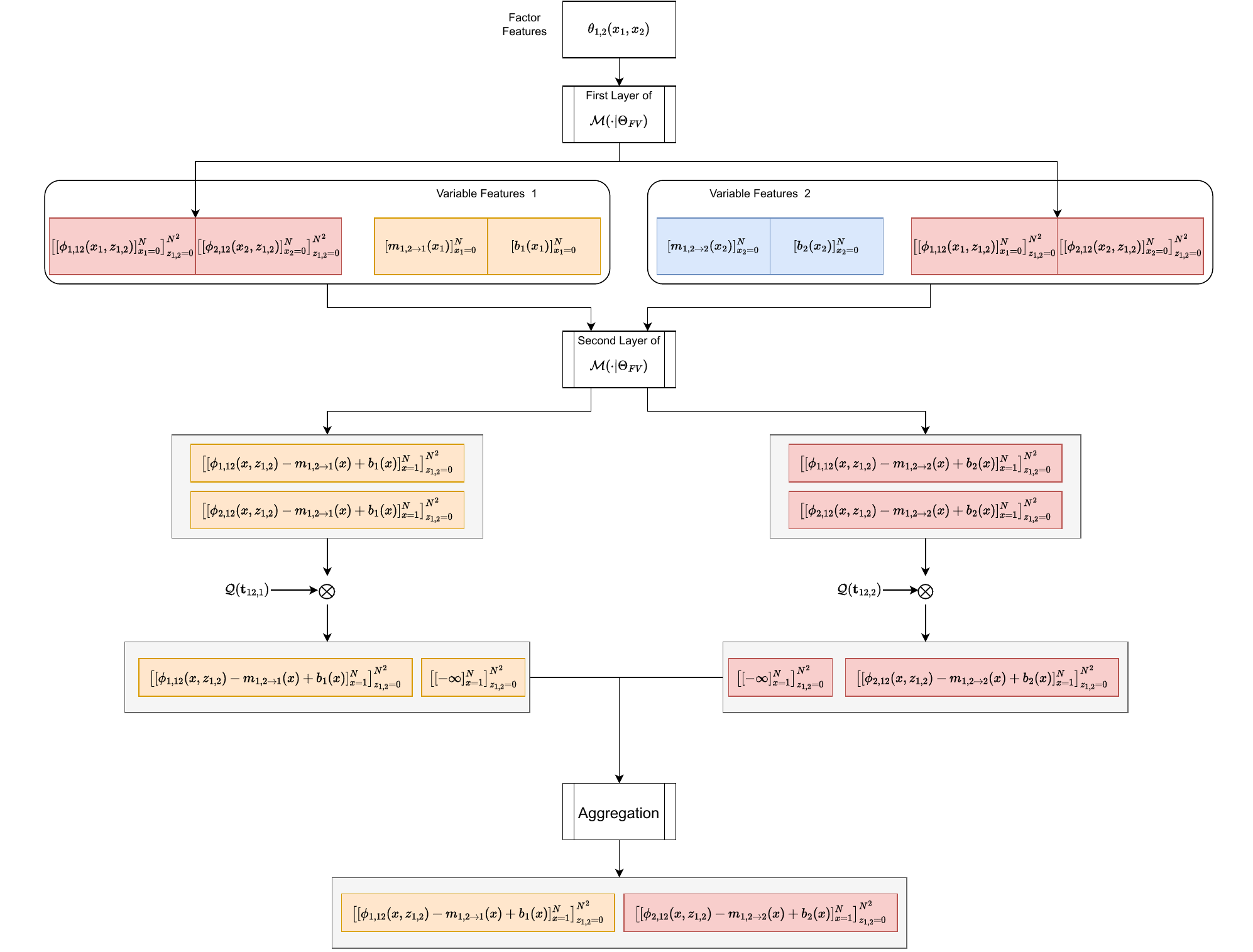}
  \caption{Example of Variable-to-Factor module that recovers the operations in \eqref{eq:to_z}.}
  \label{fig:recover_to_z_v_to_f}
\end{figure}

Let's consider a simple MAP inference problem over a simple graphical model as follows,
\begin{align}
  \label{eq:exam_map}
  \max_{\xb} \left[\theta_{1,2}(x_1, x_2) + \theta_{2,3}(x_2, x_3)\right],
\end{align}
where each variable $x_i\in \{0, 1, \ldots N - 1\}$, and the log-potentials are all real-valued functions. Then we show the most complicated procedure of \eqref{eq:decomposed}, that is \eqref{eq:to_z}, can be recovered by a Variable-to-Factor module. In such a Variable-to-Factor (shown in Figure \ref{fig:recover_to_z_v_to_f}) module, in the first layer of $\Mcal(\cdot|\Theta_{FV})$, the edge potentials are mapped into the decomposed log-potentials defined in Lemma \ref{lemma:decompose}. This operation only requires a linear transformation. Then the decomposed log-potentials will be concatenated as the input of the second layer of $\Mcal(\cdot|\Theta_{FV})$ (recall that the Variable-to-Factor requires both factor and variable feature as input), then by another linear transformation we can get the term inside the max-operation in \eqref{eq:to_z}, plus a redundant term with the same shape. Then the $\Qcal$ network, works as a selector, will set the redundant term to $-\infty$, then by the aggregation part, the redundant term will be filtered. Finally, by applying MLP to max over $x$ and do the summation, the factor $\{1,2\}$ can have a feature vector that consists of $b_{1,2\rightarrow 1}(z_c)$ and $b_{1,2\rightarrow 2}(z_c)$. 

\section{Experiments}

\subsection{Additional Information on MAP Inference over PGM}
\paragraph{Data}
We construct four datasets. All variables are binary. The instances start with a chain structure with unary potential on every node and pairwise potentials between consecutive nodes. A higher-order potential is then imposed to every node for the first three datasets. %

The node potentials are all randomly generated from the uniform distribution over $[0, 1]$. We use two kinds of pairwise potentials, one randomly generated (as in Table~\ref{tab:pws_uni}), the other encouraging two adjacent nodes to both take state $1$ (as in Table~\ref{tab:pws} and Table~\ref{tab:pws_ran}), i.e. the potential function gives high value to configuration $(1,1)$ and low value to all other configurations.
For example, in Dataset1, the potential value for $x_1$ to take the state 0 and $x_2$ to take the state 1 is 0.2; in Dataset3, the potential value for $x_1$ and $x_2$ to take the state 1 at the same time is sampled from a uniform distribution over [0, 2].

\begin{minipage}[t]{1.0\linewidth}
  \begin{minipage}{0.3\textwidth}
    \scriptsize
    \centering
    \setlength{\tabcolsep}{3pt}
    \begin{tabular}{l|cc}
      \toprule
      \thead{pairwise\\ potential}  &\thead{$x_2=0$} & \thead{$x_2=1$} \\
      \midrule
      \thead{$x_1=0$} & 0 & 0.1  \\
      \thead{$x_1=1$} & 0.2 & 1\\
      \bottomrule
    \end{tabular}
    \captionof{table}{\scriptsize Pairwise Potential for Dataset1}
    \label{tab:pws}
  \end{minipage}\hfil
  \begin{minipage}{0.3\textwidth}
    \scriptsize
    \centering
    \setlength{\tabcolsep}{3pt}
    \begin{tabular}{l|cc}
      \toprule
      \thead{pairwise\\ potential}  &\thead{$x_2=0$} & \thead{$x_2=1$} \\
      \midrule
      \thead{$x_1=0$} & U[0,1] & U[0,1]  \\
      \thead{$x_1=1$} & U[0,1] & U[0,1]\\
      \bottomrule
    \end{tabular}
    \captionof{table}{\scriptsize Pairwise Potential for Dataset2,4}
    \label{tab:pws_uni}
  \end{minipage}\hfil
  \begin{minipage}{0.3\textwidth}
    \scriptsize
    \centering
    \setlength{\tabcolsep}{3pt}
    \begin{tabular}{l|cc}
      \toprule
      \thead{pairwise\\ potential}  &\thead{$x_2=0$} & \thead{$x_2=1$} \\
      \midrule
      \thead{$x_1=0$} & 0 & 0  \\
      \thead{$x_1=1$} & 0 & U[0,2]\\
      \bottomrule
    \end{tabular}
    \captionof{table}{\scriptsize Pairwise Potential for Dataset3}
    \label{tab:pws_ran}
  \end{minipage}
\end{minipage}

For Dataset1,2,3, we additionally add the budget higher-order
potential \citep{martins2015ad} at every node; these potentials allow
at most $k$ of the 8 variables that are within their scope to take the
state 1. For the first two datasets, the value $k$ is set to 5; for
the third dataset, it is set to a random integer in
\{1,2,3,4,5,6,7,8\}. For Dataset4, there is no higher-order
potential. 

As a result of the constructions, different datasets have different
inputs for the FGNN; for each dataset, the inputs for each instance are the parameters
of the PGM that are not fixed. For Dataset1, only the node potentials
are not fixed, hence each input instance is a factor graph with the
randomly generated node potential added as the input node feature for
each variable node. Dataset2 and Dataset4 are similar in terms of the
input format, both including randomly generate node potentials as
variable node features and randomly generated pairwise 
potential parameters as the corresponding pairwise factor
node features. Finally, for Dataset3, the variable nodes, the pairwise
factor nodes and the high order factor nodes all have corresponding
input features.

\paragraph{Architecture} We use a multi-layer
factor graph neural network with architecture
\textsc{FGNN(64) - Res[FC(64) - FGNN(64) - FC(64)] - MLP(128)
  - Res[FC(64) - FGNN(64) - FC(128)] - FC(256) - Res[FC(256) - FGNN(64) -
  FC(256)] - FC(128) - Res[FC(128) - FGNN(64) - FC(128)] - FC(64) -
  Res[FC(64) - FGNN(64) - FC(64)] - FGNN(2)}. 
Here one FGNN($C_{\text{out}}$) is a FGNN layer with $C_{\text{out}}$
as output feature dimension with ReLU \citep{nair2010rectified} as activation. One FC($C_{\text{out}}$) is a fully connected layer with $C_{\text{out}}$ as output feature dimension and
ReLU as activation.  \text{Res[$\cdot$]} is a neural network with
residual link from its input to output; these additional architecture
components can assist learning.

\paragraph{Running Time}
We report the inference time of one instance and the training time of
one epoch for the synthetic datasets in
Table~\ref{tab:inference-time}. The results show that our method runs
in a reasonable amount of time. 
\begin{table*}[h]
  \scriptsize
  \centering
  \begin{tabular}{lccccccc}
    \toprule
    
    {($\mu$s)} &  {PointNet} & {DGCNN} &  { AD3 (exact/approx)}&{Max-Product}&{MPLP}&{MPNN}&{Ours}\\
    
    \midrule
    D1  & 45 (43) & 285 (107) & 5 / 5& 6 & 57 &131 (72) &144 (75)\\
    D2& -- & --  & 532 / 325 &1228 & 55 &131 (72)&341 (162)\\
    D3   & -- & --& 91092 / 1059 &4041& 55 &121 (74) & 382 (170)\\
    \bottomrule
  \end{tabular}
  \caption{\small Inference time in microseconds of one instance on synthetic datasets and GPU training time of one epoch in milliseconds (in bracket) for applicable methods.}
  \label{tab:inference-time}
\end{table*}

\subsection{Implementation details on MAP Solvers} In the experiment,
the AD3 code is from the official code
repo \footnote{\url{https://github.com/andre-martins/AD3}}, which comes
with a python interface. For Max-Product algorithm, we use the
implementation from libdai and convert the budget higher potential as a
table function. For the MPLP algorithm, we implemented it in C++ to
directly support the budget higher-order potential. The re-implemented
version is compared with the original version
\footnote{\url{https://people.csail.mit.edu/dsontag/code/mplp_ver2.tgz}},
and its performance is better than the original one in our experiment. So we provide the result of the re-implemented version.

\subsection{Dataset Generation and Training Details of LDPC decoding}

\paragraph{Data}
Each instance of training/evaluation data is generated as follows:
\begin{algorithm}
  \renewcommand{\algorithmicensure}{\textbf{Output:}}
  
  \caption{Data Generation for LDPC decoding}
  \begin{algorithmic}
    \ENSURE{$\yb$: a $96$-bit noisy signal; $\text{SNR}_{dB}$:
      signal-to-noise ratio, a scalar}\;
    \STATE Uniformly sample a 48-bit binary signal $\xb$, where for
    each $0 < i \leqslant 48$, $P(x_i=1)=P(x_i=0)=0.5$\;
    \STATE Encode $\xb$ using the ``96.3.963'' scheme \citep{mackay2009david} to get a
    96-bit signal $\yb$\;
    \STATE  sample $\text{SNR}_{dB} \in
    \{0, 1, ,2, 3, 4\}$ and   $\sigma_b \in
    \{0, 1, ,2 3, 4, 5\}$ uniformly\;
    
    \STATE For each $0 < i \leqslant 96$,uniformly, sample
    \begin{itemize}
    \item $\eta_i \in
      \Ucal(0, 1)$,
    \item  $n_i \in \Ncal(0, \sigma^2)$ s.t. $\text{SNR}_{dB}=20\log_{10}{1/\sigma}$
    \item $z_i\in \Ncal(0, \sigma_b^2)$ 
    \end{itemize}
    \STATE Set noisy signal $\tilde \yb$ to
    \begin{itemize}
    \item $\tilde y_i = y_i + n_i + \mathbb{I}(\eta_i \leqslant 0.05) z_i$
    \end{itemize}
    
  \end{algorithmic}
\end{algorithm}

During the training of FGNN, the node feature include the
noisy signal $\tilde \yb$ and the signal-to-noise ratio
$\text{SNR}_{dB}$. For
FGNN, for each factor $f$, the vector $[\tilde y_i]_{i\in f}$ is provided as
feature vector. Meanwhile, for each edge from factor node $f$ to one of its
variable node $i$, the factor feature and the variable node feature
are put together to get the edge feature.

\paragraph{Architecture}
In our FGNN, every layer
share the same $\Qcal$ network, which is 2-layer network as follows
\textsc{MLP(64)-MLP(4)}. Here the first layer comes with a ReLU activation function and the second layer is with no activation function.

The overall structure of our FGNN is as follows
\textsc{Input  - 
  Res[FC(64) - FGNN(64) - FC(64)]   -  Res[FC(64) - FGNN(64) - FC(64)]
  - FC(64) - FGNN(64) - FC(128) - FC(256) - FGNN(128) - FC(256) - 
  - Res[FC(256) - FGNN(128) - FC(256)]  - FC(128) - FGNN(128) - FC(128)  - 
  FC(64) - FGNN(64) - FC(64)  -  Res[FC(64) - FGNN(64) - FC(64)] - FC(128) - 
  FC(128) - FC(1)}. In the network, a batch-normalization layer and a
ReLU activation function is after each FC layer and FGNN layer except
for the last FC layer.

\subsection{Additional experiments on Molecular data}
\subsubsection{Improvement over MPNN when distance information is excluded}

In the main paper, we reported the improved performance of FGNN over MPNN on Alchemy dataset. The FGNN is built on MPNN as backend feature extractor. Therefore, its improved performance is the result of capturing dependencies not modeled by MPNN. This begs the question: can higher-order message passing capture more information when the backend MPNN module is further constrained. For this, we limit the input to MPNN module and see if FGNN can further improve its gain with respect to MPNN. %

One of the main reasons for the superior performance of MPNN on molecular datasets is that it can capture the 3D geometric structure of the molecule~\citep{chen2019alchemy,gilmer2017neural}. MPNN is provided with edge features which include bond type and spatial distance between the pair of atoms. Further, it operates on a complete graph where extra \emph{virtual edges} are added between every pair of atoms with no bond. The edge feature for such \emph{virtual edges} contains the spatial distance between the pair of atoms. Consequently, MPNN can capture 3D geometric structure of the molecule with such a complete graph. 

In the following experiments, we evaluate whether higher-order message passing can help capture structure of the molecule in the absence of pairwise distance edge features. %
We do include 3D atom positions in the node features and hence the information about the geometric structure of the molecule is indirectly provided. Based on the pairwise distance feature, we divide the experimental setup in three categories .
\begin{itemize}[{leftmargin=*}]
\item \textbf{Sparse graph without distance:} Input graph is a sparse graph i.e., edge exists only if bond exists between atoms, with edge features containing the bond type without the  distance between the pair of atoms. 
\item \textbf{Sparse graph with distance:} Input graph is a sparse graph with edge features containing bond type and distance between the pair of atoms.%
\item \textbf{Complete graph with distance:} Input graph is a complete graph with extra \emph{virtual edges} containing distance information between pair of atoms in the edge. This setup is the standard MPNN model.
\end{itemize}
In all three cases, regardless of input to the MPNN module, the higher-order message passing works only on the sparse graph.
\begin{table}[ht]
  \footnotesize
  \caption{Comparison  of FGNN with MPNN on Alchemy dataset with/without distance information}
  \label{table-suppl-alchemy}
  \centering
  \begin{tabular}{lccccccccc}
    \toprule
    \multirow{2}{*}{Target} & 
                              \multicolumn{3}{c}{Sparse graph without distance}& 
                                                                                 \multicolumn{3}{c}{Sparse graph with distance}& 
                                                                                                                                 \multicolumn{3}{c}{Complete graph with distance} \\
    \cmidrule(lr){2-4} \cmidrule(lr){5-7} \cmidrule(lr){8-10}
                            & MPNN & FGNN & Gain(\%) & MPNN & FGNN & Gain(\%) & MPNN & FGNN & Gain(\%) \\
    \toprule
    $\mu$                 & 0.3546              & 0.3071              & 13.39           & 0.3655              & 0.3012              & 17.60           & 0.1026     & 0.1041               & -1.41         \\
    $\alpha$              & 0.1971              & 0.0873              & 55.68  & 0.1639              & 0.0888              & 45.80         & 0.0558              & 0.0451              & 19.06          \\
    $\epsilon_{homo}$     & 0.1723              & 0.1281              & 25.66   & 0.1539              & 0.1220              & 20.73         & 0.1151              & 0.1005              & 12.75           \\
    $\epsilon_{lumo}$     & 0.1280              & 0.0905              & 29.31     & 0.1120              & 0.0847              & 24.37         & 0.0817              & 0.0664     & 18.74          \\
    $\Delta_{\epsilon}$   & 0.1266              & 0.0902              & 28.75     & 0.1084              & 0.0854              & 21.18         & 0.0832              & 0.0692     & 16.88          \\
    $\langle R^2 \rangle$ & 0.1439              & 0.0629     & 56.31     & 0.2926              & 0.0629     & 78.50        & 0.0271              & 0.0099     & 63.47          \\
    $ZPVE$                & 0.1388              & 0.0412              & 70.28    & 0.1020              & 0.0399              & 60.80         & 0.0259              & 0.0115              & 55.42          \\
    $U_0$                 & 0.0806              & 0.0206              & 74.39         & 0.0606              & 0.0236              & 61.05         & 0.0131               & 0.0045              & 65.80          \\
    $U$                   & 0.0806              & 0.0206              & 74.37      & 0.0606              & 0.0236              & 61.05         & 0.0131              & 0.0045              & 65.90          \\
    $H$                   & 0.0806              & 0.0209              & 74.07                & 0.0606              & 0.0236              & 60.98          & 0.0131              & 0.0045              & 65.77          \\
    $G$                   & 0.0806              & 0.0208              & 74.13           & 0.0606              & 0.0236              & 61.02         & 0.0131              & 0.0045              & 65.77          \\
    $C_v$                 & 0.2177              & 0.0913              & 58.04                  & 0.1729              & 0.0931              & 46.13         & 0.0559               & 0.0481              & 13.93          \\
    \toprule
    MAE & 0.1501         & 0.0818             & \textbf{45.50}   & 0.1428         & 0.0810             & \textbf{43.24}         & 0.0499        & 0.0394        & \textbf{21.18}         
  \end{tabular}
\end{table}

Results in Table~\ref{table-suppl-alchemy} shows that the margin of improvement of FGNN over MPNN is significantly higher when pairwise distance feature is not included in the graph. This suggests MPNN is not able to sufficiently capture the 3D molecular shape in both the cases where sparse graph is used, In such scenarios, capturing higher-order structures with FGNN is more helpful in reducing the MAE. Furthermore, results of both the cases are similar when sparse graph is used and inclusion of pairwise distance in the edge feature does not lead to significant performance gains. Following this, it can be inferred that bond types are indicative of distance between the atoms as well. 

\subsubsection{Results of ablation models on QM9 dataset:}
In the main paper, we considered ablation models of FGNN based on conditioning of factor parameters. These models are CAT (central atom type), BT (bond type), CABT (central atom and bond type) and CABTA (central atom, bond type and neighbouring atom type). We reported results on Alchemy dataset where we found that conditioning on bond type was sufficient for good performance. To further verify the results, we evaluate the ablation models on QM9 dataset.

Results in Table~\ref{table-qm9_sup_abl} show that unlike Alchemy dataset, CABT model performs better in almost all the targets on QM9 dataset. This perhaps suggests that in QM9 dataset, higher-order constraints are more centered around the atom and are better captured by having separate parameters for different central atom types. Collectively, the ablation study on QM9 and Alchemy datasets suggests that conditioning the parameters on neighbouring atom type (CABTA) is not helpful and only increases the paramter size. It is sufficient if edge type and central atom type information is directly captured in the model. 
\begin{table}[thbp]
  \footnotesize
  \caption{Ablation models regression results on QM9 dataset}
  \label{table-qm9_sup_abl}
  \centering
  \begin{tabular}{lrrrr}
    \toprule
    \multirow{2}{*}{Target} &  
                              \multicolumn{4}{c}{FGNN Ablation Models} \\
    \cmidrule(lr){2-5} %
                            & CAT & BT & CABT & CABTA \\
    \midrule
    $\mu$ & \textbf{0.088}     &0.095               & 0.092              & 0.097                 \\
    $\alpha$ &  0.192               & 0.204              & \textbf{0.183}     & 0.185                \\
    $\epsilon_{homo}$        & 0.0021               &0.00217              &  \textbf{0.00199}     & 0.00203               \\
    $\epsilon_{lumo}$ &  0.00212              &0.0022               &  \textbf{0.00205}     & 0.00208               \\
    $\Delta_{\epsilon}$  & 0.0029               &0.00301              &  \textbf{0.00279}     & 0.00286               \\
    $\langle R^2 \rangle$     & 2.88              & 3.01              &\textbf{2.81}     & 3.17                \\
    $ZPVE$ &  0.00021              & 0.00021              &  \textbf{0.00018}     & 0.00022                \\
    $U_0$ & 0.1283              & 0.0995              &   \textbf{0.0907}     & 0.1649            \\
    $U$ &  0.1283              & 0.1029              &  \textbf{0.0907}     & 0.1650               \\
    $H$ & 0.1283              & 0.1008              & \textbf{0.0906}     & 0.1649                \\
    $G$ &  0.1283              & 0.1011              &  \textbf{0.0906}     & 0.1651                \\
    $C_v$ &  0.0868              & 0.0900              & \textbf{0.0847}     & 0.0872                 \\
    \midrule
    MAE & 0.3141 & 0.3183 &  \textbf{0.2947} & 0.3512 
  \end{tabular}
\end{table}

\vskip 0.2in
\bibliography{reference}

\end{document}

%% file: figure/ocr_a.pgf
%% Creator: Matplotlib, PGF backend
%%
%% To include the figure in your LaTeX document, write
%%   \input{<filename>.pgf}
%%
%% Make sure the required packages are loaded in your preamble
%%   \usepackage{pgf}
%%
%% Figures using additional raster images can only be included by \input if
%% they are in the same directory as the main LaTeX file. For loading figures
%% from other directories you can use the `import` package
%%   \usepackage{import}
%% and then include the figures with
%%   \import{<path to file>}{<filename>.pgf}
%%
%% Matplotlib used the following preamble
%%
\begingroup%
\makeatletter%
\begin{pgfpicture}%
\pgfpathrectangle{\pgfpointorigin}{\pgfqpoint{8.000000in}{6.000000in}}%
\pgfusepath{use as bounding box, clip}%
\begin{pgfscope}%
\pgfsetbuttcap%
\pgfsetmiterjoin%
\definecolor{currentfill}{rgb}{1.000000,1.000000,1.000000}%
\pgfsetfillcolor{currentfill}%
\pgfsetlinewidth{0.000000pt}%
\definecolor{currentstroke}{rgb}{1.000000,1.000000,1.000000}%
\pgfsetstrokecolor{currentstroke}%
\pgfsetdash{}{0pt}%
\pgfpathmoveto{\pgfqpoint{0.000000in}{0.000000in}}%
\pgfpathlineto{\pgfqpoint{8.000000in}{0.000000in}}%
\pgfpathlineto{\pgfqpoint{8.000000in}{6.000000in}}%
\pgfpathlineto{\pgfqpoint{0.000000in}{6.000000in}}%
\pgfpathclose%
\pgfusepath{fill}%
\end{pgfscope}%
\begin{pgfscope}%
\pgfsetbuttcap%
\pgfsetmiterjoin%
\definecolor{currentfill}{rgb}{1.000000,1.000000,1.000000}%
\pgfsetfillcolor{currentfill}%
\pgfsetlinewidth{0.000000pt}%
\definecolor{currentstroke}{rgb}{0.000000,0.000000,0.000000}%
\pgfsetstrokecolor{currentstroke}%
\pgfsetstrokeopacity{0.000000}%
\pgfsetdash{}{0pt}%
\pgfpathmoveto{\pgfqpoint{1.000000in}{0.660000in}}%
\pgfpathlineto{\pgfqpoint{7.200000in}{0.660000in}}%
\pgfpathlineto{\pgfqpoint{7.200000in}{5.280000in}}%
\pgfpathlineto{\pgfqpoint{1.000000in}{5.280000in}}%
\pgfpathclose%
\pgfusepath{fill}%
\end{pgfscope}%
\begin{pgfscope}%
\pgfpathrectangle{\pgfqpoint{1.000000in}{0.660000in}}{\pgfqpoint{6.200000in}{4.620000in}}%
\pgfusepath{clip}%
\pgfsetbuttcap%
\pgfsetmiterjoin%
\definecolor{currentfill}{rgb}{0.121569,0.466667,0.705882}%
\pgfsetfillcolor{currentfill}%
\pgfsetlinewidth{0.000000pt}%
\definecolor{currentstroke}{rgb}{0.000000,0.000000,0.000000}%
\pgfsetstrokecolor{currentstroke}%
\pgfsetstrokeopacity{0.000000}%
\pgfsetdash{}{0pt}%
\pgfpathmoveto{\pgfqpoint{1.281818in}{-17.820000in}}%
\pgfpathlineto{\pgfqpoint{2.069697in}{-17.820000in}}%
\pgfpathlineto{\pgfqpoint{2.069697in}{1.669470in}}%
\pgfpathlineto{\pgfqpoint{1.281818in}{1.669470in}}%
\pgfpathclose%
\pgfusepath{fill}%
\end{pgfscope}%
\begin{pgfscope}%
\pgfpathrectangle{\pgfqpoint{1.000000in}{0.660000in}}{\pgfqpoint{6.200000in}{4.620000in}}%
\pgfusepath{clip}%
\pgfsetbuttcap%
\pgfsetmiterjoin%
\definecolor{currentfill}{rgb}{0.121569,0.466667,0.705882}%
\pgfsetfillcolor{currentfill}%
\pgfsetlinewidth{0.000000pt}%
\definecolor{currentstroke}{rgb}{0.000000,0.000000,0.000000}%
\pgfsetstrokecolor{currentstroke}%
\pgfsetstrokeopacity{0.000000}%
\pgfsetdash{}{0pt}%
\pgfpathmoveto{\pgfqpoint{2.493939in}{-17.820000in}}%
\pgfpathlineto{\pgfqpoint{3.281818in}{-17.820000in}}%
\pgfpathlineto{\pgfqpoint{3.281818in}{3.427380in}}%
\pgfpathlineto{\pgfqpoint{2.493939in}{3.427380in}}%
\pgfpathclose%
\pgfusepath{fill}%
\end{pgfscope}%
\begin{pgfscope}%
\pgfpathrectangle{\pgfqpoint{1.000000in}{0.660000in}}{\pgfqpoint{6.200000in}{4.620000in}}%
\pgfusepath{clip}%
\pgfsetbuttcap%
\pgfsetmiterjoin%
\definecolor{currentfill}{rgb}{0.121569,0.466667,0.705882}%
\pgfsetfillcolor{currentfill}%
\pgfsetlinewidth{0.000000pt}%
\definecolor{currentstroke}{rgb}{0.000000,0.000000,0.000000}%
\pgfsetstrokecolor{currentstroke}%
\pgfsetstrokeopacity{0.000000}%
\pgfsetdash{}{0pt}%
\pgfpathmoveto{\pgfqpoint{3.706061in}{-17.820000in}}%
\pgfpathlineto{\pgfqpoint{4.493939in}{-17.820000in}}%
\pgfpathlineto{\pgfqpoint{4.493939in}{4.164270in}}%
\pgfpathlineto{\pgfqpoint{3.706061in}{4.164270in}}%
\pgfpathclose%
\pgfusepath{fill}%
\end{pgfscope}%
\begin{pgfscope}%
\pgfpathrectangle{\pgfqpoint{1.000000in}{0.660000in}}{\pgfqpoint{6.200000in}{4.620000in}}%
\pgfusepath{clip}%
\pgfsetbuttcap%
\pgfsetmiterjoin%
\definecolor{currentfill}{rgb}{0.121569,0.466667,0.705882}%
\pgfsetfillcolor{currentfill}%
\pgfsetlinewidth{0.000000pt}%
\definecolor{currentstroke}{rgb}{0.000000,0.000000,0.000000}%
\pgfsetstrokecolor{currentstroke}%
\pgfsetstrokeopacity{0.000000}%
\pgfsetdash{}{0pt}%
\pgfpathmoveto{\pgfqpoint{4.918182in}{-17.820000in}}%
\pgfpathlineto{\pgfqpoint{5.706061in}{-17.820000in}}%
\pgfpathlineto{\pgfqpoint{5.706061in}{4.344450in}}%
\pgfpathlineto{\pgfqpoint{4.918182in}{4.344450in}}%
\pgfpathclose%
\pgfusepath{fill}%
\end{pgfscope}%
\begin{pgfscope}%
\pgfpathrectangle{\pgfqpoint{1.000000in}{0.660000in}}{\pgfqpoint{6.200000in}{4.620000in}}%
\pgfusepath{clip}%
\pgfsetbuttcap%
\pgfsetmiterjoin%
\definecolor{currentfill}{rgb}{0.121569,0.466667,0.705882}%
\pgfsetfillcolor{currentfill}%
\pgfsetlinewidth{0.000000pt}%
\definecolor{currentstroke}{rgb}{0.000000,0.000000,0.000000}%
\pgfsetstrokecolor{currentstroke}%
\pgfsetstrokeopacity{0.000000}%
\pgfsetdash{}{0pt}%
\pgfpathmoveto{\pgfqpoint{6.130303in}{-17.820000in}}%
\pgfpathlineto{\pgfqpoint{6.918182in}{-17.820000in}}%
\pgfpathlineto{\pgfqpoint{6.918182in}{4.307490in}}%
\pgfpathlineto{\pgfqpoint{6.130303in}{4.307490in}}%
\pgfpathclose%
\pgfusepath{fill}%
\end{pgfscope}%
\begin{pgfscope}%
\pgfsetbuttcap%
\pgfsetroundjoin%
\definecolor{currentfill}{rgb}{0.000000,0.000000,0.000000}%
\pgfsetfillcolor{currentfill}%
\pgfsetlinewidth{0.803000pt}%
\definecolor{currentstroke}{rgb}{0.000000,0.000000,0.000000}%
\pgfsetstrokecolor{currentstroke}%
\pgfsetdash{}{0pt}%
\pgfsys@defobject{currentmarker}{\pgfqpoint{0.000000in}{-0.048611in}}{\pgfqpoint{0.000000in}{0.000000in}}{%
\pgfpathmoveto{\pgfqpoint{0.000000in}{0.000000in}}%
\pgfpathlineto{\pgfqpoint{0.000000in}{-0.048611in}}%
\pgfusepath{stroke,fill}%
}%
\begin{pgfscope}%
\pgfsys@transformshift{1.675758in}{0.660000in}%
\pgfsys@useobject{currentmarker}{}%
\end{pgfscope}%
\end{pgfscope}%
\begin{pgfscope}%
\definecolor{textcolor}{rgb}{0.000000,0.000000,0.000000}%
\pgfsetstrokecolor{textcolor}%
\pgfsetfillcolor{textcolor}%
\pgftext[x=1.675758in,y=0.562778in,,top]{\color{textcolor}\rmfamily\fontsize{18.000000}{21.600000}\selectfont 0}%
\end{pgfscope}%
\begin{pgfscope}%
\pgfsetbuttcap%
\pgfsetroundjoin%
\definecolor{currentfill}{rgb}{0.000000,0.000000,0.000000}%
\pgfsetfillcolor{currentfill}%
\pgfsetlinewidth{0.803000pt}%
\definecolor{currentstroke}{rgb}{0.000000,0.000000,0.000000}%
\pgfsetstrokecolor{currentstroke}%
\pgfsetdash{}{0pt}%
\pgfsys@defobject{currentmarker}{\pgfqpoint{0.000000in}{-0.048611in}}{\pgfqpoint{0.000000in}{0.000000in}}{%
\pgfpathmoveto{\pgfqpoint{0.000000in}{0.000000in}}%
\pgfpathlineto{\pgfqpoint{0.000000in}{-0.048611in}}%
\pgfusepath{stroke,fill}%
}%
\begin{pgfscope}%
\pgfsys@transformshift{2.887879in}{0.660000in}%
\pgfsys@useobject{currentmarker}{}%
\end{pgfscope}%
\end{pgfscope}%
\begin{pgfscope}%
\definecolor{textcolor}{rgb}{0.000000,0.000000,0.000000}%
\pgfsetstrokecolor{textcolor}%
\pgfsetfillcolor{textcolor}%
\pgftext[x=2.887879in,y=0.562778in,,top]{\color{textcolor}\rmfamily\fontsize{18.000000}{21.600000}\selectfont 1}%
\end{pgfscope}%
\begin{pgfscope}%
\pgfsetbuttcap%
\pgfsetroundjoin%
\definecolor{currentfill}{rgb}{0.000000,0.000000,0.000000}%
\pgfsetfillcolor{currentfill}%
\pgfsetlinewidth{0.803000pt}%
\definecolor{currentstroke}{rgb}{0.000000,0.000000,0.000000}%
\pgfsetstrokecolor{currentstroke}%
\pgfsetdash{}{0pt}%
\pgfsys@defobject{currentmarker}{\pgfqpoint{0.000000in}{-0.048611in}}{\pgfqpoint{0.000000in}{0.000000in}}{%
\pgfpathmoveto{\pgfqpoint{0.000000in}{0.000000in}}%
\pgfpathlineto{\pgfqpoint{0.000000in}{-0.048611in}}%
\pgfusepath{stroke,fill}%
}%
\begin{pgfscope}%
\pgfsys@transformshift{4.100000in}{0.660000in}%
\pgfsys@useobject{currentmarker}{}%
\end{pgfscope}%
\end{pgfscope}%
\begin{pgfscope}%
\definecolor{textcolor}{rgb}{0.000000,0.000000,0.000000}%
\pgfsetstrokecolor{textcolor}%
\pgfsetfillcolor{textcolor}%
\pgftext[x=4.100000in,y=0.562778in,,top]{\color{textcolor}\rmfamily\fontsize{18.000000}{21.600000}\selectfont 2}%
\end{pgfscope}%
\begin{pgfscope}%
\pgfsetbuttcap%
\pgfsetroundjoin%
\definecolor{currentfill}{rgb}{0.000000,0.000000,0.000000}%
\pgfsetfillcolor{currentfill}%
\pgfsetlinewidth{0.803000pt}%
\definecolor{currentstroke}{rgb}{0.000000,0.000000,0.000000}%
\pgfsetstrokecolor{currentstroke}%
\pgfsetdash{}{0pt}%
\pgfsys@defobject{currentmarker}{\pgfqpoint{0.000000in}{-0.048611in}}{\pgfqpoint{0.000000in}{0.000000in}}{%
\pgfpathmoveto{\pgfqpoint{0.000000in}{0.000000in}}%
\pgfpathlineto{\pgfqpoint{0.000000in}{-0.048611in}}%
\pgfusepath{stroke,fill}%
}%
\begin{pgfscope}%
\pgfsys@transformshift{5.312121in}{0.660000in}%
\pgfsys@useobject{currentmarker}{}%
\end{pgfscope}%
\end{pgfscope}%
\begin{pgfscope}%
\definecolor{textcolor}{rgb}{0.000000,0.000000,0.000000}%
\pgfsetstrokecolor{textcolor}%
\pgfsetfillcolor{textcolor}%
\pgftext[x=5.312121in,y=0.562778in,,top]{\color{textcolor}\rmfamily\fontsize{18.000000}{21.600000}\selectfont 3}%
\end{pgfscope}%
\begin{pgfscope}%
\pgfsetbuttcap%
\pgfsetroundjoin%
\definecolor{currentfill}{rgb}{0.000000,0.000000,0.000000}%
\pgfsetfillcolor{currentfill}%
\pgfsetlinewidth{0.803000pt}%
\definecolor{currentstroke}{rgb}{0.000000,0.000000,0.000000}%
\pgfsetstrokecolor{currentstroke}%
\pgfsetdash{}{0pt}%
\pgfsys@defobject{currentmarker}{\pgfqpoint{0.000000in}{-0.048611in}}{\pgfqpoint{0.000000in}{0.000000in}}{%
\pgfpathmoveto{\pgfqpoint{0.000000in}{0.000000in}}%
\pgfpathlineto{\pgfqpoint{0.000000in}{-0.048611in}}%
\pgfusepath{stroke,fill}%
}%
\begin{pgfscope}%
\pgfsys@transformshift{6.524242in}{0.660000in}%
\pgfsys@useobject{currentmarker}{}%
\end{pgfscope}%
\end{pgfscope}%
\begin{pgfscope}%
\definecolor{textcolor}{rgb}{0.000000,0.000000,0.000000}%
\pgfsetstrokecolor{textcolor}%
\pgfsetfillcolor{textcolor}%
\pgftext[x=6.524242in,y=0.562778in,,top]{\color{textcolor}\rmfamily\fontsize{18.000000}{21.600000}\selectfont 4}%
\end{pgfscope}%
\begin{pgfscope}%
\definecolor{textcolor}{rgb}{0.000000,0.000000,0.000000}%
\pgfsetstrokecolor{textcolor}%
\pgfsetfillcolor{textcolor}%
\pgftext[x=4.100000in,y=0.293874in,,top]{\color{textcolor}\rmfamily\fontsize{24.000000}{28.800000}\selectfont Maximum order of the factors}%
\end{pgfscope}%
\begin{pgfscope}%
\pgfpathrectangle{\pgfqpoint{1.000000in}{0.660000in}}{\pgfqpoint{6.200000in}{4.620000in}}%
\pgfusepath{clip}%
\pgfsetrectcap%
\pgfsetroundjoin%
\pgfsetlinewidth{0.803000pt}%
\definecolor{currentstroke}{rgb}{0.690196,0.690196,0.690196}%
\pgfsetstrokecolor{currentstroke}%
\pgfsetdash{}{0pt}%
\pgfpathmoveto{\pgfqpoint{1.000000in}{0.660000in}}%
\pgfpathlineto{\pgfqpoint{7.200000in}{0.660000in}}%
\pgfusepath{stroke}%
\end{pgfscope}%
\begin{pgfscope}%
\pgfsetbuttcap%
\pgfsetroundjoin%
\definecolor{currentfill}{rgb}{0.000000,0.000000,0.000000}%
\pgfsetfillcolor{currentfill}%
\pgfsetlinewidth{0.803000pt}%
\definecolor{currentstroke}{rgb}{0.000000,0.000000,0.000000}%
\pgfsetstrokecolor{currentstroke}%
\pgfsetdash{}{0pt}%
\pgfsys@defobject{currentmarker}{\pgfqpoint{-0.048611in}{0.000000in}}{\pgfqpoint{0.000000in}{0.000000in}}{%
\pgfpathmoveto{\pgfqpoint{0.000000in}{0.000000in}}%
\pgfpathlineto{\pgfqpoint{-0.048611in}{0.000000in}}%
\pgfusepath{stroke,fill}%
}%
\begin{pgfscope}%
\pgfsys@transformshift{1.000000in}{0.660000in}%
\pgfsys@useobject{currentmarker}{}%
\end{pgfscope}%
\end{pgfscope}%
\begin{pgfscope}%
\definecolor{textcolor}{rgb}{0.000000,0.000000,0.000000}%
\pgfsetstrokecolor{textcolor}%
\pgfsetfillcolor{textcolor}%
\pgftext[x=0.554634in,y=0.590556in,left,base]{\color{textcolor}\rmfamily\fontsize{14.000000}{16.800000}\selectfont \(\displaystyle 80.0\)}%
\end{pgfscope}%
\begin{pgfscope}%
\pgfpathrectangle{\pgfqpoint{1.000000in}{0.660000in}}{\pgfqpoint{6.200000in}{4.620000in}}%
\pgfusepath{clip}%
\pgfsetrectcap%
\pgfsetroundjoin%
\pgfsetlinewidth{0.803000pt}%
\definecolor{currentstroke}{rgb}{0.690196,0.690196,0.690196}%
\pgfsetstrokecolor{currentstroke}%
\pgfsetdash{}{0pt}%
\pgfpathmoveto{\pgfqpoint{1.000000in}{1.237500in}}%
\pgfpathlineto{\pgfqpoint{7.200000in}{1.237500in}}%
\pgfusepath{stroke}%
\end{pgfscope}%
\begin{pgfscope}%
\pgfsetbuttcap%
\pgfsetroundjoin%
\definecolor{currentfill}{rgb}{0.000000,0.000000,0.000000}%
\pgfsetfillcolor{currentfill}%
\pgfsetlinewidth{0.803000pt}%
\definecolor{currentstroke}{rgb}{0.000000,0.000000,0.000000}%
\pgfsetstrokecolor{currentstroke}%
\pgfsetdash{}{0pt}%
\pgfsys@defobject{currentmarker}{\pgfqpoint{-0.048611in}{0.000000in}}{\pgfqpoint{0.000000in}{0.000000in}}{%
\pgfpathmoveto{\pgfqpoint{0.000000in}{0.000000in}}%
\pgfpathlineto{\pgfqpoint{-0.048611in}{0.000000in}}%
\pgfusepath{stroke,fill}%
}%
\begin{pgfscope}%
\pgfsys@transformshift{1.000000in}{1.237500in}%
\pgfsys@useobject{currentmarker}{}%
\end{pgfscope}%
\end{pgfscope}%
\begin{pgfscope}%
\definecolor{textcolor}{rgb}{0.000000,0.000000,0.000000}%
\pgfsetstrokecolor{textcolor}%
\pgfsetfillcolor{textcolor}%
\pgftext[x=0.554634in,y=1.168056in,left,base]{\color{textcolor}\rmfamily\fontsize{14.000000}{16.800000}\selectfont \(\displaystyle 82.5\)}%
\end{pgfscope}%
\begin{pgfscope}%
\pgfpathrectangle{\pgfqpoint{1.000000in}{0.660000in}}{\pgfqpoint{6.200000in}{4.620000in}}%
\pgfusepath{clip}%
\pgfsetrectcap%
\pgfsetroundjoin%
\pgfsetlinewidth{0.803000pt}%
\definecolor{currentstroke}{rgb}{0.690196,0.690196,0.690196}%
\pgfsetstrokecolor{currentstroke}%
\pgfsetdash{}{0pt}%
\pgfpathmoveto{\pgfqpoint{1.000000in}{1.815000in}}%
\pgfpathlineto{\pgfqpoint{7.200000in}{1.815000in}}%
\pgfusepath{stroke}%
\end{pgfscope}%
\begin{pgfscope}%
\pgfsetbuttcap%
\pgfsetroundjoin%
\definecolor{currentfill}{rgb}{0.000000,0.000000,0.000000}%
\pgfsetfillcolor{currentfill}%
\pgfsetlinewidth{0.803000pt}%
\definecolor{currentstroke}{rgb}{0.000000,0.000000,0.000000}%
\pgfsetstrokecolor{currentstroke}%
\pgfsetdash{}{0pt}%
\pgfsys@defobject{currentmarker}{\pgfqpoint{-0.048611in}{0.000000in}}{\pgfqpoint{0.000000in}{0.000000in}}{%
\pgfpathmoveto{\pgfqpoint{0.000000in}{0.000000in}}%
\pgfpathlineto{\pgfqpoint{-0.048611in}{0.000000in}}%
\pgfusepath{stroke,fill}%
}%
\begin{pgfscope}%
\pgfsys@transformshift{1.000000in}{1.815000in}%
\pgfsys@useobject{currentmarker}{}%
\end{pgfscope}%
\end{pgfscope}%
\begin{pgfscope}%
\definecolor{textcolor}{rgb}{0.000000,0.000000,0.000000}%
\pgfsetstrokecolor{textcolor}%
\pgfsetfillcolor{textcolor}%
\pgftext[x=0.554634in,y=1.745556in,left,base]{\color{textcolor}\rmfamily\fontsize{14.000000}{16.800000}\selectfont \(\displaystyle 85.0\)}%
\end{pgfscope}%
\begin{pgfscope}%
\pgfpathrectangle{\pgfqpoint{1.000000in}{0.660000in}}{\pgfqpoint{6.200000in}{4.620000in}}%
\pgfusepath{clip}%
\pgfsetrectcap%
\pgfsetroundjoin%
\pgfsetlinewidth{0.803000pt}%
\definecolor{currentstroke}{rgb}{0.690196,0.690196,0.690196}%
\pgfsetstrokecolor{currentstroke}%
\pgfsetdash{}{0pt}%
\pgfpathmoveto{\pgfqpoint{1.000000in}{2.392500in}}%
\pgfpathlineto{\pgfqpoint{7.200000in}{2.392500in}}%
\pgfusepath{stroke}%
\end{pgfscope}%
\begin{pgfscope}%
\pgfsetbuttcap%
\pgfsetroundjoin%
\definecolor{currentfill}{rgb}{0.000000,0.000000,0.000000}%
\pgfsetfillcolor{currentfill}%
\pgfsetlinewidth{0.803000pt}%
\definecolor{currentstroke}{rgb}{0.000000,0.000000,0.000000}%
\pgfsetstrokecolor{currentstroke}%
\pgfsetdash{}{0pt}%
\pgfsys@defobject{currentmarker}{\pgfqpoint{-0.048611in}{0.000000in}}{\pgfqpoint{0.000000in}{0.000000in}}{%
\pgfpathmoveto{\pgfqpoint{0.000000in}{0.000000in}}%
\pgfpathlineto{\pgfqpoint{-0.048611in}{0.000000in}}%
\pgfusepath{stroke,fill}%
}%
\begin{pgfscope}%
\pgfsys@transformshift{1.000000in}{2.392500in}%
\pgfsys@useobject{currentmarker}{}%
\end{pgfscope}%
\end{pgfscope}%
\begin{pgfscope}%
\definecolor{textcolor}{rgb}{0.000000,0.000000,0.000000}%
\pgfsetstrokecolor{textcolor}%
\pgfsetfillcolor{textcolor}%
\pgftext[x=0.554634in,y=2.323056in,left,base]{\color{textcolor}\rmfamily\fontsize{14.000000}{16.800000}\selectfont \(\displaystyle 87.5\)}%
\end{pgfscope}%
\begin{pgfscope}%
\pgfpathrectangle{\pgfqpoint{1.000000in}{0.660000in}}{\pgfqpoint{6.200000in}{4.620000in}}%
\pgfusepath{clip}%
\pgfsetrectcap%
\pgfsetroundjoin%
\pgfsetlinewidth{0.803000pt}%
\definecolor{currentstroke}{rgb}{0.690196,0.690196,0.690196}%
\pgfsetstrokecolor{currentstroke}%
\pgfsetdash{}{0pt}%
\pgfpathmoveto{\pgfqpoint{1.000000in}{2.970000in}}%
\pgfpathlineto{\pgfqpoint{7.200000in}{2.970000in}}%
\pgfusepath{stroke}%
\end{pgfscope}%
\begin{pgfscope}%
\pgfsetbuttcap%
\pgfsetroundjoin%
\definecolor{currentfill}{rgb}{0.000000,0.000000,0.000000}%
\pgfsetfillcolor{currentfill}%
\pgfsetlinewidth{0.803000pt}%
\definecolor{currentstroke}{rgb}{0.000000,0.000000,0.000000}%
\pgfsetstrokecolor{currentstroke}%
\pgfsetdash{}{0pt}%
\pgfsys@defobject{currentmarker}{\pgfqpoint{-0.048611in}{0.000000in}}{\pgfqpoint{0.000000in}{0.000000in}}{%
\pgfpathmoveto{\pgfqpoint{0.000000in}{0.000000in}}%
\pgfpathlineto{\pgfqpoint{-0.048611in}{0.000000in}}%
\pgfusepath{stroke,fill}%
}%
\begin{pgfscope}%
\pgfsys@transformshift{1.000000in}{2.970000in}%
\pgfsys@useobject{currentmarker}{}%
\end{pgfscope}%
\end{pgfscope}%
\begin{pgfscope}%
\definecolor{textcolor}{rgb}{0.000000,0.000000,0.000000}%
\pgfsetstrokecolor{textcolor}%
\pgfsetfillcolor{textcolor}%
\pgftext[x=0.554634in,y=2.900556in,left,base]{\color{textcolor}\rmfamily\fontsize{14.000000}{16.800000}\selectfont \(\displaystyle 90.0\)}%
\end{pgfscope}%
\begin{pgfscope}%
\pgfpathrectangle{\pgfqpoint{1.000000in}{0.660000in}}{\pgfqpoint{6.200000in}{4.620000in}}%
\pgfusepath{clip}%
\pgfsetrectcap%
\pgfsetroundjoin%
\pgfsetlinewidth{0.803000pt}%
\definecolor{currentstroke}{rgb}{0.690196,0.690196,0.690196}%
\pgfsetstrokecolor{currentstroke}%
\pgfsetdash{}{0pt}%
\pgfpathmoveto{\pgfqpoint{1.000000in}{3.547500in}}%
\pgfpathlineto{\pgfqpoint{7.200000in}{3.547500in}}%
\pgfusepath{stroke}%
\end{pgfscope}%
\begin{pgfscope}%
\pgfsetbuttcap%
\pgfsetroundjoin%
\definecolor{currentfill}{rgb}{0.000000,0.000000,0.000000}%
\pgfsetfillcolor{currentfill}%
\pgfsetlinewidth{0.803000pt}%
\definecolor{currentstroke}{rgb}{0.000000,0.000000,0.000000}%
\pgfsetstrokecolor{currentstroke}%
\pgfsetdash{}{0pt}%
\pgfsys@defobject{currentmarker}{\pgfqpoint{-0.048611in}{0.000000in}}{\pgfqpoint{0.000000in}{0.000000in}}{%
\pgfpathmoveto{\pgfqpoint{0.000000in}{0.000000in}}%
\pgfpathlineto{\pgfqpoint{-0.048611in}{0.000000in}}%
\pgfusepath{stroke,fill}%
}%
\begin{pgfscope}%
\pgfsys@transformshift{1.000000in}{3.547500in}%
\pgfsys@useobject{currentmarker}{}%
\end{pgfscope}%
\end{pgfscope}%
\begin{pgfscope}%
\definecolor{textcolor}{rgb}{0.000000,0.000000,0.000000}%
\pgfsetstrokecolor{textcolor}%
\pgfsetfillcolor{textcolor}%
\pgftext[x=0.554634in,y=3.478056in,left,base]{\color{textcolor}\rmfamily\fontsize{14.000000}{16.800000}\selectfont \(\displaystyle 92.5\)}%
\end{pgfscope}%
\begin{pgfscope}%
\pgfpathrectangle{\pgfqpoint{1.000000in}{0.660000in}}{\pgfqpoint{6.200000in}{4.620000in}}%
\pgfusepath{clip}%
\pgfsetrectcap%
\pgfsetroundjoin%
\pgfsetlinewidth{0.803000pt}%
\definecolor{currentstroke}{rgb}{0.690196,0.690196,0.690196}%
\pgfsetstrokecolor{currentstroke}%
\pgfsetdash{}{0pt}%
\pgfpathmoveto{\pgfqpoint{1.000000in}{4.125000in}}%
\pgfpathlineto{\pgfqpoint{7.200000in}{4.125000in}}%
\pgfusepath{stroke}%
\end{pgfscope}%
\begin{pgfscope}%
\pgfsetbuttcap%
\pgfsetroundjoin%
\definecolor{currentfill}{rgb}{0.000000,0.000000,0.000000}%
\pgfsetfillcolor{currentfill}%
\pgfsetlinewidth{0.803000pt}%
\definecolor{currentstroke}{rgb}{0.000000,0.000000,0.000000}%
\pgfsetstrokecolor{currentstroke}%
\pgfsetdash{}{0pt}%
\pgfsys@defobject{currentmarker}{\pgfqpoint{-0.048611in}{0.000000in}}{\pgfqpoint{0.000000in}{0.000000in}}{%
\pgfpathmoveto{\pgfqpoint{0.000000in}{0.000000in}}%
\pgfpathlineto{\pgfqpoint{-0.048611in}{0.000000in}}%
\pgfusepath{stroke,fill}%
}%
\begin{pgfscope}%
\pgfsys@transformshift{1.000000in}{4.125000in}%
\pgfsys@useobject{currentmarker}{}%
\end{pgfscope}%
\end{pgfscope}%
\begin{pgfscope}%
\definecolor{textcolor}{rgb}{0.000000,0.000000,0.000000}%
\pgfsetstrokecolor{textcolor}%
\pgfsetfillcolor{textcolor}%
\pgftext[x=0.554634in,y=4.055556in,left,base]{\color{textcolor}\rmfamily\fontsize{14.000000}{16.800000}\selectfont \(\displaystyle 95.0\)}%
\end{pgfscope}%
\begin{pgfscope}%
\pgfpathrectangle{\pgfqpoint{1.000000in}{0.660000in}}{\pgfqpoint{6.200000in}{4.620000in}}%
\pgfusepath{clip}%
\pgfsetrectcap%
\pgfsetroundjoin%
\pgfsetlinewidth{0.803000pt}%
\definecolor{currentstroke}{rgb}{0.690196,0.690196,0.690196}%
\pgfsetstrokecolor{currentstroke}%
\pgfsetdash{}{0pt}%
\pgfpathmoveto{\pgfqpoint{1.000000in}{4.702500in}}%
\pgfpathlineto{\pgfqpoint{7.200000in}{4.702500in}}%
\pgfusepath{stroke}%
\end{pgfscope}%
\begin{pgfscope}%
\pgfsetbuttcap%
\pgfsetroundjoin%
\definecolor{currentfill}{rgb}{0.000000,0.000000,0.000000}%
\pgfsetfillcolor{currentfill}%
\pgfsetlinewidth{0.803000pt}%
\definecolor{currentstroke}{rgb}{0.000000,0.000000,0.000000}%
\pgfsetstrokecolor{currentstroke}%
\pgfsetdash{}{0pt}%
\pgfsys@defobject{currentmarker}{\pgfqpoint{-0.048611in}{0.000000in}}{\pgfqpoint{0.000000in}{0.000000in}}{%
\pgfpathmoveto{\pgfqpoint{0.000000in}{0.000000in}}%
\pgfpathlineto{\pgfqpoint{-0.048611in}{0.000000in}}%
\pgfusepath{stroke,fill}%
}%
\begin{pgfscope}%
\pgfsys@transformshift{1.000000in}{4.702500in}%
\pgfsys@useobject{currentmarker}{}%
\end{pgfscope}%
\end{pgfscope}%
\begin{pgfscope}%
\definecolor{textcolor}{rgb}{0.000000,0.000000,0.000000}%
\pgfsetstrokecolor{textcolor}%
\pgfsetfillcolor{textcolor}%
\pgftext[x=0.554634in,y=4.633056in,left,base]{\color{textcolor}\rmfamily\fontsize{14.000000}{16.800000}\selectfont \(\displaystyle 97.5\)}%
\end{pgfscope}%
\begin{pgfscope}%
\pgfpathrectangle{\pgfqpoint{1.000000in}{0.660000in}}{\pgfqpoint{6.200000in}{4.620000in}}%
\pgfusepath{clip}%
\pgfsetrectcap%
\pgfsetroundjoin%
\pgfsetlinewidth{0.803000pt}%
\definecolor{currentstroke}{rgb}{0.690196,0.690196,0.690196}%
\pgfsetstrokecolor{currentstroke}%
\pgfsetdash{}{0pt}%
\pgfpathmoveto{\pgfqpoint{1.000000in}{5.280000in}}%
\pgfpathlineto{\pgfqpoint{7.200000in}{5.280000in}}%
\pgfusepath{stroke}%
\end{pgfscope}%
\begin{pgfscope}%
\pgfsetbuttcap%
\pgfsetroundjoin%
\definecolor{currentfill}{rgb}{0.000000,0.000000,0.000000}%
\pgfsetfillcolor{currentfill}%
\pgfsetlinewidth{0.803000pt}%
\definecolor{currentstroke}{rgb}{0.000000,0.000000,0.000000}%
\pgfsetstrokecolor{currentstroke}%
\pgfsetdash{}{0pt}%
\pgfsys@defobject{currentmarker}{\pgfqpoint{-0.048611in}{0.000000in}}{\pgfqpoint{0.000000in}{0.000000in}}{%
\pgfpathmoveto{\pgfqpoint{0.000000in}{0.000000in}}%
\pgfpathlineto{\pgfqpoint{-0.048611in}{0.000000in}}%
\pgfusepath{stroke,fill}%
}%
\begin{pgfscope}%
\pgfsys@transformshift{1.000000in}{5.280000in}%
\pgfsys@useobject{currentmarker}{}%
\end{pgfscope}%
\end{pgfscope}%
\begin{pgfscope}%
\definecolor{textcolor}{rgb}{0.000000,0.000000,0.000000}%
\pgfsetstrokecolor{textcolor}%
\pgfsetfillcolor{textcolor}%
\pgftext[x=0.456719in,y=5.210556in,left,base]{\color{textcolor}\rmfamily\fontsize{14.000000}{16.800000}\selectfont \(\displaystyle 100.0\)}%
\end{pgfscope}%
\begin{pgfscope}%
\definecolor{textcolor}{rgb}{0.000000,0.000000,0.000000}%
\pgfsetstrokecolor{textcolor}%
\pgfsetfillcolor{textcolor}%
\pgftext[x=0.401163in,y=2.970000in,,bottom,rotate=90.000000]{\color{textcolor}\rmfamily\fontsize{24.000000}{28.800000}\selectfont Accuracy}%
\end{pgfscope}%
\begin{pgfscope}%
\pgfpathrectangle{\pgfqpoint{1.000000in}{0.660000in}}{\pgfqpoint{6.200000in}{4.620000in}}%
\pgfusepath{clip}%
\pgfsetbuttcap%
\pgfsetroundjoin%
\pgfsetlinewidth{1.505625pt}%
\definecolor{currentstroke}{rgb}{0.000000,0.000000,0.000000}%
\pgfsetstrokecolor{currentstroke}%
\pgfsetdash{}{0pt}%
\pgfpathmoveto{\pgfqpoint{1.675758in}{1.493910in}}%
\pgfpathlineto{\pgfqpoint{1.675758in}{1.845030in}}%
\pgfusepath{stroke}%
\end{pgfscope}%
\begin{pgfscope}%
\pgfpathrectangle{\pgfqpoint{1.000000in}{0.660000in}}{\pgfqpoint{6.200000in}{4.620000in}}%
\pgfusepath{clip}%
\pgfsetbuttcap%
\pgfsetroundjoin%
\pgfsetlinewidth{1.505625pt}%
\definecolor{currentstroke}{rgb}{0.000000,0.000000,0.000000}%
\pgfsetstrokecolor{currentstroke}%
\pgfsetdash{}{0pt}%
\pgfpathmoveto{\pgfqpoint{2.887879in}{3.177900in}}%
\pgfpathlineto{\pgfqpoint{2.887879in}{3.676860in}}%
\pgfusepath{stroke}%
\end{pgfscope}%
\begin{pgfscope}%
\pgfpathrectangle{\pgfqpoint{1.000000in}{0.660000in}}{\pgfqpoint{6.200000in}{4.620000in}}%
\pgfusepath{clip}%
\pgfsetbuttcap%
\pgfsetroundjoin%
\pgfsetlinewidth{1.505625pt}%
\definecolor{currentstroke}{rgb}{0.000000,0.000000,0.000000}%
\pgfsetstrokecolor{currentstroke}%
\pgfsetdash{}{0pt}%
\pgfpathmoveto{\pgfqpoint{4.100000in}{3.932115in}}%
\pgfpathlineto{\pgfqpoint{4.100000in}{4.396425in}}%
\pgfusepath{stroke}%
\end{pgfscope}%
\begin{pgfscope}%
\pgfpathrectangle{\pgfqpoint{1.000000in}{0.660000in}}{\pgfqpoint{6.200000in}{4.620000in}}%
\pgfusepath{clip}%
\pgfsetbuttcap%
\pgfsetroundjoin%
\pgfsetlinewidth{1.505625pt}%
\definecolor{currentstroke}{rgb}{0.000000,0.000000,0.000000}%
\pgfsetstrokecolor{currentstroke}%
\pgfsetdash{}{0pt}%
\pgfpathmoveto{\pgfqpoint{5.312121in}{4.143480in}}%
\pgfpathlineto{\pgfqpoint{5.312121in}{4.545420in}}%
\pgfusepath{stroke}%
\end{pgfscope}%
\begin{pgfscope}%
\pgfpathrectangle{\pgfqpoint{1.000000in}{0.660000in}}{\pgfqpoint{6.200000in}{4.620000in}}%
\pgfusepath{clip}%
\pgfsetbuttcap%
\pgfsetroundjoin%
\pgfsetlinewidth{1.505625pt}%
\definecolor{currentstroke}{rgb}{0.000000,0.000000,0.000000}%
\pgfsetstrokecolor{currentstroke}%
\pgfsetdash{}{0pt}%
\pgfpathmoveto{\pgfqpoint{6.524242in}{4.081110in}}%
\pgfpathlineto{\pgfqpoint{6.524242in}{4.533870in}}%
\pgfusepath{stroke}%
\end{pgfscope}%
\begin{pgfscope}%
\pgfsetrectcap%
\pgfsetmiterjoin%
\pgfsetlinewidth{0.803000pt}%
\definecolor{currentstroke}{rgb}{0.000000,0.000000,0.000000}%
\pgfsetstrokecolor{currentstroke}%
\pgfsetdash{}{0pt}%
\pgfpathmoveto{\pgfqpoint{1.000000in}{0.660000in}}%
\pgfpathlineto{\pgfqpoint{1.000000in}{5.280000in}}%
\pgfusepath{stroke}%
\end{pgfscope}%
\begin{pgfscope}%
\pgfsetrectcap%
\pgfsetmiterjoin%
\pgfsetlinewidth{0.803000pt}%
\definecolor{currentstroke}{rgb}{0.000000,0.000000,0.000000}%
\pgfsetstrokecolor{currentstroke}%
\pgfsetdash{}{0pt}%
\pgfpathmoveto{\pgfqpoint{7.200000in}{0.660000in}}%
\pgfpathlineto{\pgfqpoint{7.200000in}{5.280000in}}%
\pgfusepath{stroke}%
\end{pgfscope}%
\begin{pgfscope}%
\pgfsetrectcap%
\pgfsetmiterjoin%
\pgfsetlinewidth{0.803000pt}%
\definecolor{currentstroke}{rgb}{0.000000,0.000000,0.000000}%
\pgfsetstrokecolor{currentstroke}%
\pgfsetdash{}{0pt}%
\pgfpathmoveto{\pgfqpoint{1.000000in}{0.660000in}}%
\pgfpathlineto{\pgfqpoint{7.200000in}{0.660000in}}%
\pgfusepath{stroke}%
\end{pgfscope}%
\begin{pgfscope}%
\pgfsetrectcap%
\pgfsetmiterjoin%
\pgfsetlinewidth{0.803000pt}%
\definecolor{currentstroke}{rgb}{0.000000,0.000000,0.000000}%
\pgfsetstrokecolor{currentstroke}%
\pgfsetdash{}{0pt}%
\pgfpathmoveto{\pgfqpoint{1.000000in}{5.280000in}}%
\pgfpathlineto{\pgfqpoint{7.200000in}{5.280000in}}%
\pgfusepath{stroke}%
\end{pgfscope}%
\end{pgfpicture}%
\makeatother%
\endgroup%

%% file: figure/ocr_b.pgf
%% Creator: Matplotlib, PGF backend
%%
%% To include the figure in your LaTeX document, write
%%   \input{<filename>.pgf}
%%
%% Make sure the required packages are loaded in your preamble
%%   \usepackage{pgf}
%%
%% Figures using additional raster images can only be included by \input if
%% they are in the same directory as the main LaTeX file. For loading figures
%% from other directories you can use the `import` package
%%   \usepackage{import}
%% and then include the figures with
%%   \import{<path to file>}{<filename>.pgf}
%%
%% Matplotlib used the following preamble
%%
\begingroup%
\makeatletter%
\begin{pgfpicture}%
\pgfpathrectangle{\pgfpointorigin}{\pgfqpoint{8.000000in}{6.000000in}}%
\pgfusepath{use as bounding box, clip}%
\begin{pgfscope}%
\pgfsetbuttcap%
\pgfsetmiterjoin%
\definecolor{currentfill}{rgb}{1.000000,1.000000,1.000000}%
\pgfsetfillcolor{currentfill}%
\pgfsetlinewidth{0.000000pt}%
\definecolor{currentstroke}{rgb}{1.000000,1.000000,1.000000}%
\pgfsetstrokecolor{currentstroke}%
\pgfsetdash{}{0pt}%
\pgfpathmoveto{\pgfqpoint{0.000000in}{0.000000in}}%
\pgfpathlineto{\pgfqpoint{8.000000in}{0.000000in}}%
\pgfpathlineto{\pgfqpoint{8.000000in}{6.000000in}}%
\pgfpathlineto{\pgfqpoint{0.000000in}{6.000000in}}%
\pgfpathclose%
\pgfusepath{fill}%
\end{pgfscope}%
\begin{pgfscope}%
\pgfsetbuttcap%
\pgfsetmiterjoin%
\definecolor{currentfill}{rgb}{1.000000,1.000000,1.000000}%
\pgfsetfillcolor{currentfill}%
\pgfsetlinewidth{0.000000pt}%
\definecolor{currentstroke}{rgb}{0.000000,0.000000,0.000000}%
\pgfsetstrokecolor{currentstroke}%
\pgfsetstrokeopacity{0.000000}%
\pgfsetdash{}{0pt}%
\pgfpathmoveto{\pgfqpoint{1.000000in}{0.660000in}}%
\pgfpathlineto{\pgfqpoint{7.200000in}{0.660000in}}%
\pgfpathlineto{\pgfqpoint{7.200000in}{5.280000in}}%
\pgfpathlineto{\pgfqpoint{1.000000in}{5.280000in}}%
\pgfpathclose%
\pgfusepath{fill}%
\end{pgfscope}%
\begin{pgfscope}%
\pgfsetbuttcap%
\pgfsetroundjoin%
\definecolor{currentfill}{rgb}{0.000000,0.000000,0.000000}%
\pgfsetfillcolor{currentfill}%
\pgfsetlinewidth{0.803000pt}%
\definecolor{currentstroke}{rgb}{0.000000,0.000000,0.000000}%
\pgfsetstrokecolor{currentstroke}%
\pgfsetdash{}{0pt}%
\pgfsys@defobject{currentmarker}{\pgfqpoint{0.000000in}{-0.048611in}}{\pgfqpoint{0.000000in}{0.000000in}}{%
\pgfpathmoveto{\pgfqpoint{0.000000in}{0.000000in}}%
\pgfpathlineto{\pgfqpoint{0.000000in}{-0.048611in}}%
\pgfusepath{stroke,fill}%
}%
\begin{pgfscope}%
\pgfsys@transformshift{1.281818in}{0.660000in}%
\pgfsys@useobject{currentmarker}{}%
\end{pgfscope}%
\end{pgfscope}%
\begin{pgfscope}%
\definecolor{textcolor}{rgb}{0.000000,0.000000,0.000000}%
\pgfsetstrokecolor{textcolor}%
\pgfsetfillcolor{textcolor}%
\pgftext[x=1.281818in,y=0.562778in,,top]{\color{textcolor}\rmfamily\fontsize{18.000000}{21.600000}\selectfont 0}%
\end{pgfscope}%
\begin{pgfscope}%
\pgfsetbuttcap%
\pgfsetroundjoin%
\definecolor{currentfill}{rgb}{0.000000,0.000000,0.000000}%
\pgfsetfillcolor{currentfill}%
\pgfsetlinewidth{0.803000pt}%
\definecolor{currentstroke}{rgb}{0.000000,0.000000,0.000000}%
\pgfsetstrokecolor{currentstroke}%
\pgfsetdash{}{0pt}%
\pgfsys@defobject{currentmarker}{\pgfqpoint{0.000000in}{-0.048611in}}{\pgfqpoint{0.000000in}{0.000000in}}{%
\pgfpathmoveto{\pgfqpoint{0.000000in}{0.000000in}}%
\pgfpathlineto{\pgfqpoint{0.000000in}{-0.048611in}}%
\pgfusepath{stroke,fill}%
}%
\begin{pgfscope}%
\pgfsys@transformshift{2.221212in}{0.660000in}%
\pgfsys@useobject{currentmarker}{}%
\end{pgfscope}%
\end{pgfscope}%
\begin{pgfscope}%
\definecolor{textcolor}{rgb}{0.000000,0.000000,0.000000}%
\pgfsetstrokecolor{textcolor}%
\pgfsetfillcolor{textcolor}%
\pgftext[x=2.221212in,y=0.562778in,,top]{\color{textcolor}\rmfamily\fontsize{18.000000}{21.600000}\selectfont 1}%
\end{pgfscope}%
\begin{pgfscope}%
\pgfsetbuttcap%
\pgfsetroundjoin%
\definecolor{currentfill}{rgb}{0.000000,0.000000,0.000000}%
\pgfsetfillcolor{currentfill}%
\pgfsetlinewidth{0.803000pt}%
\definecolor{currentstroke}{rgb}{0.000000,0.000000,0.000000}%
\pgfsetstrokecolor{currentstroke}%
\pgfsetdash{}{0pt}%
\pgfsys@defobject{currentmarker}{\pgfqpoint{0.000000in}{-0.048611in}}{\pgfqpoint{0.000000in}{0.000000in}}{%
\pgfpathmoveto{\pgfqpoint{0.000000in}{0.000000in}}%
\pgfpathlineto{\pgfqpoint{0.000000in}{-0.048611in}}%
\pgfusepath{stroke,fill}%
}%
\begin{pgfscope}%
\pgfsys@transformshift{3.160606in}{0.660000in}%
\pgfsys@useobject{currentmarker}{}%
\end{pgfscope}%
\end{pgfscope}%
\begin{pgfscope}%
\definecolor{textcolor}{rgb}{0.000000,0.000000,0.000000}%
\pgfsetstrokecolor{textcolor}%
\pgfsetfillcolor{textcolor}%
\pgftext[x=3.160606in,y=0.562778in,,top]{\color{textcolor}\rmfamily\fontsize{18.000000}{21.600000}\selectfont 2}%
\end{pgfscope}%
\begin{pgfscope}%
\pgfsetbuttcap%
\pgfsetroundjoin%
\definecolor{currentfill}{rgb}{0.000000,0.000000,0.000000}%
\pgfsetfillcolor{currentfill}%
\pgfsetlinewidth{0.803000pt}%
\definecolor{currentstroke}{rgb}{0.000000,0.000000,0.000000}%
\pgfsetstrokecolor{currentstroke}%
\pgfsetdash{}{0pt}%
\pgfsys@defobject{currentmarker}{\pgfqpoint{0.000000in}{-0.048611in}}{\pgfqpoint{0.000000in}{0.000000in}}{%
\pgfpathmoveto{\pgfqpoint{0.000000in}{0.000000in}}%
\pgfpathlineto{\pgfqpoint{0.000000in}{-0.048611in}}%
\pgfusepath{stroke,fill}%
}%
\begin{pgfscope}%
\pgfsys@transformshift{4.100000in}{0.660000in}%
\pgfsys@useobject{currentmarker}{}%
\end{pgfscope}%
\end{pgfscope}%
\begin{pgfscope}%
\definecolor{textcolor}{rgb}{0.000000,0.000000,0.000000}%
\pgfsetstrokecolor{textcolor}%
\pgfsetfillcolor{textcolor}%
\pgftext[x=4.100000in,y=0.562778in,,top]{\color{textcolor}\rmfamily\fontsize{18.000000}{21.600000}\selectfont 3}%
\end{pgfscope}%
\begin{pgfscope}%
\pgfsetbuttcap%
\pgfsetroundjoin%
\definecolor{currentfill}{rgb}{0.000000,0.000000,0.000000}%
\pgfsetfillcolor{currentfill}%
\pgfsetlinewidth{0.803000pt}%
\definecolor{currentstroke}{rgb}{0.000000,0.000000,0.000000}%
\pgfsetstrokecolor{currentstroke}%
\pgfsetdash{}{0pt}%
\pgfsys@defobject{currentmarker}{\pgfqpoint{0.000000in}{-0.048611in}}{\pgfqpoint{0.000000in}{0.000000in}}{%
\pgfpathmoveto{\pgfqpoint{0.000000in}{0.000000in}}%
\pgfpathlineto{\pgfqpoint{0.000000in}{-0.048611in}}%
\pgfusepath{stroke,fill}%
}%
\begin{pgfscope}%
\pgfsys@transformshift{5.039394in}{0.660000in}%
\pgfsys@useobject{currentmarker}{}%
\end{pgfscope}%
\end{pgfscope}%
\begin{pgfscope}%
\definecolor{textcolor}{rgb}{0.000000,0.000000,0.000000}%
\pgfsetstrokecolor{textcolor}%
\pgfsetfillcolor{textcolor}%
\pgftext[x=5.039394in,y=0.562778in,,top]{\color{textcolor}\rmfamily\fontsize{18.000000}{21.600000}\selectfont 4}%
\end{pgfscope}%
\begin{pgfscope}%
\pgfsetbuttcap%
\pgfsetroundjoin%
\definecolor{currentfill}{rgb}{0.000000,0.000000,0.000000}%
\pgfsetfillcolor{currentfill}%
\pgfsetlinewidth{0.803000pt}%
\definecolor{currentstroke}{rgb}{0.000000,0.000000,0.000000}%
\pgfsetstrokecolor{currentstroke}%
\pgfsetdash{}{0pt}%
\pgfsys@defobject{currentmarker}{\pgfqpoint{0.000000in}{-0.048611in}}{\pgfqpoint{0.000000in}{0.000000in}}{%
\pgfpathmoveto{\pgfqpoint{0.000000in}{0.000000in}}%
\pgfpathlineto{\pgfqpoint{0.000000in}{-0.048611in}}%
\pgfusepath{stroke,fill}%
}%
\begin{pgfscope}%
\pgfsys@transformshift{5.978788in}{0.660000in}%
\pgfsys@useobject{currentmarker}{}%
\end{pgfscope}%
\end{pgfscope}%
\begin{pgfscope}%
\definecolor{textcolor}{rgb}{0.000000,0.000000,0.000000}%
\pgfsetstrokecolor{textcolor}%
\pgfsetfillcolor{textcolor}%
\pgftext[x=5.978788in,y=0.562778in,,top]{\color{textcolor}\rmfamily\fontsize{18.000000}{21.600000}\selectfont 5}%
\end{pgfscope}%
\begin{pgfscope}%
\pgfsetbuttcap%
\pgfsetroundjoin%
\definecolor{currentfill}{rgb}{0.000000,0.000000,0.000000}%
\pgfsetfillcolor{currentfill}%
\pgfsetlinewidth{0.803000pt}%
\definecolor{currentstroke}{rgb}{0.000000,0.000000,0.000000}%
\pgfsetstrokecolor{currentstroke}%
\pgfsetdash{}{0pt}%
\pgfsys@defobject{currentmarker}{\pgfqpoint{0.000000in}{-0.048611in}}{\pgfqpoint{0.000000in}{0.000000in}}{%
\pgfpathmoveto{\pgfqpoint{0.000000in}{0.000000in}}%
\pgfpathlineto{\pgfqpoint{0.000000in}{-0.048611in}}%
\pgfusepath{stroke,fill}%
}%
\begin{pgfscope}%
\pgfsys@transformshift{6.918182in}{0.660000in}%
\pgfsys@useobject{currentmarker}{}%
\end{pgfscope}%
\end{pgfscope}%
\begin{pgfscope}%
\definecolor{textcolor}{rgb}{0.000000,0.000000,0.000000}%
\pgfsetstrokecolor{textcolor}%
\pgfsetfillcolor{textcolor}%
\pgftext[x=6.918182in,y=0.562778in,,top]{\color{textcolor}\rmfamily\fontsize{18.000000}{21.600000}\selectfont 6}%
\end{pgfscope}%
\begin{pgfscope}%
\definecolor{textcolor}{rgb}{0.000000,0.000000,0.000000}%
\pgfsetstrokecolor{textcolor}%
\pgfsetfillcolor{textcolor}%
\pgftext[x=4.100000in,y=0.293874in,,top]{\color{textcolor}\rmfamily\fontsize{24.000000}{28.800000}\selectfont Order of the factors}%
\end{pgfscope}%
\begin{pgfscope}%
\pgfpathrectangle{\pgfqpoint{1.000000in}{0.660000in}}{\pgfqpoint{6.200000in}{4.620000in}}%
\pgfusepath{clip}%
\pgfsetrectcap%
\pgfsetroundjoin%
\pgfsetlinewidth{0.803000pt}%
\definecolor{currentstroke}{rgb}{0.690196,0.690196,0.690196}%
\pgfsetstrokecolor{currentstroke}%
\pgfsetdash{}{0pt}%
\pgfpathmoveto{\pgfqpoint{1.000000in}{0.660000in}}%
\pgfpathlineto{\pgfqpoint{7.200000in}{0.660000in}}%
\pgfusepath{stroke}%
\end{pgfscope}%
\begin{pgfscope}%
\pgfsetbuttcap%
\pgfsetroundjoin%
\definecolor{currentfill}{rgb}{0.000000,0.000000,0.000000}%
\pgfsetfillcolor{currentfill}%
\pgfsetlinewidth{0.803000pt}%
\definecolor{currentstroke}{rgb}{0.000000,0.000000,0.000000}%
\pgfsetstrokecolor{currentstroke}%
\pgfsetdash{}{0pt}%
\pgfsys@defobject{currentmarker}{\pgfqpoint{-0.048611in}{0.000000in}}{\pgfqpoint{0.000000in}{0.000000in}}{%
\pgfpathmoveto{\pgfqpoint{0.000000in}{0.000000in}}%
\pgfpathlineto{\pgfqpoint{-0.048611in}{0.000000in}}%
\pgfusepath{stroke,fill}%
}%
\begin{pgfscope}%
\pgfsys@transformshift{1.000000in}{0.660000in}%
\pgfsys@useobject{currentmarker}{}%
\end{pgfscope}%
\end{pgfscope}%
\begin{pgfscope}%
\definecolor{textcolor}{rgb}{0.000000,0.000000,0.000000}%
\pgfsetstrokecolor{textcolor}%
\pgfsetfillcolor{textcolor}%
\pgftext[x=0.652550in,y=0.590556in,left,base]{\color{textcolor}\rmfamily\fontsize{14.000000}{16.800000}\selectfont \(\displaystyle 0.0\)}%
\end{pgfscope}%
\begin{pgfscope}%
\pgfpathrectangle{\pgfqpoint{1.000000in}{0.660000in}}{\pgfqpoint{6.200000in}{4.620000in}}%
\pgfusepath{clip}%
\pgfsetrectcap%
\pgfsetroundjoin%
\pgfsetlinewidth{0.803000pt}%
\definecolor{currentstroke}{rgb}{0.690196,0.690196,0.690196}%
\pgfsetstrokecolor{currentstroke}%
\pgfsetdash{}{0pt}%
\pgfpathmoveto{\pgfqpoint{1.000000in}{1.430000in}}%
\pgfpathlineto{\pgfqpoint{7.200000in}{1.430000in}}%
\pgfusepath{stroke}%
\end{pgfscope}%
\begin{pgfscope}%
\pgfsetbuttcap%
\pgfsetroundjoin%
\definecolor{currentfill}{rgb}{0.000000,0.000000,0.000000}%
\pgfsetfillcolor{currentfill}%
\pgfsetlinewidth{0.803000pt}%
\definecolor{currentstroke}{rgb}{0.000000,0.000000,0.000000}%
\pgfsetstrokecolor{currentstroke}%
\pgfsetdash{}{0pt}%
\pgfsys@defobject{currentmarker}{\pgfqpoint{-0.048611in}{0.000000in}}{\pgfqpoint{0.000000in}{0.000000in}}{%
\pgfpathmoveto{\pgfqpoint{0.000000in}{0.000000in}}%
\pgfpathlineto{\pgfqpoint{-0.048611in}{0.000000in}}%
\pgfusepath{stroke,fill}%
}%
\begin{pgfscope}%
\pgfsys@transformshift{1.000000in}{1.430000in}%
\pgfsys@useobject{currentmarker}{}%
\end{pgfscope}%
\end{pgfscope}%
\begin{pgfscope}%
\definecolor{textcolor}{rgb}{0.000000,0.000000,0.000000}%
\pgfsetstrokecolor{textcolor}%
\pgfsetfillcolor{textcolor}%
\pgftext[x=0.652550in,y=1.360556in,left,base]{\color{textcolor}\rmfamily\fontsize{14.000000}{16.800000}\selectfont \(\displaystyle 0.5\)}%
\end{pgfscope}%
\begin{pgfscope}%
\pgfpathrectangle{\pgfqpoint{1.000000in}{0.660000in}}{\pgfqpoint{6.200000in}{4.620000in}}%
\pgfusepath{clip}%
\pgfsetrectcap%
\pgfsetroundjoin%
\pgfsetlinewidth{0.803000pt}%
\definecolor{currentstroke}{rgb}{0.690196,0.690196,0.690196}%
\pgfsetstrokecolor{currentstroke}%
\pgfsetdash{}{0pt}%
\pgfpathmoveto{\pgfqpoint{1.000000in}{2.200000in}}%
\pgfpathlineto{\pgfqpoint{7.200000in}{2.200000in}}%
\pgfusepath{stroke}%
\end{pgfscope}%
\begin{pgfscope}%
\pgfsetbuttcap%
\pgfsetroundjoin%
\definecolor{currentfill}{rgb}{0.000000,0.000000,0.000000}%
\pgfsetfillcolor{currentfill}%
\pgfsetlinewidth{0.803000pt}%
\definecolor{currentstroke}{rgb}{0.000000,0.000000,0.000000}%
\pgfsetstrokecolor{currentstroke}%
\pgfsetdash{}{0pt}%
\pgfsys@defobject{currentmarker}{\pgfqpoint{-0.048611in}{0.000000in}}{\pgfqpoint{0.000000in}{0.000000in}}{%
\pgfpathmoveto{\pgfqpoint{0.000000in}{0.000000in}}%
\pgfpathlineto{\pgfqpoint{-0.048611in}{0.000000in}}%
\pgfusepath{stroke,fill}%
}%
\begin{pgfscope}%
\pgfsys@transformshift{1.000000in}{2.200000in}%
\pgfsys@useobject{currentmarker}{}%
\end{pgfscope}%
\end{pgfscope}%
\begin{pgfscope}%
\definecolor{textcolor}{rgb}{0.000000,0.000000,0.000000}%
\pgfsetstrokecolor{textcolor}%
\pgfsetfillcolor{textcolor}%
\pgftext[x=0.652550in,y=2.130556in,left,base]{\color{textcolor}\rmfamily\fontsize{14.000000}{16.800000}\selectfont \(\displaystyle 1.0\)}%
\end{pgfscope}%
\begin{pgfscope}%
\pgfpathrectangle{\pgfqpoint{1.000000in}{0.660000in}}{\pgfqpoint{6.200000in}{4.620000in}}%
\pgfusepath{clip}%
\pgfsetrectcap%
\pgfsetroundjoin%
\pgfsetlinewidth{0.803000pt}%
\definecolor{currentstroke}{rgb}{0.690196,0.690196,0.690196}%
\pgfsetstrokecolor{currentstroke}%
\pgfsetdash{}{0pt}%
\pgfpathmoveto{\pgfqpoint{1.000000in}{2.970000in}}%
\pgfpathlineto{\pgfqpoint{7.200000in}{2.970000in}}%
\pgfusepath{stroke}%
\end{pgfscope}%
\begin{pgfscope}%
\pgfsetbuttcap%
\pgfsetroundjoin%
\definecolor{currentfill}{rgb}{0.000000,0.000000,0.000000}%
\pgfsetfillcolor{currentfill}%
\pgfsetlinewidth{0.803000pt}%
\definecolor{currentstroke}{rgb}{0.000000,0.000000,0.000000}%
\pgfsetstrokecolor{currentstroke}%
\pgfsetdash{}{0pt}%
\pgfsys@defobject{currentmarker}{\pgfqpoint{-0.048611in}{0.000000in}}{\pgfqpoint{0.000000in}{0.000000in}}{%
\pgfpathmoveto{\pgfqpoint{0.000000in}{0.000000in}}%
\pgfpathlineto{\pgfqpoint{-0.048611in}{0.000000in}}%
\pgfusepath{stroke,fill}%
}%
\begin{pgfscope}%
\pgfsys@transformshift{1.000000in}{2.970000in}%
\pgfsys@useobject{currentmarker}{}%
\end{pgfscope}%
\end{pgfscope}%
\begin{pgfscope}%
\definecolor{textcolor}{rgb}{0.000000,0.000000,0.000000}%
\pgfsetstrokecolor{textcolor}%
\pgfsetfillcolor{textcolor}%
\pgftext[x=0.652550in,y=2.900556in,left,base]{\color{textcolor}\rmfamily\fontsize{14.000000}{16.800000}\selectfont \(\displaystyle 1.5\)}%
\end{pgfscope}%
\begin{pgfscope}%
\pgfpathrectangle{\pgfqpoint{1.000000in}{0.660000in}}{\pgfqpoint{6.200000in}{4.620000in}}%
\pgfusepath{clip}%
\pgfsetrectcap%
\pgfsetroundjoin%
\pgfsetlinewidth{0.803000pt}%
\definecolor{currentstroke}{rgb}{0.690196,0.690196,0.690196}%
\pgfsetstrokecolor{currentstroke}%
\pgfsetdash{}{0pt}%
\pgfpathmoveto{\pgfqpoint{1.000000in}{3.740000in}}%
\pgfpathlineto{\pgfqpoint{7.200000in}{3.740000in}}%
\pgfusepath{stroke}%
\end{pgfscope}%
\begin{pgfscope}%
\pgfsetbuttcap%
\pgfsetroundjoin%
\definecolor{currentfill}{rgb}{0.000000,0.000000,0.000000}%
\pgfsetfillcolor{currentfill}%
\pgfsetlinewidth{0.803000pt}%
\definecolor{currentstroke}{rgb}{0.000000,0.000000,0.000000}%
\pgfsetstrokecolor{currentstroke}%
\pgfsetdash{}{0pt}%
\pgfsys@defobject{currentmarker}{\pgfqpoint{-0.048611in}{0.000000in}}{\pgfqpoint{0.000000in}{0.000000in}}{%
\pgfpathmoveto{\pgfqpoint{0.000000in}{0.000000in}}%
\pgfpathlineto{\pgfqpoint{-0.048611in}{0.000000in}}%
\pgfusepath{stroke,fill}%
}%
\begin{pgfscope}%
\pgfsys@transformshift{1.000000in}{3.740000in}%
\pgfsys@useobject{currentmarker}{}%
\end{pgfscope}%
\end{pgfscope}%
\begin{pgfscope}%
\definecolor{textcolor}{rgb}{0.000000,0.000000,0.000000}%
\pgfsetstrokecolor{textcolor}%
\pgfsetfillcolor{textcolor}%
\pgftext[x=0.652550in,y=3.670556in,left,base]{\color{textcolor}\rmfamily\fontsize{14.000000}{16.800000}\selectfont \(\displaystyle 2.0\)}%
\end{pgfscope}%
\begin{pgfscope}%
\pgfpathrectangle{\pgfqpoint{1.000000in}{0.660000in}}{\pgfqpoint{6.200000in}{4.620000in}}%
\pgfusepath{clip}%
\pgfsetrectcap%
\pgfsetroundjoin%
\pgfsetlinewidth{0.803000pt}%
\definecolor{currentstroke}{rgb}{0.690196,0.690196,0.690196}%
\pgfsetstrokecolor{currentstroke}%
\pgfsetdash{}{0pt}%
\pgfpathmoveto{\pgfqpoint{1.000000in}{4.510000in}}%
\pgfpathlineto{\pgfqpoint{7.200000in}{4.510000in}}%
\pgfusepath{stroke}%
\end{pgfscope}%
\begin{pgfscope}%
\pgfsetbuttcap%
\pgfsetroundjoin%
\definecolor{currentfill}{rgb}{0.000000,0.000000,0.000000}%
\pgfsetfillcolor{currentfill}%
\pgfsetlinewidth{0.803000pt}%
\definecolor{currentstroke}{rgb}{0.000000,0.000000,0.000000}%
\pgfsetstrokecolor{currentstroke}%
\pgfsetdash{}{0pt}%
\pgfsys@defobject{currentmarker}{\pgfqpoint{-0.048611in}{0.000000in}}{\pgfqpoint{0.000000in}{0.000000in}}{%
\pgfpathmoveto{\pgfqpoint{0.000000in}{0.000000in}}%
\pgfpathlineto{\pgfqpoint{-0.048611in}{0.000000in}}%
\pgfusepath{stroke,fill}%
}%
\begin{pgfscope}%
\pgfsys@transformshift{1.000000in}{4.510000in}%
\pgfsys@useobject{currentmarker}{}%
\end{pgfscope}%
\end{pgfscope}%
\begin{pgfscope}%
\definecolor{textcolor}{rgb}{0.000000,0.000000,0.000000}%
\pgfsetstrokecolor{textcolor}%
\pgfsetfillcolor{textcolor}%
\pgftext[x=0.652550in,y=4.440556in,left,base]{\color{textcolor}\rmfamily\fontsize{14.000000}{16.800000}\selectfont \(\displaystyle 2.5\)}%
\end{pgfscope}%
\begin{pgfscope}%
\pgfpathrectangle{\pgfqpoint{1.000000in}{0.660000in}}{\pgfqpoint{6.200000in}{4.620000in}}%
\pgfusepath{clip}%
\pgfsetrectcap%
\pgfsetroundjoin%
\pgfsetlinewidth{0.803000pt}%
\definecolor{currentstroke}{rgb}{0.690196,0.690196,0.690196}%
\pgfsetstrokecolor{currentstroke}%
\pgfsetdash{}{0pt}%
\pgfpathmoveto{\pgfqpoint{1.000000in}{5.280000in}}%
\pgfpathlineto{\pgfqpoint{7.200000in}{5.280000in}}%
\pgfusepath{stroke}%
\end{pgfscope}%
\begin{pgfscope}%
\pgfsetbuttcap%
\pgfsetroundjoin%
\definecolor{currentfill}{rgb}{0.000000,0.000000,0.000000}%
\pgfsetfillcolor{currentfill}%
\pgfsetlinewidth{0.803000pt}%
\definecolor{currentstroke}{rgb}{0.000000,0.000000,0.000000}%
\pgfsetstrokecolor{currentstroke}%
\pgfsetdash{}{0pt}%
\pgfsys@defobject{currentmarker}{\pgfqpoint{-0.048611in}{0.000000in}}{\pgfqpoint{0.000000in}{0.000000in}}{%
\pgfpathmoveto{\pgfqpoint{0.000000in}{0.000000in}}%
\pgfpathlineto{\pgfqpoint{-0.048611in}{0.000000in}}%
\pgfusepath{stroke,fill}%
}%
\begin{pgfscope}%
\pgfsys@transformshift{1.000000in}{5.280000in}%
\pgfsys@useobject{currentmarker}{}%
\end{pgfscope}%
\end{pgfscope}%
\begin{pgfscope}%
\definecolor{textcolor}{rgb}{0.000000,0.000000,0.000000}%
\pgfsetstrokecolor{textcolor}%
\pgfsetfillcolor{textcolor}%
\pgftext[x=0.652550in,y=5.210556in,left,base]{\color{textcolor}\rmfamily\fontsize{14.000000}{16.800000}\selectfont \(\displaystyle 3.0\)}%
\end{pgfscope}%
\begin{pgfscope}%
\definecolor{textcolor}{rgb}{0.000000,0.000000,0.000000}%
\pgfsetstrokecolor{textcolor}%
\pgfsetfillcolor{textcolor}%
\pgftext[x=0.596994in,y=2.970000in,,bottom,rotate=90.000000]{\color{textcolor}\rmfamily\fontsize{24.000000}{28.800000}\selectfont Time (sec)}%
\end{pgfscope}%
\begin{pgfscope}%
\pgfpathrectangle{\pgfqpoint{1.000000in}{0.660000in}}{\pgfqpoint{6.200000in}{4.620000in}}%
\pgfusepath{clip}%
\pgfsetrectcap%
\pgfsetroundjoin%
\pgfsetlinewidth{1.505625pt}%
\definecolor{currentstroke}{rgb}{0.000000,0.501961,0.000000}%
\pgfsetstrokecolor{currentstroke}%
\pgfsetdash{}{0pt}%
\pgfpathmoveto{\pgfqpoint{1.281818in}{1.212552in}}%
\pgfpathlineto{\pgfqpoint{2.221212in}{1.953600in}}%
\pgfpathlineto{\pgfqpoint{3.160606in}{2.397120in}}%
\pgfpathlineto{\pgfqpoint{4.100000in}{2.857580in}}%
\pgfpathlineto{\pgfqpoint{5.039394in}{3.355000in}}%
\pgfpathlineto{\pgfqpoint{5.978788in}{3.847800in}}%
\pgfpathlineto{\pgfqpoint{6.918182in}{4.386800in}}%
\pgfusepath{stroke}%
\end{pgfscope}%
\begin{pgfscope}%
\pgfpathrectangle{\pgfqpoint{1.000000in}{0.660000in}}{\pgfqpoint{6.200000in}{4.620000in}}%
\pgfusepath{clip}%
\pgfsetbuttcap%
\pgfsetroundjoin%
\definecolor{currentfill}{rgb}{0.000000,0.501961,0.000000}%
\pgfsetfillcolor{currentfill}%
\pgfsetlinewidth{1.003750pt}%
\definecolor{currentstroke}{rgb}{0.000000,0.501961,0.000000}%
\pgfsetstrokecolor{currentstroke}%
\pgfsetdash{}{0pt}%
\pgfsys@defobject{currentmarker}{\pgfqpoint{-0.041667in}{-0.041667in}}{\pgfqpoint{0.041667in}{0.041667in}}{%
\pgfpathmoveto{\pgfqpoint{0.000000in}{-0.041667in}}%
\pgfpathcurveto{\pgfqpoint{0.011050in}{-0.041667in}}{\pgfqpoint{0.021649in}{-0.037276in}}{\pgfqpoint{0.029463in}{-0.029463in}}%
\pgfpathcurveto{\pgfqpoint{0.037276in}{-0.021649in}}{\pgfqpoint{0.041667in}{-0.011050in}}{\pgfqpoint{0.041667in}{0.000000in}}%
\pgfpathcurveto{\pgfqpoint{0.041667in}{0.011050in}}{\pgfqpoint{0.037276in}{0.021649in}}{\pgfqpoint{0.029463in}{0.029463in}}%
\pgfpathcurveto{\pgfqpoint{0.021649in}{0.037276in}}{\pgfqpoint{0.011050in}{0.041667in}}{\pgfqpoint{0.000000in}{0.041667in}}%
\pgfpathcurveto{\pgfqpoint{-0.011050in}{0.041667in}}{\pgfqpoint{-0.021649in}{0.037276in}}{\pgfqpoint{-0.029463in}{0.029463in}}%
\pgfpathcurveto{\pgfqpoint{-0.037276in}{0.021649in}}{\pgfqpoint{-0.041667in}{0.011050in}}{\pgfqpoint{-0.041667in}{0.000000in}}%
\pgfpathcurveto{\pgfqpoint{-0.041667in}{-0.011050in}}{\pgfqpoint{-0.037276in}{-0.021649in}}{\pgfqpoint{-0.029463in}{-0.029463in}}%
\pgfpathcurveto{\pgfqpoint{-0.021649in}{-0.037276in}}{\pgfqpoint{-0.011050in}{-0.041667in}}{\pgfqpoint{0.000000in}{-0.041667in}}%
\pgfpathclose%
\pgfusepath{stroke,fill}%
}%
\begin{pgfscope}%
\pgfsys@transformshift{1.281818in}{1.212552in}%
\pgfsys@useobject{currentmarker}{}%
\end{pgfscope}%
\begin{pgfscope}%
\pgfsys@transformshift{2.221212in}{1.953600in}%
\pgfsys@useobject{currentmarker}{}%
\end{pgfscope}%
\begin{pgfscope}%
\pgfsys@transformshift{3.160606in}{2.397120in}%
\pgfsys@useobject{currentmarker}{}%
\end{pgfscope}%
\begin{pgfscope}%
\pgfsys@transformshift{4.100000in}{2.857580in}%
\pgfsys@useobject{currentmarker}{}%
\end{pgfscope}%
\begin{pgfscope}%
\pgfsys@transformshift{5.039394in}{3.355000in}%
\pgfsys@useobject{currentmarker}{}%
\end{pgfscope}%
\begin{pgfscope}%
\pgfsys@transformshift{5.978788in}{3.847800in}%
\pgfsys@useobject{currentmarker}{}%
\end{pgfscope}%
\begin{pgfscope}%
\pgfsys@transformshift{6.918182in}{4.386800in}%
\pgfsys@useobject{currentmarker}{}%
\end{pgfscope}%
\end{pgfscope}%
\begin{pgfscope}%
\pgfsetrectcap%
\pgfsetmiterjoin%
\pgfsetlinewidth{0.803000pt}%
\definecolor{currentstroke}{rgb}{0.000000,0.000000,0.000000}%
\pgfsetstrokecolor{currentstroke}%
\pgfsetdash{}{0pt}%
\pgfpathmoveto{\pgfqpoint{1.000000in}{0.660000in}}%
\pgfpathlineto{\pgfqpoint{1.000000in}{5.280000in}}%
\pgfusepath{stroke}%
\end{pgfscope}%
\begin{pgfscope}%
\pgfsetrectcap%
\pgfsetmiterjoin%
\pgfsetlinewidth{0.803000pt}%
\definecolor{currentstroke}{rgb}{0.000000,0.000000,0.000000}%
\pgfsetstrokecolor{currentstroke}%
\pgfsetdash{}{0pt}%
\pgfpathmoveto{\pgfqpoint{7.200000in}{0.660000in}}%
\pgfpathlineto{\pgfqpoint{7.200000in}{5.280000in}}%
\pgfusepath{stroke}%
\end{pgfscope}%
\begin{pgfscope}%
\pgfsetrectcap%
\pgfsetmiterjoin%
\pgfsetlinewidth{0.803000pt}%
\definecolor{currentstroke}{rgb}{0.000000,0.000000,0.000000}%
\pgfsetstrokecolor{currentstroke}%
\pgfsetdash{}{0pt}%
\pgfpathmoveto{\pgfqpoint{1.000000in}{0.660000in}}%
\pgfpathlineto{\pgfqpoint{7.200000in}{0.660000in}}%
\pgfusepath{stroke}%
\end{pgfscope}%
\begin{pgfscope}%
\pgfsetrectcap%
\pgfsetmiterjoin%
\pgfsetlinewidth{0.803000pt}%
\definecolor{currentstroke}{rgb}{0.000000,0.000000,0.000000}%
\pgfsetstrokecolor{currentstroke}%
\pgfsetdash{}{0pt}%
\pgfpathmoveto{\pgfqpoint{1.000000in}{5.280000in}}%
\pgfpathlineto{\pgfqpoint{7.200000in}{5.280000in}}%
\pgfusepath{stroke}%
\end{pgfscope}%
\end{pgfpicture}%
\makeatother%
\endgroup%

%% file: figure/ocr_c.pgf
%% Creator: Matplotlib, PGF backend
%%
%% To include the figure in your LaTeX document, write
%%   \input{<filename>.pgf}
%%
%% Make sure the required packages are loaded in your preamble
%%   \usepackage{pgf}
%%
%% Figures using additional raster images can only be included by \input if
%% they are in the same directory as the main LaTeX file. For loading figures
%% from other directories you can use the `import` package
%%   \usepackage{import}
%% and then include the figures with
%%   \import{<path to file>}{<filename>.pgf}
%%
%% Matplotlib used the following preamble
%%
\begingroup%
\makeatletter%
\begin{pgfpicture}%
\pgfpathrectangle{\pgfpointorigin}{\pgfqpoint{8.000000in}{6.000000in}}%
\pgfusepath{use as bounding box, clip}%
\begin{pgfscope}%
\pgfsetbuttcap%
\pgfsetmiterjoin%
\definecolor{currentfill}{rgb}{1.000000,1.000000,1.000000}%
\pgfsetfillcolor{currentfill}%
\pgfsetlinewidth{0.000000pt}%
\definecolor{currentstroke}{rgb}{1.000000,1.000000,1.000000}%
\pgfsetstrokecolor{currentstroke}%
\pgfsetdash{}{0pt}%
\pgfpathmoveto{\pgfqpoint{0.000000in}{0.000000in}}%
\pgfpathlineto{\pgfqpoint{8.000000in}{0.000000in}}%
\pgfpathlineto{\pgfqpoint{8.000000in}{6.000000in}}%
\pgfpathlineto{\pgfqpoint{0.000000in}{6.000000in}}%
\pgfpathclose%
\pgfusepath{fill}%
\end{pgfscope}%
\begin{pgfscope}%
\pgfsetbuttcap%
\pgfsetmiterjoin%
\definecolor{currentfill}{rgb}{1.000000,1.000000,1.000000}%
\pgfsetfillcolor{currentfill}%
\pgfsetlinewidth{0.000000pt}%
\definecolor{currentstroke}{rgb}{0.000000,0.000000,0.000000}%
\pgfsetstrokecolor{currentstroke}%
\pgfsetstrokeopacity{0.000000}%
\pgfsetdash{}{0pt}%
\pgfpathmoveto{\pgfqpoint{1.000000in}{0.660000in}}%
\pgfpathlineto{\pgfqpoint{7.200000in}{0.660000in}}%
\pgfpathlineto{\pgfqpoint{7.200000in}{5.280000in}}%
\pgfpathlineto{\pgfqpoint{1.000000in}{5.280000in}}%
\pgfpathclose%
\pgfusepath{fill}%
\end{pgfscope}%
\begin{pgfscope}%
\pgfpathrectangle{\pgfqpoint{1.000000in}{0.660000in}}{\pgfqpoint{6.200000in}{4.620000in}}%
\pgfusepath{clip}%
\pgfsetbuttcap%
\pgfsetmiterjoin%
\definecolor{currentfill}{rgb}{1.000000,0.388235,0.278431}%
\pgfsetfillcolor{currentfill}%
\pgfsetlinewidth{0.000000pt}%
\definecolor{currentstroke}{rgb}{0.000000,0.000000,0.000000}%
\pgfsetstrokecolor{currentstroke}%
\pgfsetstrokeopacity{0.000000}%
\pgfsetdash{}{0pt}%
\pgfpathmoveto{\pgfqpoint{1.281818in}{-30.756000in}}%
\pgfpathlineto{\pgfqpoint{2.059248in}{-30.756000in}}%
\pgfpathlineto{\pgfqpoint{2.059248in}{2.175360in}}%
\pgfpathlineto{\pgfqpoint{1.281818in}{2.175360in}}%
\pgfpathclose%
\pgfusepath{fill}%
\end{pgfscope}%
\begin{pgfscope}%
\pgfpathrectangle{\pgfqpoint{1.000000in}{0.660000in}}{\pgfqpoint{6.200000in}{4.620000in}}%
\pgfusepath{clip}%
\pgfsetbuttcap%
\pgfsetmiterjoin%
\definecolor{currentfill}{rgb}{1.000000,0.388235,0.278431}%
\pgfsetfillcolor{currentfill}%
\pgfsetlinewidth{0.000000pt}%
\definecolor{currentstroke}{rgb}{0.000000,0.000000,0.000000}%
\pgfsetstrokecolor{currentstroke}%
\pgfsetstrokeopacity{0.000000}%
\pgfsetdash{}{0pt}%
\pgfpathmoveto{\pgfqpoint{2.253605in}{-30.756000in}}%
\pgfpathlineto{\pgfqpoint{3.031034in}{-30.756000in}}%
\pgfpathlineto{\pgfqpoint{3.031034in}{2.341680in}}%
\pgfpathlineto{\pgfqpoint{2.253605in}{2.341680in}}%
\pgfpathclose%
\pgfusepath{fill}%
\end{pgfscope}%
\begin{pgfscope}%
\pgfpathrectangle{\pgfqpoint{1.000000in}{0.660000in}}{\pgfqpoint{6.200000in}{4.620000in}}%
\pgfusepath{clip}%
\pgfsetbuttcap%
\pgfsetmiterjoin%
\definecolor{currentfill}{rgb}{1.000000,0.388235,0.278431}%
\pgfsetfillcolor{currentfill}%
\pgfsetlinewidth{0.000000pt}%
\definecolor{currentstroke}{rgb}{0.000000,0.000000,0.000000}%
\pgfsetstrokecolor{currentstroke}%
\pgfsetstrokeopacity{0.000000}%
\pgfsetdash{}{0pt}%
\pgfpathmoveto{\pgfqpoint{3.225392in}{-30.756000in}}%
\pgfpathlineto{\pgfqpoint{4.002821in}{-30.756000in}}%
\pgfpathlineto{\pgfqpoint{4.002821in}{2.766720in}}%
\pgfpathlineto{\pgfqpoint{3.225392in}{2.766720in}}%
\pgfpathclose%
\pgfusepath{fill}%
\end{pgfscope}%
\begin{pgfscope}%
\pgfpathrectangle{\pgfqpoint{1.000000in}{0.660000in}}{\pgfqpoint{6.200000in}{4.620000in}}%
\pgfusepath{clip}%
\pgfsetbuttcap%
\pgfsetmiterjoin%
\definecolor{currentfill}{rgb}{1.000000,0.388235,0.278431}%
\pgfsetfillcolor{currentfill}%
\pgfsetlinewidth{0.000000pt}%
\definecolor{currentstroke}{rgb}{0.000000,0.000000,0.000000}%
\pgfsetstrokecolor{currentstroke}%
\pgfsetstrokeopacity{0.000000}%
\pgfsetdash{}{0pt}%
\pgfpathmoveto{\pgfqpoint{4.197179in}{-30.756000in}}%
\pgfpathlineto{\pgfqpoint{4.974608in}{-30.756000in}}%
\pgfpathlineto{\pgfqpoint{4.974608in}{3.454176in}}%
\pgfpathlineto{\pgfqpoint{4.197179in}{3.454176in}}%
\pgfpathclose%
\pgfusepath{fill}%
\end{pgfscope}%
\begin{pgfscope}%
\pgfpathrectangle{\pgfqpoint{1.000000in}{0.660000in}}{\pgfqpoint{6.200000in}{4.620000in}}%
\pgfusepath{clip}%
\pgfsetbuttcap%
\pgfsetmiterjoin%
\definecolor{currentfill}{rgb}{1.000000,0.388235,0.278431}%
\pgfsetfillcolor{currentfill}%
\pgfsetlinewidth{0.000000pt}%
\definecolor{currentstroke}{rgb}{0.000000,0.000000,0.000000}%
\pgfsetstrokecolor{currentstroke}%
\pgfsetstrokeopacity{0.000000}%
\pgfsetdash{}{0pt}%
\pgfpathmoveto{\pgfqpoint{5.168966in}{-30.756000in}}%
\pgfpathlineto{\pgfqpoint{5.946395in}{-30.756000in}}%
\pgfpathlineto{\pgfqpoint{5.946395in}{4.444704in}}%
\pgfpathlineto{\pgfqpoint{5.168966in}{4.444704in}}%
\pgfpathclose%
\pgfusepath{fill}%
\end{pgfscope}%
\begin{pgfscope}%
\pgfpathrectangle{\pgfqpoint{1.000000in}{0.660000in}}{\pgfqpoint{6.200000in}{4.620000in}}%
\pgfusepath{clip}%
\pgfsetbuttcap%
\pgfsetmiterjoin%
\definecolor{currentfill}{rgb}{1.000000,0.388235,0.278431}%
\pgfsetfillcolor{currentfill}%
\pgfsetlinewidth{0.000000pt}%
\definecolor{currentstroke}{rgb}{0.000000,0.000000,0.000000}%
\pgfsetstrokecolor{currentstroke}%
\pgfsetstrokeopacity{0.000000}%
\pgfsetdash{}{0pt}%
\pgfpathmoveto{\pgfqpoint{6.140752in}{-30.756000in}}%
\pgfpathlineto{\pgfqpoint{6.918182in}{-30.756000in}}%
\pgfpathlineto{\pgfqpoint{6.918182in}{4.588848in}}%
\pgfpathlineto{\pgfqpoint{6.140752in}{4.588848in}}%
\pgfpathclose%
\pgfusepath{fill}%
\end{pgfscope}%
\begin{pgfscope}%
\pgfsetbuttcap%
\pgfsetroundjoin%
\definecolor{currentfill}{rgb}{0.000000,0.000000,0.000000}%
\pgfsetfillcolor{currentfill}%
\pgfsetlinewidth{0.803000pt}%
\definecolor{currentstroke}{rgb}{0.000000,0.000000,0.000000}%
\pgfsetstrokecolor{currentstroke}%
\pgfsetdash{}{0pt}%
\pgfsys@defobject{currentmarker}{\pgfqpoint{0.000000in}{-0.048611in}}{\pgfqpoint{0.000000in}{0.000000in}}{%
\pgfpathmoveto{\pgfqpoint{0.000000in}{0.000000in}}%
\pgfpathlineto{\pgfqpoint{0.000000in}{-0.048611in}}%
\pgfusepath{stroke,fill}%
}%
\begin{pgfscope}%
\pgfsys@transformshift{1.670533in}{0.660000in}%
\pgfsys@useobject{currentmarker}{}%
\end{pgfscope}%
\end{pgfscope}%
\begin{pgfscope}%
\definecolor{textcolor}{rgb}{0.000000,0.000000,0.000000}%
\pgfsetstrokecolor{textcolor}%
\pgfsetfillcolor{textcolor}%
\pgftext[x=1.670533in,y=0.562778in,,top]{\color{textcolor}\rmfamily\fontsize{18.000000}{21.600000}\selectfont 64}%
\end{pgfscope}%
\begin{pgfscope}%
\pgfsetbuttcap%
\pgfsetroundjoin%
\definecolor{currentfill}{rgb}{0.000000,0.000000,0.000000}%
\pgfsetfillcolor{currentfill}%
\pgfsetlinewidth{0.803000pt}%
\definecolor{currentstroke}{rgb}{0.000000,0.000000,0.000000}%
\pgfsetstrokecolor{currentstroke}%
\pgfsetdash{}{0pt}%
\pgfsys@defobject{currentmarker}{\pgfqpoint{0.000000in}{-0.048611in}}{\pgfqpoint{0.000000in}{0.000000in}}{%
\pgfpathmoveto{\pgfqpoint{0.000000in}{0.000000in}}%
\pgfpathlineto{\pgfqpoint{0.000000in}{-0.048611in}}%
\pgfusepath{stroke,fill}%
}%
\begin{pgfscope}%
\pgfsys@transformshift{2.642320in}{0.660000in}%
\pgfsys@useobject{currentmarker}{}%
\end{pgfscope}%
\end{pgfscope}%
\begin{pgfscope}%
\definecolor{textcolor}{rgb}{0.000000,0.000000,0.000000}%
\pgfsetstrokecolor{textcolor}%
\pgfsetfillcolor{textcolor}%
\pgftext[x=2.642320in,y=0.562778in,,top]{\color{textcolor}\rmfamily\fontsize{18.000000}{21.600000}\selectfont 128}%
\end{pgfscope}%
\begin{pgfscope}%
\pgfsetbuttcap%
\pgfsetroundjoin%
\definecolor{currentfill}{rgb}{0.000000,0.000000,0.000000}%
\pgfsetfillcolor{currentfill}%
\pgfsetlinewidth{0.803000pt}%
\definecolor{currentstroke}{rgb}{0.000000,0.000000,0.000000}%
\pgfsetstrokecolor{currentstroke}%
\pgfsetdash{}{0pt}%
\pgfsys@defobject{currentmarker}{\pgfqpoint{0.000000in}{-0.048611in}}{\pgfqpoint{0.000000in}{0.000000in}}{%
\pgfpathmoveto{\pgfqpoint{0.000000in}{0.000000in}}%
\pgfpathlineto{\pgfqpoint{0.000000in}{-0.048611in}}%
\pgfusepath{stroke,fill}%
}%
\begin{pgfscope}%
\pgfsys@transformshift{3.614107in}{0.660000in}%
\pgfsys@useobject{currentmarker}{}%
\end{pgfscope}%
\end{pgfscope}%
\begin{pgfscope}%
\definecolor{textcolor}{rgb}{0.000000,0.000000,0.000000}%
\pgfsetstrokecolor{textcolor}%
\pgfsetfillcolor{textcolor}%
\pgftext[x=3.614107in,y=0.562778in,,top]{\color{textcolor}\rmfamily\fontsize{18.000000}{21.600000}\selectfont 256}%
\end{pgfscope}%
\begin{pgfscope}%
\pgfsetbuttcap%
\pgfsetroundjoin%
\definecolor{currentfill}{rgb}{0.000000,0.000000,0.000000}%
\pgfsetfillcolor{currentfill}%
\pgfsetlinewidth{0.803000pt}%
\definecolor{currentstroke}{rgb}{0.000000,0.000000,0.000000}%
\pgfsetstrokecolor{currentstroke}%
\pgfsetdash{}{0pt}%
\pgfsys@defobject{currentmarker}{\pgfqpoint{0.000000in}{-0.048611in}}{\pgfqpoint{0.000000in}{0.000000in}}{%
\pgfpathmoveto{\pgfqpoint{0.000000in}{0.000000in}}%
\pgfpathlineto{\pgfqpoint{0.000000in}{-0.048611in}}%
\pgfusepath{stroke,fill}%
}%
\begin{pgfscope}%
\pgfsys@transformshift{4.585893in}{0.660000in}%
\pgfsys@useobject{currentmarker}{}%
\end{pgfscope}%
\end{pgfscope}%
\begin{pgfscope}%
\definecolor{textcolor}{rgb}{0.000000,0.000000,0.000000}%
\pgfsetstrokecolor{textcolor}%
\pgfsetfillcolor{textcolor}%
\pgftext[x=4.585893in,y=0.562778in,,top]{\color{textcolor}\rmfamily\fontsize{18.000000}{21.600000}\selectfont 512}%
\end{pgfscope}%
\begin{pgfscope}%
\pgfsetbuttcap%
\pgfsetroundjoin%
\definecolor{currentfill}{rgb}{0.000000,0.000000,0.000000}%
\pgfsetfillcolor{currentfill}%
\pgfsetlinewidth{0.803000pt}%
\definecolor{currentstroke}{rgb}{0.000000,0.000000,0.000000}%
\pgfsetstrokecolor{currentstroke}%
\pgfsetdash{}{0pt}%
\pgfsys@defobject{currentmarker}{\pgfqpoint{0.000000in}{-0.048611in}}{\pgfqpoint{0.000000in}{0.000000in}}{%
\pgfpathmoveto{\pgfqpoint{0.000000in}{0.000000in}}%
\pgfpathlineto{\pgfqpoint{0.000000in}{-0.048611in}}%
\pgfusepath{stroke,fill}%
}%
\begin{pgfscope}%
\pgfsys@transformshift{5.557680in}{0.660000in}%
\pgfsys@useobject{currentmarker}{}%
\end{pgfscope}%
\end{pgfscope}%
\begin{pgfscope}%
\definecolor{textcolor}{rgb}{0.000000,0.000000,0.000000}%
\pgfsetstrokecolor{textcolor}%
\pgfsetfillcolor{textcolor}%
\pgftext[x=5.557680in,y=0.562778in,,top]{\color{textcolor}\rmfamily\fontsize{18.000000}{21.600000}\selectfont 1024}%
\end{pgfscope}%
\begin{pgfscope}%
\pgfsetbuttcap%
\pgfsetroundjoin%
\definecolor{currentfill}{rgb}{0.000000,0.000000,0.000000}%
\pgfsetfillcolor{currentfill}%
\pgfsetlinewidth{0.803000pt}%
\definecolor{currentstroke}{rgb}{0.000000,0.000000,0.000000}%
\pgfsetstrokecolor{currentstroke}%
\pgfsetdash{}{0pt}%
\pgfsys@defobject{currentmarker}{\pgfqpoint{0.000000in}{-0.048611in}}{\pgfqpoint{0.000000in}{0.000000in}}{%
\pgfpathmoveto{\pgfqpoint{0.000000in}{0.000000in}}%
\pgfpathlineto{\pgfqpoint{0.000000in}{-0.048611in}}%
\pgfusepath{stroke,fill}%
}%
\begin{pgfscope}%
\pgfsys@transformshift{6.529467in}{0.660000in}%
\pgfsys@useobject{currentmarker}{}%
\end{pgfscope}%
\end{pgfscope}%
\begin{pgfscope}%
\definecolor{textcolor}{rgb}{0.000000,0.000000,0.000000}%
\pgfsetstrokecolor{textcolor}%
\pgfsetfillcolor{textcolor}%
\pgftext[x=6.529467in,y=0.562778in,,top]{\color{textcolor}\rmfamily\fontsize{18.000000}{21.600000}\selectfont 2048}%
\end{pgfscope}%
\begin{pgfscope}%
\definecolor{textcolor}{rgb}{0.000000,0.000000,0.000000}%
\pgfsetstrokecolor{textcolor}%
\pgfsetfillcolor{textcolor}%
\pgftext[x=4.100000in,y=0.293874in,,top]{\color{textcolor}\rmfamily\fontsize{24.000000}{28.800000}\selectfont Rank (R) of factors}%
\end{pgfscope}%
\begin{pgfscope}%
\pgfpathrectangle{\pgfqpoint{1.000000in}{0.660000in}}{\pgfqpoint{6.200000in}{4.620000in}}%
\pgfusepath{clip}%
\pgfsetrectcap%
\pgfsetroundjoin%
\pgfsetlinewidth{0.803000pt}%
\definecolor{currentstroke}{rgb}{0.690196,0.690196,0.690196}%
\pgfsetstrokecolor{currentstroke}%
\pgfsetdash{}{0pt}%
\pgfpathmoveto{\pgfqpoint{1.000000in}{1.029600in}}%
\pgfpathlineto{\pgfqpoint{7.200000in}{1.029600in}}%
\pgfusepath{stroke}%
\end{pgfscope}%
\begin{pgfscope}%
\pgfsetbuttcap%
\pgfsetroundjoin%
\definecolor{currentfill}{rgb}{0.000000,0.000000,0.000000}%
\pgfsetfillcolor{currentfill}%
\pgfsetlinewidth{0.803000pt}%
\definecolor{currentstroke}{rgb}{0.000000,0.000000,0.000000}%
\pgfsetstrokecolor{currentstroke}%
\pgfsetdash{}{0pt}%
\pgfsys@defobject{currentmarker}{\pgfqpoint{-0.048611in}{0.000000in}}{\pgfqpoint{0.000000in}{0.000000in}}{%
\pgfpathmoveto{\pgfqpoint{0.000000in}{0.000000in}}%
\pgfpathlineto{\pgfqpoint{-0.048611in}{0.000000in}}%
\pgfusepath{stroke,fill}%
}%
\begin{pgfscope}%
\pgfsys@transformshift{1.000000in}{1.029600in}%
\pgfsys@useobject{currentmarker}{}%
\end{pgfscope}%
\end{pgfscope}%
\begin{pgfscope}%
\definecolor{textcolor}{rgb}{0.000000,0.000000,0.000000}%
\pgfsetstrokecolor{textcolor}%
\pgfsetfillcolor{textcolor}%
\pgftext[x=0.706947in,y=0.960156in,left,base]{\color{textcolor}\rmfamily\fontsize{14.000000}{16.800000}\selectfont \(\displaystyle 86\)}%
\end{pgfscope}%
\begin{pgfscope}%
\pgfpathrectangle{\pgfqpoint{1.000000in}{0.660000in}}{\pgfqpoint{6.200000in}{4.620000in}}%
\pgfusepath{clip}%
\pgfsetrectcap%
\pgfsetroundjoin%
\pgfsetlinewidth{0.803000pt}%
\definecolor{currentstroke}{rgb}{0.690196,0.690196,0.690196}%
\pgfsetstrokecolor{currentstroke}%
\pgfsetdash{}{0pt}%
\pgfpathmoveto{\pgfqpoint{1.000000in}{1.768800in}}%
\pgfpathlineto{\pgfqpoint{7.200000in}{1.768800in}}%
\pgfusepath{stroke}%
\end{pgfscope}%
\begin{pgfscope}%
\pgfsetbuttcap%
\pgfsetroundjoin%
\definecolor{currentfill}{rgb}{0.000000,0.000000,0.000000}%
\pgfsetfillcolor{currentfill}%
\pgfsetlinewidth{0.803000pt}%
\definecolor{currentstroke}{rgb}{0.000000,0.000000,0.000000}%
\pgfsetstrokecolor{currentstroke}%
\pgfsetdash{}{0pt}%
\pgfsys@defobject{currentmarker}{\pgfqpoint{-0.048611in}{0.000000in}}{\pgfqpoint{0.000000in}{0.000000in}}{%
\pgfpathmoveto{\pgfqpoint{0.000000in}{0.000000in}}%
\pgfpathlineto{\pgfqpoint{-0.048611in}{0.000000in}}%
\pgfusepath{stroke,fill}%
}%
\begin{pgfscope}%
\pgfsys@transformshift{1.000000in}{1.768800in}%
\pgfsys@useobject{currentmarker}{}%
\end{pgfscope}%
\end{pgfscope}%
\begin{pgfscope}%
\definecolor{textcolor}{rgb}{0.000000,0.000000,0.000000}%
\pgfsetstrokecolor{textcolor}%
\pgfsetfillcolor{textcolor}%
\pgftext[x=0.706947in,y=1.699356in,left,base]{\color{textcolor}\rmfamily\fontsize{14.000000}{16.800000}\selectfont \(\displaystyle 88\)}%
\end{pgfscope}%
\begin{pgfscope}%
\pgfpathrectangle{\pgfqpoint{1.000000in}{0.660000in}}{\pgfqpoint{6.200000in}{4.620000in}}%
\pgfusepath{clip}%
\pgfsetrectcap%
\pgfsetroundjoin%
\pgfsetlinewidth{0.803000pt}%
\definecolor{currentstroke}{rgb}{0.690196,0.690196,0.690196}%
\pgfsetstrokecolor{currentstroke}%
\pgfsetdash{}{0pt}%
\pgfpathmoveto{\pgfqpoint{1.000000in}{2.508000in}}%
\pgfpathlineto{\pgfqpoint{7.200000in}{2.508000in}}%
\pgfusepath{stroke}%
\end{pgfscope}%
\begin{pgfscope}%
\pgfsetbuttcap%
\pgfsetroundjoin%
\definecolor{currentfill}{rgb}{0.000000,0.000000,0.000000}%
\pgfsetfillcolor{currentfill}%
\pgfsetlinewidth{0.803000pt}%
\definecolor{currentstroke}{rgb}{0.000000,0.000000,0.000000}%
\pgfsetstrokecolor{currentstroke}%
\pgfsetdash{}{0pt}%
\pgfsys@defobject{currentmarker}{\pgfqpoint{-0.048611in}{0.000000in}}{\pgfqpoint{0.000000in}{0.000000in}}{%
\pgfpathmoveto{\pgfqpoint{0.000000in}{0.000000in}}%
\pgfpathlineto{\pgfqpoint{-0.048611in}{0.000000in}}%
\pgfusepath{stroke,fill}%
}%
\begin{pgfscope}%
\pgfsys@transformshift{1.000000in}{2.508000in}%
\pgfsys@useobject{currentmarker}{}%
\end{pgfscope}%
\end{pgfscope}%
\begin{pgfscope}%
\definecolor{textcolor}{rgb}{0.000000,0.000000,0.000000}%
\pgfsetstrokecolor{textcolor}%
\pgfsetfillcolor{textcolor}%
\pgftext[x=0.706947in,y=2.438556in,left,base]{\color{textcolor}\rmfamily\fontsize{14.000000}{16.800000}\selectfont \(\displaystyle 90\)}%
\end{pgfscope}%
\begin{pgfscope}%
\pgfpathrectangle{\pgfqpoint{1.000000in}{0.660000in}}{\pgfqpoint{6.200000in}{4.620000in}}%
\pgfusepath{clip}%
\pgfsetrectcap%
\pgfsetroundjoin%
\pgfsetlinewidth{0.803000pt}%
\definecolor{currentstroke}{rgb}{0.690196,0.690196,0.690196}%
\pgfsetstrokecolor{currentstroke}%
\pgfsetdash{}{0pt}%
\pgfpathmoveto{\pgfqpoint{1.000000in}{3.247200in}}%
\pgfpathlineto{\pgfqpoint{7.200000in}{3.247200in}}%
\pgfusepath{stroke}%
\end{pgfscope}%
\begin{pgfscope}%
\pgfsetbuttcap%
\pgfsetroundjoin%
\definecolor{currentfill}{rgb}{0.000000,0.000000,0.000000}%
\pgfsetfillcolor{currentfill}%
\pgfsetlinewidth{0.803000pt}%
\definecolor{currentstroke}{rgb}{0.000000,0.000000,0.000000}%
\pgfsetstrokecolor{currentstroke}%
\pgfsetdash{}{0pt}%
\pgfsys@defobject{currentmarker}{\pgfqpoint{-0.048611in}{0.000000in}}{\pgfqpoint{0.000000in}{0.000000in}}{%
\pgfpathmoveto{\pgfqpoint{0.000000in}{0.000000in}}%
\pgfpathlineto{\pgfqpoint{-0.048611in}{0.000000in}}%
\pgfusepath{stroke,fill}%
}%
\begin{pgfscope}%
\pgfsys@transformshift{1.000000in}{3.247200in}%
\pgfsys@useobject{currentmarker}{}%
\end{pgfscope}%
\end{pgfscope}%
\begin{pgfscope}%
\definecolor{textcolor}{rgb}{0.000000,0.000000,0.000000}%
\pgfsetstrokecolor{textcolor}%
\pgfsetfillcolor{textcolor}%
\pgftext[x=0.706947in,y=3.177756in,left,base]{\color{textcolor}\rmfamily\fontsize{14.000000}{16.800000}\selectfont \(\displaystyle 92\)}%
\end{pgfscope}%
\begin{pgfscope}%
\pgfpathrectangle{\pgfqpoint{1.000000in}{0.660000in}}{\pgfqpoint{6.200000in}{4.620000in}}%
\pgfusepath{clip}%
\pgfsetrectcap%
\pgfsetroundjoin%
\pgfsetlinewidth{0.803000pt}%
\definecolor{currentstroke}{rgb}{0.690196,0.690196,0.690196}%
\pgfsetstrokecolor{currentstroke}%
\pgfsetdash{}{0pt}%
\pgfpathmoveto{\pgfqpoint{1.000000in}{3.986400in}}%
\pgfpathlineto{\pgfqpoint{7.200000in}{3.986400in}}%
\pgfusepath{stroke}%
\end{pgfscope}%
\begin{pgfscope}%
\pgfsetbuttcap%
\pgfsetroundjoin%
\definecolor{currentfill}{rgb}{0.000000,0.000000,0.000000}%
\pgfsetfillcolor{currentfill}%
\pgfsetlinewidth{0.803000pt}%
\definecolor{currentstroke}{rgb}{0.000000,0.000000,0.000000}%
\pgfsetstrokecolor{currentstroke}%
\pgfsetdash{}{0pt}%
\pgfsys@defobject{currentmarker}{\pgfqpoint{-0.048611in}{0.000000in}}{\pgfqpoint{0.000000in}{0.000000in}}{%
\pgfpathmoveto{\pgfqpoint{0.000000in}{0.000000in}}%
\pgfpathlineto{\pgfqpoint{-0.048611in}{0.000000in}}%
\pgfusepath{stroke,fill}%
}%
\begin{pgfscope}%
\pgfsys@transformshift{1.000000in}{3.986400in}%
\pgfsys@useobject{currentmarker}{}%
\end{pgfscope}%
\end{pgfscope}%
\begin{pgfscope}%
\definecolor{textcolor}{rgb}{0.000000,0.000000,0.000000}%
\pgfsetstrokecolor{textcolor}%
\pgfsetfillcolor{textcolor}%
\pgftext[x=0.706947in,y=3.916956in,left,base]{\color{textcolor}\rmfamily\fontsize{14.000000}{16.800000}\selectfont \(\displaystyle 94\)}%
\end{pgfscope}%
\begin{pgfscope}%
\pgfpathrectangle{\pgfqpoint{1.000000in}{0.660000in}}{\pgfqpoint{6.200000in}{4.620000in}}%
\pgfusepath{clip}%
\pgfsetrectcap%
\pgfsetroundjoin%
\pgfsetlinewidth{0.803000pt}%
\definecolor{currentstroke}{rgb}{0.690196,0.690196,0.690196}%
\pgfsetstrokecolor{currentstroke}%
\pgfsetdash{}{0pt}%
\pgfpathmoveto{\pgfqpoint{1.000000in}{4.725600in}}%
\pgfpathlineto{\pgfqpoint{7.200000in}{4.725600in}}%
\pgfusepath{stroke}%
\end{pgfscope}%
\begin{pgfscope}%
\pgfsetbuttcap%
\pgfsetroundjoin%
\definecolor{currentfill}{rgb}{0.000000,0.000000,0.000000}%
\pgfsetfillcolor{currentfill}%
\pgfsetlinewidth{0.803000pt}%
\definecolor{currentstroke}{rgb}{0.000000,0.000000,0.000000}%
\pgfsetstrokecolor{currentstroke}%
\pgfsetdash{}{0pt}%
\pgfsys@defobject{currentmarker}{\pgfqpoint{-0.048611in}{0.000000in}}{\pgfqpoint{0.000000in}{0.000000in}}{%
\pgfpathmoveto{\pgfqpoint{0.000000in}{0.000000in}}%
\pgfpathlineto{\pgfqpoint{-0.048611in}{0.000000in}}%
\pgfusepath{stroke,fill}%
}%
\begin{pgfscope}%
\pgfsys@transformshift{1.000000in}{4.725600in}%
\pgfsys@useobject{currentmarker}{}%
\end{pgfscope}%
\end{pgfscope}%
\begin{pgfscope}%
\definecolor{textcolor}{rgb}{0.000000,0.000000,0.000000}%
\pgfsetstrokecolor{textcolor}%
\pgfsetfillcolor{textcolor}%
\pgftext[x=0.706947in,y=4.656156in,left,base]{\color{textcolor}\rmfamily\fontsize{14.000000}{16.800000}\selectfont \(\displaystyle 96\)}%
\end{pgfscope}%
\begin{pgfscope}%
\definecolor{textcolor}{rgb}{0.000000,0.000000,0.000000}%
\pgfsetstrokecolor{textcolor}%
\pgfsetfillcolor{textcolor}%
\pgftext[x=0.651391in,y=2.970000in,,bottom,rotate=90.000000]{\color{textcolor}\rmfamily\fontsize{24.000000}{28.800000}\selectfont Accuracy}%
\end{pgfscope}%
\begin{pgfscope}%
\pgfpathrectangle{\pgfqpoint{1.000000in}{0.660000in}}{\pgfqpoint{6.200000in}{4.620000in}}%
\pgfusepath{clip}%
\pgfsetbuttcap%
\pgfsetroundjoin%
\pgfsetlinewidth{1.505625pt}%
\definecolor{currentstroke}{rgb}{0.000000,0.000000,0.000000}%
\pgfsetstrokecolor{currentstroke}%
\pgfsetdash{}{0pt}%
\pgfpathmoveto{\pgfqpoint{1.670533in}{1.476816in}}%
\pgfpathlineto{\pgfqpoint{1.670533in}{2.873904in}}%
\pgfusepath{stroke}%
\end{pgfscope}%
\begin{pgfscope}%
\pgfpathrectangle{\pgfqpoint{1.000000in}{0.660000in}}{\pgfqpoint{6.200000in}{4.620000in}}%
\pgfusepath{clip}%
\pgfsetbuttcap%
\pgfsetroundjoin%
\pgfsetlinewidth{1.505625pt}%
\definecolor{currentstroke}{rgb}{0.000000,0.000000,0.000000}%
\pgfsetstrokecolor{currentstroke}%
\pgfsetdash{}{0pt}%
\pgfpathmoveto{\pgfqpoint{2.642320in}{1.850112in}}%
\pgfpathlineto{\pgfqpoint{2.642320in}{2.833248in}}%
\pgfusepath{stroke}%
\end{pgfscope}%
\begin{pgfscope}%
\pgfpathrectangle{\pgfqpoint{1.000000in}{0.660000in}}{\pgfqpoint{6.200000in}{4.620000in}}%
\pgfusepath{clip}%
\pgfsetbuttcap%
\pgfsetroundjoin%
\pgfsetlinewidth{1.505625pt}%
\definecolor{currentstroke}{rgb}{0.000000,0.000000,0.000000}%
\pgfsetstrokecolor{currentstroke}%
\pgfsetdash{}{0pt}%
\pgfpathmoveto{\pgfqpoint{3.614107in}{2.160576in}}%
\pgfpathlineto{\pgfqpoint{3.614107in}{3.372864in}}%
\pgfusepath{stroke}%
\end{pgfscope}%
\begin{pgfscope}%
\pgfpathrectangle{\pgfqpoint{1.000000in}{0.660000in}}{\pgfqpoint{6.200000in}{4.620000in}}%
\pgfusepath{clip}%
\pgfsetbuttcap%
\pgfsetroundjoin%
\pgfsetlinewidth{1.505625pt}%
\definecolor{currentstroke}{rgb}{0.000000,0.000000,0.000000}%
\pgfsetstrokecolor{currentstroke}%
\pgfsetdash{}{0pt}%
\pgfpathmoveto{\pgfqpoint{4.585893in}{2.740848in}}%
\pgfpathlineto{\pgfqpoint{4.585893in}{4.167504in}}%
\pgfusepath{stroke}%
\end{pgfscope}%
\begin{pgfscope}%
\pgfpathrectangle{\pgfqpoint{1.000000in}{0.660000in}}{\pgfqpoint{6.200000in}{4.620000in}}%
\pgfusepath{clip}%
\pgfsetbuttcap%
\pgfsetroundjoin%
\pgfsetlinewidth{1.505625pt}%
\definecolor{currentstroke}{rgb}{0.000000,0.000000,0.000000}%
\pgfsetstrokecolor{currentstroke}%
\pgfsetdash{}{0pt}%
\pgfpathmoveto{\pgfqpoint{5.557680in}{4.116130in}}%
\pgfpathlineto{\pgfqpoint{5.557680in}{4.773278in}}%
\pgfusepath{stroke}%
\end{pgfscope}%
\begin{pgfscope}%
\pgfpathrectangle{\pgfqpoint{1.000000in}{0.660000in}}{\pgfqpoint{6.200000in}{4.620000in}}%
\pgfusepath{clip}%
\pgfsetbuttcap%
\pgfsetroundjoin%
\pgfsetlinewidth{1.505625pt}%
\definecolor{currentstroke}{rgb}{0.000000,0.000000,0.000000}%
\pgfsetstrokecolor{currentstroke}%
\pgfsetdash{}{0pt}%
\pgfpathmoveto{\pgfqpoint{6.529467in}{4.396656in}}%
\pgfpathlineto{\pgfqpoint{6.529467in}{4.781040in}}%
\pgfusepath{stroke}%
\end{pgfscope}%
\begin{pgfscope}%
\pgfsetrectcap%
\pgfsetmiterjoin%
\pgfsetlinewidth{0.803000pt}%
\definecolor{currentstroke}{rgb}{0.000000,0.000000,0.000000}%
\pgfsetstrokecolor{currentstroke}%
\pgfsetdash{}{0pt}%
\pgfpathmoveto{\pgfqpoint{1.000000in}{0.660000in}}%
\pgfpathlineto{\pgfqpoint{1.000000in}{5.280000in}}%
\pgfusepath{stroke}%
\end{pgfscope}%
\begin{pgfscope}%
\pgfsetrectcap%
\pgfsetmiterjoin%
\pgfsetlinewidth{0.803000pt}%
\definecolor{currentstroke}{rgb}{0.000000,0.000000,0.000000}%
\pgfsetstrokecolor{currentstroke}%
\pgfsetdash{}{0pt}%
\pgfpathmoveto{\pgfqpoint{7.200000in}{0.660000in}}%
\pgfpathlineto{\pgfqpoint{7.200000in}{5.280000in}}%
\pgfusepath{stroke}%
\end{pgfscope}%
\begin{pgfscope}%
\pgfsetrectcap%
\pgfsetmiterjoin%
\pgfsetlinewidth{0.803000pt}%
\definecolor{currentstroke}{rgb}{0.000000,0.000000,0.000000}%
\pgfsetstrokecolor{currentstroke}%
\pgfsetdash{}{0pt}%
\pgfpathmoveto{\pgfqpoint{1.000000in}{0.660000in}}%
\pgfpathlineto{\pgfqpoint{7.200000in}{0.660000in}}%
\pgfusepath{stroke}%
\end{pgfscope}%
\begin{pgfscope}%
\pgfsetrectcap%
\pgfsetmiterjoin%
\pgfsetlinewidth{0.803000pt}%
\definecolor{currentstroke}{rgb}{0.000000,0.000000,0.000000}%
\pgfsetstrokecolor{currentstroke}%
\pgfsetdash{}{0pt}%
\pgfpathmoveto{\pgfqpoint{1.000000in}{5.280000in}}%
\pgfpathlineto{\pgfqpoint{7.200000in}{5.280000in}}%
\pgfusepath{stroke}%
\end{pgfscope}%
\end{pgfpicture}%
\makeatother%
\endgroup%